\documentclass[11pt]{article}

\def\modifymargins#1#2{
\newdimen\addtoh
\newdimen\addtow
\addtoh=#1
\addtow=#2

\advance\topmargin by -\addtoh
\multiply\addtoh by 2
\advance\textheight by \addtoh

\advance\oddsidemargin by -\addtow
\advance\evensidemargin by -\addtow
\multiply\addtow by 2
\advance\textwidth by \addtow
}

\expandafter\ifx\csname ttytty\endcsname\relax
\modifymargins{40pt}{30pt}
\else\modifymargins{60pt}{0pt}\def\_{\tt\char 95}
\raggedright\parindent 40pt

\def\D#1{{\bf D#1}}
\def\Dlog#1{{\bf Dlog#1}}
\fi
\let\xinput=\input
\def\ttyfig#1#2{\xinput{#1.latex}}
\def\ttytex#1{#1\nop}

\def\possnewtheorem#1#2{
\expandafter\ifx\csname #1\endcsname\relax
\newtheorem{#1}{#2}
\fi
}

\def\possnewtheoremthree#1[#2]#3{
\expandafter\ifx\csname #1\endcsname\relax
\newtheorem{#1}[#2]{#3}
\fi
}

\possnewtheorem{theorem}{Theorem}
\possnewtheorem{corollary}{Corollary}
\possnewtheorem{lemma}{Lemma}
\possnewtheoremthree{proposition}[theorem]{Proposition}
\possnewtheorem{definition}{Definition}
\possnewtheorem{question}{Question}
\possnewtheorem{example}{Example}
\possnewtheorem{nontheorem}{Counterexample}
\possnewtheorem{property}{Property}
\possnewtheorem{assumption}{Assumption}
\possnewtheorem{conjecture}{Conjecture}
\possnewtheorem{notation}{Notation}

\def\proof{\noindent {\sl Proof.\ \ }}
\def\qed{\hfill{\boxit{}}
  \ifdim\lastskip<\medskipamount \removelastskip\penalty55\medskip\fi}
\long\def\boxit#1{\vbox{\hrule\hbox{\vrule\kern3pt
                  \vbox{\kern3pt#1\kern3pt}\kern3pt\vrule}\hrule}}

\expandafter\ifx\csname shortcite\endcsname\relax
\let\shortcite=\cite
\fi

\def\np{{\rm NP}}
\def\conp{{\rm coNP}}
\def\bh#1{\if#1{}{\rm BH}\else\mbox{BH$_{#1}$}\fi}
\def\Dp{${\rm D}^p$}
\def\D#1{\mbox{$\Delta^p_{#1}$}}
\def\Dlog#1{\mbox{$\Delta^p_{#1}[\log n]$}}
\def\P#1{\mbox{$\Pi^p_{#1}$}}

\def\mod{M\!od}
\def\form{F\!orm}
\def\color[#1]#2{}
\def\nop#1{}
\let\plural=\relax

\def\ope{\mathrm{ope}}		
\def\rev{\mathrm{rev}}		
\def\lex{\mathrm{lex}}		
\def\refi{\mathrm{ref}}		
\def\sev{\mathrm{sev}}		
\def\nat{\mathrm{nat}}		
\def\res{\mathrm{res}}		
\def\rad{\mathrm{rad}}		
\def\sevr{\mathrm{sevr}}	
\def\psev{\mathrm{psev}}	
\def\msev{\mathrm{msev}}	
\def\full{\mathrm{full}}	

\def\maxset{\mathrm{maxset}}

\def\under{\mathrm{under}}
\def\longest{\mathrm{longest}}

\title{On Mixed Iterated Revisions}
\author{Paolo Liberatore~\thanks{
DIIAG, Sapienza University of Rome.
{\tt liberato@diag.uniroma1.it}
}}

\begin{document}

\maketitle

\begin{abstract}

Several forms of iterable belief change exist, differing in the kind of change
and its strength: some operators introduce formulae, others remove them; some
add formulae unconditionally, others only as additions to the previous beliefs;
some only relative to the current situation, others in all possible cases. A
sequence of changes may involve several of them: for example, the first step is
a revision, the second a contraction and the third a refinement of the previous
beliefs. The ten operators considered in this article are shown to be all
reducible to three: lexicographic revision, refinement and severe withdrawal.
In turn, these three can be expressed in terms of lexicographic revision at the
cost of restructuring the sequence. This restructuring needs not to be done
explicitly: an algorithm that works on the original sequence is shown. The
complexity of mixed sequences of belief change operators is also analyzed. Most
of them require only a polynomial number of calls to a satisfiability checker,
some are even easier.

\end{abstract}

\sloppy


\section{Introduction}

New information come in different forms. At the one end of the spectrum, it is
believed to be always true (lexicographic revision~\cite{spoh-88,naya-94}); at
the far opposite, it is known to be false
(contraction~\cite{alch-gard-maki-85,saue-etal-20-b}). Middle cases exist: it
may be believed only as long as the current situation is concerned (natural
revision~\cite{bout-96-a,koni-98}), or it may be believed only as long as it
does not contradict the previous beliefs
(refinement~\cite{papi-01,boot-etal-06}. Sequence of changes are hardly all of
the same form, like someone who embraces every single new theory with all of
his hearth or instead so skeptical to refuse every piece of information that
contradicts what known.

\begin{example}

An example against natural revision~\cite{bout-96-a,koni-98} is that of the red
bird~\cite{jin-thie-07}: an animal looks like a bird ($b$) and upon coming
close turned out to be red ($r$); finding out not to be a bird ($\neg b$) makes
natural revision discard it being red in spite of no evidence of the contrary.

This is a situation where natural revision is not to be used. Yet, a small
variant of the conditions turns the very same formulae, with the very same
meaning of variables, into a case for it.

A hunter meets a peasant friend in the countryside, who told him having seen a
strange bird in the thicket a couple of miles away ($b$). Lured by the unique
trophy he could make out of it, and by bragging about its hunting at the
village f\^ete that evening, the hunter rushes to there.

Midway, he encounters the village postman. In a hurry, the hunter explains he
could not stop and why. The postman answers he understands, as he actually saw
something red in the thicket ($r$). This would explain why the peasant friend
told the bird was strange, because no red bird has ever been seen around there.

Arrived at the thicket, the hunter checks for anything red but the bushes and
trees are too thick to see anything inside. Entering is out of discussion, as
any bird would fly away upon hearing the noise. Keeping ready with his rifle,
the hunter throws a stone in the thicket, but nothing happens. A second and
third stone confirm that no bird is there. The peasant friend and the postman
made fun of him by having him run with no reason. No bird is there ($\neg b$),
nothing indicates something red ($r$ is no longer believed).

While the sequence of formulae is exactly the same (first $b$, then $r$,
finally $\neg b$), with the same meaning of the variables ($b$=bird, $r$=red),
natural revision gives exactly the expected result: since the first piece of
information was made up, there is no reason to believe the second.

Should hunters always use natural revisions?

Of course not.

Seeing the hunter throwing stones for no clear reason, the village policeman
approaches the hunter to ask why. Concealing how he was made a fool is useless,
as the peasant friend and the postman will tell the story to everyone at the
village f\^ete that evening; everyone will know even before the parade. The
policeman laughs, but to the hunter's surprise it's not about the practical
joke. A red animal would be unique in the area. Even if one was there, and the
hunter managed to shot it, he would have been fined and confiscated the trophy.
He could not have shown it in his living room, and certainly not brag about it
at the village f\^ete. If it is red, it cannot be hunted ($r \rightarrow \neg
h$).

As it comes from a police officer while on duty, this information is totally
reliable. It is also a general rule, not specific to the current situation. No
matter if it was a bird ($b$ or $\neg b$), if it is red it cannot be hunted ($r
\rightarrow \neg h$). The situation where a red bird that can be hunted was
there ($b$, $r$ and $h$) is less likely than one where it can ($b$, $r$, $\neg
h$), and the same if it is red but not a bird ($\neg b$, $r$ and $h$ is less
likely than $\neg b$, $r$ and $\neg h$). That hunting it is forbidden is more
likely in every situation, even if a red animal that is not a bird later turn
out to be there.

The hunter is better revising lexicographically (by $\neg h$) this time if he
wants to avoid being fined, after previously having revised naturally (by $\neg
b$). A mixed sequence of revisions is the best course of actions.

A related question is why to apply a sequence of revisions in the first place.
Why not just coming back to where the hunter started, with no information about
the thicket, rather than revising by $b$, then $r$ and then by $\neg b$? Once
the bird is nowhere to be found, everything could be just canceled. The
epistemic state is the same as the beginning.

While talking with the policeman, the hunter hears a sound of feathers from the
thicket. Feathers mean bird ($b$). Maybe one that does not fly, or is wounded
and unable to fly. The peasant and the postman might have been truthful, after
all. Nothing indicates they joked any longer. None of them, including the
postman. The bird might be red ($r$) after all. This would not be the case if
the hunter just forgot everything was told.

\end{example}

Natural revision, then lexicographic revision. Not two lexicographic revisions,
not two natural revisions. A mixed sequence of revisions is what to do in this
example. As Booth and Meyer~\shortcite{boot-meye-06} put it: "it is clear that
an agent need not, and in most cases, ought not to stick to the same revision
operator every time that it has to perform a revision."

Other forms of revisions exist. They may mix in any order. A lexicographic
revision may be followed by a natural revision, a restrained revision and a
radical revision. Such a sequence is used as a running example:

\[
\emptyset \lex(y) \nat(\neg x) \res(x \wedge z) \rad(\neg z)
\]

The sequence begins with $\emptyset$, the complete lack of knowledge. The first
information acquired is $y$, and is considered valid in all possible
situations; it is a lexicographic revision. The next is $\neg x$, but is deemed
valid only in the current conditions; this is a natural revision. The next is
$x \wedge z$, but is only accepted as long as it does not contradict what is
currently known; it is a refinement. Finally, $\neg z$ is so firmly believed
that all cases where it does not hold are excluded; this is a radical revision.

No semantics is better than the others. Natural revision has its place. As well
as lexicographic revision. As well as refinement, radical revision, severe
antiwithdrawals and severe revision. None of them is the best. Each is right in
the appropriate conditions and wrong in the others. For example, a sequence of
severe revisions is problematic because it coarsens the plausibility degree of
different scenarios~\cite{ferm-rott-04,rott-06}; yet, it is a common experience
that sometimes new information makes unlikely conditions as likely as others. A
truck full of parrots to be sold as pets crushed nearby, freeing thousands of
exotic animals; a red bird is now as likely to be stumbled upon as a local wild
animal. New information may raise the plausibility of a situation to the level
of another, making them equally likely. This is a consequence of the new
information, not a fault of the revision operator. The problem comes when using
severe revisions only, without other revisions that separate the likeliness of
different conditions~\cite{ferm-rott-04,rott-06}.

The solution is not to search for a new framework that encompasses all possible
cases, but to deal with mixes of different revision kinds. How to decide which
type of revision to use at each time depends on how the information has been
obtained and on the information itself. This is a separate
problem~\cite{boot-nitt-08,libe-16}, not considered here.

The problem considered here is to determine the outcome of a sequence of mixed
revisions. Semantically, each modifies the order of plausibility of the models
in a different way. For example, natural revision ``promotes'' some models to
the maximal plausibilty; lexicographic revision makes some models being more
plausible than others. Keeping in memory a complete description of these
orderings is unfeasible, even in the propositional case: the number of models
is exponential in the number of variables. Several operators such as
lexicographic revision, refinement and restrained revision can generate
orderings that compare equal no two models, making exponential every
representation that is linear in the number of equivalence classes of the
ordering, such as the ordered sequence of formulae by Rott~\cite{rott-09} and
the partitions by Meyer, Ghose and Chopra~\cite{meye-etal-02}. Many
distance-based one-step revisions suffer from a similar problem: the result of
even a single revision may be exponential in the size of the involved
formulae~\cite{cado-etal-00-b}. Iterated revisions typically do not employ
distances, and the problem can be overcome:

\begin{itemize}

\item the ten belief change operators considered in this article (lexicographic
revision, refinement, severe withdrawal, natural revision, restrained revision,
plain severe revision, severe revision, moderate severe revision, very radical
revision) can be reduced to three: lexicographic revision, refinement and
severe withdrawal; these reductions are local: they replace an operator without
changing the rest of the sequence before and after it;

\item refinement and severe withdrawal can be reduced to lexicographic
revision; this however requires structuring the sequence of belief change
operators; however, the result is a sequence that behave like the original on
subsequent changes;

\item this restructuring needs not to be done explicitly; an algorithm that
works on the original sequence is shown; it does not change the sequence, but
behaves as if it were restructured; apart from the calls to a satisfiability
checker, the running time is polynomial.

\end{itemize}

This mechanism determines the result of an arbitrary sequence of revisions from
an initial ordering that reflects a total lack of information. This is not a
limitation, as an arbitrary ordering can be created by a sequence of
lexicographic revisions~\cite{boot-nitt-08}.

During its execution, the algorithm calculates some partial results called
unformulae, which can be used when the sequence is extended by further
revisions. The need for a satisfiability checker is unavoidable, given that
belief change operates on propositional formulae. However, efficient solvers
have been developed~\cite{bier-etal-09,baly-etal-17}. Restricting to a less
expressive language~\cite{crei-etal-18} such as Horn logic may also reduce the
complexity of the problem, as it is generally the case for one-step
revisions~\cite{eite-gott-91-b,nebe-98,libe-scha-00,lang-etal-08}, since
satisfiability in this case can be solved efficiently.

Some complexity results are proved: some imply the ones announced without
proofs in a previous article~\cite{libe-97-c}, but extend them to the case of
mixed sequences of revisions. Entailment from a sequence of lexicographic,
natural, restrained, very radical and severe revisions, refinements and severe
antiwithdrawals is in the complexity class \D{2}, and is \D{2}-hard even if the
sequence contains only lexicographic revisions and refinements. Two groups of
belief change operators are relevant to complexity. The first is called
lexicographic-finding and comprises the ones that behave like lexicographic
revision on consistent sequences of formulae; lexicographic and moderate severe
revisions are in this group. The second is called bottom-refining as it
includes the revisions that separate the most likely scenarios when some are
consistent with the new information; natural revision, restrained revision and
severe revision are in this group. Entailment from a sequence of operators all
of the first kind or all of the second is \D{2}-complete. Three revision
operators require a separate analysis. Entailment from a sequence of very
radical revision is \Dlog{2}-complete. The same complexity comes from sequences
of plain severe and full meet revisions only.

The rest of this article is organized as follows: Section~\ref{preliminaries}
introduces the main concepts of total preorder, lexicographic revision,
refinement and severe withdrawal; Section~\ref{reductions} shows how to reduce
the other change operators to these three; Section~\ref{method} shows an
algorithm for computing the result of a sequence of revisions;
Section~\ref{hardness} presents the computational complexity results;
Section~\ref{conclusions} discusses the results in this article, compares them
with other work in the literature and presents the open problems.

\section{Preliminaries}
\label{preliminaries}

A propositional language over a finite alphabet is assumed. Given a formula
$F$, its set of models is denoted $\mod(F)$, while a formula having $S$ as its
set of models is denoted $\form(S)$. The symbol $\top$ denotes a tautology, a
formula satisfied by all models over the given alphabet.

A {\em base} is a propositional formula denoting what is believed in certain
moment. Historically, revision was defined as an operator that modifies a base
in front of new information; an ordering was employed to take choices when this
integration may be done in multiple ways, which is usually the case. Assuming
this ordering as fixed or depending on the base only is the AGM model or
revision~\cite{alch-gard-maki-85,gard-88}. Iterated revision is problematic
using this approach; the solution is to reverse the role of the base and the
ordering. Instead of being a supporting element, the ordering becomes the
protagonist. The base derives from it as the set of most plausible
formulae~\cite{darw-pear-97,koni-pere-17}. Such plausibility information can be
formalized in several equivalent ways: epistemic
entrenchments~\cite{gard-maki-88,ferm-reis-13}, systems of
spheres~\cite{grov-88,girl-etal-17},
rankings~\cite{will-94,spoh-99,kern-etal-21}, and
KM~preorders~\cite{kats-mend-91-b,pepp-etal-08}.

\subsection{Total preorders}

Katsuno and Mendelzon~\cite{kats-mend-91-b} proved that AGM revision can be
reformulated in terms of a total preorder over the set of models, where the
models of the base are exactly the minimal ones according to the ordering.
Iterated revision can be defined by demoting the base from primary information
to derived one. Instead of revising a base using the ordering as a guide, the
ordering itself is modified. The base is taken to be just the set of formulae
implied by all most plausible models.

\begin{definition}

A {\em total preorder} $C$ is a partition of the models into a finite sequence
of classes $[C(0),C(1),C(2),\ldots,C(m)]$.

\end{definition}

Such an ordering can be depicted as a stack, the top boxes containing the most
plausible models. This is equivalent to a reflexive, transitive and total
relation, but makes for simpler definitions and proofs about iterated
revisions.

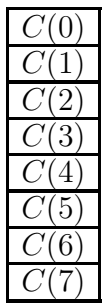
\setlength{\unitlength}{5000sp}%
\begingroup\makeatletter\ifx\SetFigFont\undefined%
\gdef\SetFigFont#1#2#3#4#5{%
  \reset@font\fontsize{#1}{#2pt}%
  \fontfamily{#3}\fontseries{#4}\fontshape{#5}%
  \selectfont}%
\fi\endgroup%
\begin{picture}(516,1506)(4828,-4144)
\thicklines
{\color[rgb]{0,0,0}\put(4861,-2851){\line( 1, 0){450}}
}%
{\color[rgb]{0,0,0}\put(4861,-3031){\line( 1, 0){450}}
}%
{\color[rgb]{0,0,0}\put(4861,-3211){\line( 1, 0){450}}
}%
{\color[rgb]{0,0,0}\put(4861,-3391){\line( 1, 0){450}}
}%
{\color[rgb]{0,0,0}\put(4861,-3571){\line( 1, 0){450}}
}%
{\color[rgb]{0,0,0}\put(4861,-3751){\line( 1, 0){450}}
}%
{\color[rgb]{0,0,0}\put(4861,-3931){\line( 1, 0){450}}
}%
{\color[rgb]{0,0,0}\put(4861,-4111){\framebox(450,1440){}}
}%
\put(5086,-2806){\makebox(0,0)[b]{\smash{{\SetFigFont{12}{24.0}{\rmdefault}{\mddefault}{\itdefault}{\color[rgb]{0,0,0}$C(0)$}%
}}}}
\put(5086,-2986){\makebox(0,0)[b]{\smash{{\SetFigFont{12}{24.0}{\rmdefault}{\mddefault}{\itdefault}{\color[rgb]{0,0,0}$C(1)$}%
}}}}
\put(5086,-3166){\makebox(0,0)[b]{\smash{{\SetFigFont{12}{24.0}{\rmdefault}{\mddefault}{\itdefault}{\color[rgb]{0,0,0}$C(2)$}%
}}}}
\put(5086,-3346){\makebox(0,0)[b]{\smash{{\SetFigFont{12}{24.0}{\rmdefault}{\mddefault}{\itdefault}{\color[rgb]{0,0,0}$C(3)$}%
}}}}
\put(5086,-3526){\makebox(0,0)[b]{\smash{{\SetFigFont{12}{24.0}{\rmdefault}{\mddefault}{\itdefault}{\color[rgb]{0,0,0}$C(4)$}%
}}}}
\put(5086,-3706){\makebox(0,0)[b]{\smash{{\SetFigFont{12}{24.0}{\rmdefault}{\mddefault}{\itdefault}{\color[rgb]{0,0,0}$C(5)$}%
}}}}
\put(5086,-3886){\makebox(0,0)[b]{\smash{{\SetFigFont{12}{24.0}{\rmdefault}{\mddefault}{\itdefault}{\color[rgb]{0,0,0}$C(6)$}%
}}}}
\put(5086,-4066){\makebox(0,0)[b]{\smash{{\SetFigFont{12}{24.0}{\rmdefault}{\mddefault}{\itdefault}{\color[rgb]{0,0,0}$C(7)$}%
}}}}
\end{picture}%
\nop{
 +------+
 | C(0) |
 +------+
 | C(1) |
 +------+
 | C(2) |
 +------+
 | C(3) |
 +------+
 | C(4) |
 +------+
 | C(5) |
 +------+
 | C(6) |
 +------+
 | C(7) |
 +------+
}

A KM total preorder is the same as a partition by Mayer, Ghose and
Chopra~\cite{meye-etal-02}, who use a formula for each class in place of its
set of models. In turns, such a partition is similar to the system for
expressing such orderings in possibilistic logic~\cite{benf-etal-01}, and
correspond to a sequence of formulae by Rott~\cite{rott-09}
{} $h_1 \prec \cdots \prec h_m$ via 
{} $C(0) \cup \ldots \cup C(i) = \mod(h_{i+1} \wedge \cdots \wedge h_m)$
and to an epistemic entrenchment~\cite{alch-gard-maki-85}.

Being a partition, $C=[C(0),\ldots,C(m)]$ contains all models. As a result,
every model is in a class. No model is ``inaccessible'', or excluded from
consideration when performing revision. Revisions producing such models could
still be formalized by giving a special status to the last class $C(m)$, as the
set of such inaccessible models, but they are not studied in this article.
Their analysis is left as an open problem.

Classes are allowed to be empty, even class zero $C(0)$. The base represented
by a total preorder $C=[C(0),\ldots,C(m)]$ cannot therefore being defined as
$\form(C(0))$ but as the minimal models according to $C$, denoted by
$\mod(C,\top)$.

More generally, given a formula $P$ the notation $\min(C,P)$ indicates the set
of minimal models of $P$ according to the ordering $C$. Formally, if $i$ is the
lowest index such that $C(i) \cap \mod(P)$ is not empty, then $\min(C,P)=C(i)
\cap \mod(P)$. Several iterated revision depends on such an index $i$ and its
corresponding set of models $C(i) \cap \mod(P)$.

Another consequence of allowing empty classes is that two total preorder may be
different yet comparing models in the same way. For example, $[\mod(T)]$ and
$[\emptyset,\mod(T)]$ both place all models in the same class, which is class
zero for the former and class one for the latter. They are in this sense
equivalent. They coincide when removing the empty classes. The minimal models
of every formula are the same~\cite{papi-01}.

\begin{definition}

Two total preorder $C$ and $C'$ are equivalent if $\min(C,P)=\min(C,P')$ for
every formula $P$.

\end{definition}

Revising by the same formula modifies equivalent orderings into equivalent
orderings. This holds for all revision semantics considered in this article.

The amount of information an ordering carries can be informally identified with
its ability of telling the relative plausibility of two models. Ideally, an
ordering should have a single minimal model, representing what is believed to
be the state of the world, and a single model in each class, allowing to
unambiguously decide which among two possible states of the world is the most
likely. Most revision indeed refine the ordering by splitting its classes. At
the other end of the spectrum, the total order $\emptyset=[\mod(T)]$ carries no
information: not only its base comprises all models and is therefore
tautological, but all models are also considered equally plausible. Studies on
the practical use of revision~\cite{libe-97-c,boot-nitt-08} assume an initial
empty ordering that is then revised to obtain a more discriminating one.
Equivalently, an ordering can be expressed as a suitable sequence of revisions
applied to the empty total preorder.

Not all operators considered in this article are revisions, only the ones that
produce an ordering whose base implies the revising formula. Some other
operators just split classes (like refinement) or merge them (like severe
withdrawal). The result of an operator $\ope$ modifying a total preorder $C$ by
a formula $P$ is defined by the infix notation $C \ope(P)$. This is a new total
preorder whose base entails $P$ if $\ope$ is a revision operator. More
specifically, AGM revisions produce a base out of the minimal models of $P$ in
$C$:

\[
\min(C \ope(P),\top) = \min(C, P)
\]

\subsection{Iterated revisions}

Several iterated belief revision operators are considered. These can be all
expressed in terms of three of them: lexicographic revision, refinement, and
severe withdrawal. Intuitively, this is because each of these three includes a
basic operation that can be performed over an ordering: moving, splitting and
merging classes. The correspondence is not exact, as the lexicographic revision
perform both moving and splitting, but can be made to move a single class from
a position of the sequence to another.

These three operators are defined in this section. The others will be then
introduced in the next, and immediately proved to be reducible to these three.
This allows to concentrate on the computational aspects only on the three basic
ones.

\subsubsection{Lexicographic revision}

Lexicographic revision is one of the two earliest iterated belief revision
operator~\cite{spoh-88}. While its authors initially rejected it, later
research have reconsidered it~\cite{naya-94,libe-97-c,rott-03-a,naya-etal-03}.
The tenant of this operator can be summarized as: revising by $P$ means that
$P$ is true no matter of everything else. Technically, all models satisfying
$P$ are more plausible that every other one.

\begin{definition}

The lexicographic revision of a total preorder $C$ by a formula $P$ is defined
as the following total preorder, where $i$ and $j$ are respectively the indexes
of the minimal and maximal classes of $C$ containing models of $P$:

\[
C \lex(P)(k) = 
\left\{
\begin{array}{ll}
C(k+i) \cap \mod(P)		& \mbox{if } k \leq j-i \\
C(k-j+i-1) \backslash \mod(P)	& \mbox{otherwise}
\end{array}
\right.
\]

\end{definition}

Alternatively, a formula directly based on sequences can be taken as the
definition of lexicographic revision:

\begin{eqnarray*}
\lefteqn{ [C(0), \ldots, C(m)] \lex(P) = } \\
&& [
C(0) \cap \mod(P), \ldots, C(m) \cap \mod(P),
C(0) \backslash \mod(P), \ldots, C(m) \backslash \mod(P)
]
\end{eqnarray*}

This definition does not exactly coincide with the previous one because of some
empty classes, which means that the two produce equivalent total preorders. A
graphical representation of revising a total preorder by a formula $P$ is the
following one:

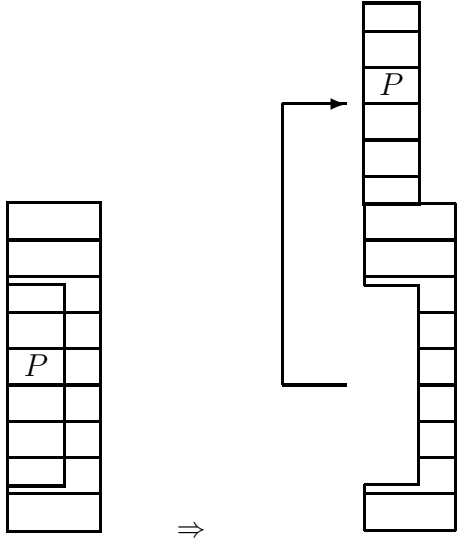
\ttytex{
\begin{tabular}{ccc}
\setlength{\unitlength}{5000sp}%
\begingroup\makeatletter\ifx\SetFigFont\undefined%
\gdef\SetFigFont#1#2#3#4#5{%
  \reset@font\fontsize{#1}{#2pt}%
  \fontfamily{#3}\fontseries{#4}\fontshape{#5}%
  \selectfont}%
\fi\endgroup%
\begin{picture}(516,1686)(4828,-4324)
\thicklines
{\color[rgb]{0,0,0}\put(4861,-2851){\line( 1, 0){450}}
}%
{\color[rgb]{0,0,0}\put(4861,-3031){\line( 1, 0){450}}
}%
{\color[rgb]{0,0,0}\put(4861,-3211){\line( 1, 0){450}}
}%
{\color[rgb]{0,0,0}\put(4861,-3391){\line( 1, 0){450}}
}%
{\color[rgb]{0,0,0}\put(4861,-3571){\line( 1, 0){450}}
}%
{\color[rgb]{0,0,0}\put(4861,-3751){\line( 1, 0){450}}
}%
{\color[rgb]{0,0,0}\put(4861,-3931){\line( 1, 0){450}}
}%
{\color[rgb]{0,0,0}\put(4861,-4111){\line( 1, 0){450}}
}%
{\color[rgb]{0,0,0}\put(4861,-4066){\framebox(270,990){}}
}%
{\color[rgb]{0,0,0}\put(4861,-4291){\framebox(450,1620){}}
}%
\put(4996,-3526){\makebox(0,0)[b]{\smash{{\SetFigFont{12}{24.0}{\rmdefault}{\mddefault}{\updefault}{\color[rgb]{0,0,0}$P$}%
}}}}
\end{picture}%
& ~ ~ $\Rightarrow$ ~ ~ &
\setlength{\unitlength}{5000sp}%
\begingroup\makeatletter\ifx\SetFigFont\undefined%
\gdef\SetFigFont#1#2#3#4#5{%
  \reset@font\fontsize{#1}{#2pt}%
  \fontfamily{#3}\fontseries{#4}\fontshape{#5}%
  \selectfont}%
\fi\endgroup%
\begin{picture}(921,2676)(4423,-4324)
\thicklines
{\color[rgb]{0,0,0}\put(4861,-2851){\line( 1, 0){450}}
}%
{\color[rgb]{0,0,0}\put(4861,-3031){\line( 1, 0){450}}
}%
{\color[rgb]{0,0,0}\put(4861,-4111){\line( 1, 0){450}}
}%
{\color[rgb]{0,0,0}\put(4861,-1816){\line( 1, 0){270}}
}%
{\color[rgb]{0,0,0}\put(4861,-1996){\line( 1, 0){270}}
}%
{\color[rgb]{0,0,0}\put(4861,-2176){\line( 1, 0){270}}
}%
{\color[rgb]{0,0,0}\put(4861,-2356){\line( 1, 0){270}}
}%
{\color[rgb]{0,0,0}\put(4861,-2536){\line( 1, 0){270}}
}%
{\color[rgb]{0,0,0}\put(5131,-3211){\line( 1, 0){180}}
}%
{\color[rgb]{0,0,0}\put(5131,-3391){\line( 1, 0){180}}
}%
{\color[rgb]{0,0,0}\put(5131,-3571){\line( 1, 0){180}}
}%
{\color[rgb]{0,0,0}\put(5131,-3751){\line( 1, 0){180}}
}%
{\color[rgb]{0,0,0}\put(5131,-3931){\line( 1, 0){180}}
}%
{\color[rgb]{0,0,0}\put(4861,-2671){\framebox(270,990){}}
}%
{\color[rgb]{0,0,0}\put(4771,-3571){\line(-1, 0){315}}
\put(4456,-3571){\line( 0, 1){1395}}
\put(4456,-2176){\vector( 1, 0){315}}
}%
{\color[rgb]{0,0,0}\put(5311,-4291){\line(-1, 0){450}}
\put(4861,-4291){\line( 0, 1){225}}
\put(4861,-4066){\line( 1, 0){270}}
\put(5131,-4066){\line( 0, 1){990}}
\put(5131,-3076){\line(-1, 0){270}}
\put(4861,-3076){\line( 0, 1){405}}
\put(4861,-2671){\line( 1, 0){450}}
\put(5311,-2671){\line( 0,-1){1620}}
}%
\put(4996,-2131){\makebox(0,0)[b]{\smash{{\SetFigFont{12}{24.0}{\rmdefault}{\mddefault}{\updefault}{\color[rgb]{0,0,0}$P$}%
}}}}
\end{picture}%
\end{tabular}
}{
                                +---+
                                |   |
           <-+                  |---|
             |                  | P |
 +------+    |                  |---|
 |      |    |                  |   |
 +------+    |                  +---+
 |+---+ |    |                 +------+
 +|---|-+    |      ===>       |      |
 || P | |   -+                 +------+
 +|---|-+                      |+---+ |
 |+---+ |                      +|   |-+
 +------+                      ||   | |
 |      |                      +|   |-+
 +------+                      |+---+ |
                               +------+
                               |      |
                               +------+
}

In words, the models of $P$ are ``cut out'' from the ordering and shifted
together to the top. Their relative ordering is not changed, but they are made
more plausible than every model of $\neg P$. By construction, $\min(C
\lex(P),\top)$ is equal to $\min(C,P)$, making this operator a revision.

\subsubsection{Refinement}

Contrary to lexicographic revision, refinement~\cite{papi-01,boot-etal-06} is
not a revision. It is still a basic form of belief revision in which belief in
a formula $P$ is strengthened, but never so much to contradict previous
information. Technically, the models of every class are split depending on
whether they satisfy $P$ or not. This way, two models are separated only if
they were previously considered equally plausible, and only if one satisfies
$P$ and the other does not.

\begin{definition}

The refinement of a total preorder $C$ by formula $P$ is the following total
preorder:

\[
C \refi(P)(k) = 
\left\{
\begin{array}{ll}
C(0) \cap \mod(P)	& \mbox{if $k=0$ and } C(0)\cap\mod(P) \not=\emptyset \\
C(0) \backslash \mod(P)	& \mbox{if $k=0$ and } C(0)\cap\mod(P) =\emptyset \\
C(k/2) \cap \mod(P)			& \mbox{if $k>0$ even}	\\
C(k/2) \backslash C \refi(P)(k-1)) 	& \mbox{if $k>0$ odd}
\end{array}
\right.
\]

\end{definition}

Alternatively, refinement can be defined directly on partitions:

\begin{eqnarray*}
\lefteqn{[C(0), \ldots, C(m)] \refi(P) = }		\\
&& [
C(0) \cap \mod(P),
C(0) \backslash \mod(P),
\ldots,
C(m) \cap \mod(P),
C(m) \backslash \mod(P)
]
\end{eqnarray*}

Some of these classes may be empty, and can therefore be removed respecting
preorder equivalence. Graphically, refining a total preorder $C$ by a formula
$P$ can be seen as follows:

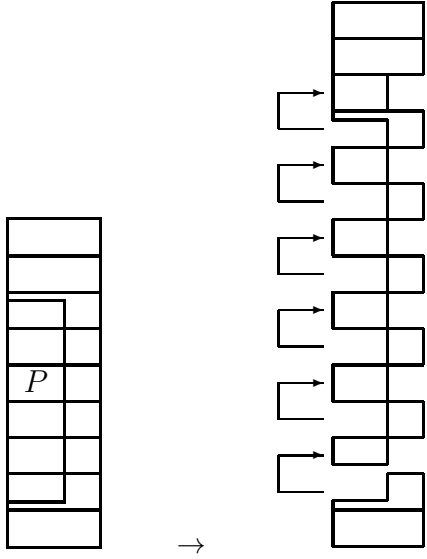
\ttytex{
\begin{tabular}{ccc}
\setlength{\unitlength}{5000sp}%
\begingroup\makeatletter\ifx\SetFigFont\undefined%
\gdef\SetFigFont#1#2#3#4#5{%
  \reset@font\fontsize{#1}{#2pt}%
  \fontfamily{#3}\fontseries{#4}\fontshape{#5}%
  \selectfont}%
\fi\endgroup%
\begin{picture}(516,1686)(4828,-4324)
\thicklines
{\color[rgb]{0,0,0}\put(4861,-2851){\line( 1, 0){450}}
}%
{\color[rgb]{0,0,0}\put(4861,-3031){\line( 1, 0){450}}
}%
{\color[rgb]{0,0,0}\put(4861,-3211){\line( 1, 0){450}}
}%
{\color[rgb]{0,0,0}\put(4861,-3391){\line( 1, 0){450}}
}%
{\color[rgb]{0,0,0}\put(4861,-3571){\line( 1, 0){450}}
}%
{\color[rgb]{0,0,0}\put(4861,-3751){\line( 1, 0){450}}
}%
{\color[rgb]{0,0,0}\put(4861,-3931){\line( 1, 0){450}}
}%
{\color[rgb]{0,0,0}\put(4861,-4111){\line( 1, 0){450}}
}%
{\color[rgb]{0,0,0}\put(4861,-4066){\framebox(270,990){}}
}%
{\color[rgb]{0,0,0}\put(4861,-4291){\framebox(450,1620){}}
}%
\put(4996,-3526){\makebox(0,0)[b]{\smash{{\SetFigFont{12}{24.0}{\rmdefault}{\mddefault}{\updefault}{\color[rgb]{0,0,0}$P$}%
}}}}
\end{picture}%
& ~ ~ $\rightarrow$ ~ ~ &
\setlength{\unitlength}{5000sp}%
\begingroup\makeatletter\ifx\SetFigFont\undefined%
\gdef\SetFigFont#1#2#3#4#5{%
  \reset@font\fontsize{#1}{#2pt}%
  \fontfamily{#3}\fontseries{#4}\fontshape{#5}%
  \selectfont}%
\fi\endgroup%
\begin{picture}(765,2766)(5389,-5224)
\thicklines
{\color[rgb]{0,0,0}\put(5671,-5011){\line( 1, 0){450}}
}%
\thinlines
{\color[rgb]{0,0,0}\put(5626,-3121){\line(-1, 0){225}}
\put(5401,-3121){\line( 0, 1){180}}
\put(5401,-2941){\vector( 1, 0){225}}
}%
{\color[rgb]{0,0,0}\put(5626,-3481){\line(-1, 0){225}}
\put(5401,-3481){\line( 0, 1){180}}
\put(5401,-3301){\vector( 1, 0){225}}
}%
{\color[rgb]{0,0,0}\put(5626,-3841){\line(-1, 0){225}}
\put(5401,-3841){\line( 0, 1){180}}
\put(5401,-3661){\vector( 1, 0){225}}
}%
{\color[rgb]{0,0,0}\put(5626,-4201){\line(-1, 0){225}}
\put(5401,-4201){\line( 0, 1){180}}
\put(5401,-4021){\vector( 1, 0){225}}
}%
{\color[rgb]{0,0,0}\put(5626,-4561){\line(-1, 0){225}}
\put(5401,-4561){\line( 0, 1){180}}
\put(5401,-4381){\vector( 1, 0){225}}
}%
{\color[rgb]{0,0,0}\put(5626,-4921){\line(-1, 0){225}}
\put(5401,-4921){\line( 0, 1){180}}
\put(5401,-4741){\vector( 1, 0){225}}
}%
\thicklines
{\color[rgb]{0,0,0}\put(5671,-2851){\line( 1, 0){450}}
}%
{\color[rgb]{0,0,0}\put(5671,-2671){\line( 1, 0){450}}
}%
{\color[rgb]{0,0,0}\put(5671,-3031){\line( 1, 0){270}}
}%
{\color[rgb]{0,0,0}\put(5941,-4651){\line( 1, 0){180}}
\put(6121,-4651){\line( 0, 1){180}}
\put(6121,-4471){\line(-1, 0){180}}
\put(5941,-4471){\line( 0, 1){180}}
\put(5941,-4291){\line( 1, 0){180}}
\put(6121,-4291){\line( 0, 1){180}}
\put(6121,-4111){\line(-1, 0){180}}
\put(5941,-4111){\line( 0, 1){180}}
\put(5941,-3931){\line( 1, 0){180}}
\put(6121,-3931){\line( 0, 1){180}}
\put(6121,-3751){\line(-1, 0){180}}
\put(5941,-3751){\line( 0, 1){180}}
\put(5941,-3571){\line( 1, 0){180}}
\put(6121,-3571){\line( 0, 1){180}}
\put(6121,-3391){\line(-1, 0){180}}
\put(5941,-3391){\line( 0, 1){180}}
\put(5941,-3211){\line( 1, 0){180}}
\put(6121,-3211){\line( 0, 1){180}}
\put(6121,-3031){\line(-1, 0){180}}
\put(5941,-3031){\line( 0, 1){180}}
\put(5941,-2851){\line( 1, 0){180}}
\put(6121,-2851){\line( 0, 1){360}}
\put(6121,-2491){\line(-1, 0){450}}
\put(5671,-2491){\line( 0,-1){585}}
\put(5671,-3076){\line( 1, 0){270}}
\put(5941,-3076){\line( 0,-1){135}}
\put(5941,-3211){\line(-1, 0){270}}
\put(5671,-3211){\line( 0,-1){180}}
\put(5671,-3391){\line( 1, 0){270}}
\put(5941,-3391){\line( 0,-1){180}}
\put(5941,-3571){\line(-1, 0){270}}
\put(5671,-3571){\line( 0,-1){180}}
\put(5671,-3751){\line( 1, 0){270}}
\put(5941,-3751){\line( 0,-1){180}}
\put(5941,-3931){\line(-1, 0){270}}
\put(5671,-3931){\line( 0,-1){180}}
\put(5671,-4111){\line( 1, 0){270}}
\put(5941,-4111){\line( 0,-1){180}}
\put(5941,-4291){\line(-1, 0){270}}
\put(5671,-4291){\line( 0,-1){180}}
\put(5671,-4471){\line( 1, 0){270}}
\put(5941,-4471){\line( 0,-1){180}}
\put(5941,-4651){\line(-1, 0){270}}
\put(5671,-4651){\line( 0,-1){135}}
\put(5671,-4786){\line( 1, 0){270}}
\put(5941,-4786){\line( 0, 1){135}}
}%
{\color[rgb]{0,0,0}\put(6121,-5191){\line(-1, 0){450}}
\put(5671,-5191){\line( 0, 1){225}}
\put(5671,-4966){\line( 1, 0){270}}
\put(5941,-4966){\line( 0, 1){135}}
\put(5941,-4831){\line( 1, 0){180}}
\put(6121,-4831){\line( 0,-1){360}}
}%
\end{picture}%
\end{tabular}
}{
+------+            +------+
|      |            |      |
+------+            +------+
|      |            |      |
+-+----+            +-+----+
| |    |              |    |
+-| -- +    ===>    +-+----+
| |  P |            | |
+-| -- +            +-+----+
| |    |              |    |
+-+----+            +-+----+
|      |            | |
+------+            +-+----+
                      |    |
                    +-+----+
                    | |
                    +-+----+
                    |      |
                    +------+
}

\subsubsection{Severe antiwithdrawal}

While this operator was defined~\cite{rott-pagn-99,ferm-rodr-98} as a form of
contraction, it is technically cleaner to use it in reverse, with the negated
formula. Removing a formula $\neg P$ is the same of creating the consistency
with $P$, but the second definition has been advocated has a most direct
formalization of the actual process of belief change~\cite{glai-00}.

In the specific case of severe antiwithdrawal, creating consistency with $P$ is
obtained by merging all classes of the ordering that are in the same class or
one of lower index with the minimal models of $P$. This is motivated by the
principle of equal-treated-equally when applied to the plausibility of models:
in order to make $P$ consistent some models of $P$ have to become minimal; but
the models of $\neg P$ that are in lower classes have the same plausibility or
greater, so they should not be excluded.

\begin{definition}

The severe anticontraction of the total preorder $C$ by formula $P$ is the
following total preorder, where $i$ is the minimal index such that $C(i) \cap
\mod(P) \not= \emptyset$:

\[
C\sev(P)(k) =
\left\{
\begin{array}{ll}
C(0) \cup \cdots \cup C(i)	&	\mbox{if } k=0 \\
C(k+i)				&	\mbox{if } k>0
\end{array}
\right.
\]

\end{definition}

Lexicographic antiwithdrawal can also be defined in terms of sequences. If $i$
is the lowest index such that $C(i) \cap \mod(P) \not= \emptyset$, then:

\begin{eqnarray*}
\lefteqn{[C(0),\ldots,C(i),C(i+1),\ldots,C(m)] \sev(P) = }		\\
&& [
C(0) \cup \cdots \cup C(i),
C(i+1),
\ldots
C(m)
]
\end{eqnarray*}

Graphically, severe antiwithdrawal merges all classes of index lower or equal
to the minimal class intersecting $\mod(P)$:

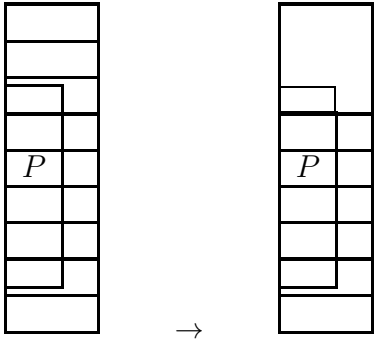
\ttytex{
\begin{tabular}{ccc}
\setlength{\unitlength}{5000sp}%
\begingroup\makeatletter\ifx\SetFigFont\undefined%
\gdef\SetFigFont#1#2#3#4#5{%
  \reset@font\fontsize{#1}{#2pt}%
  \fontfamily{#3}\fontseries{#4}\fontshape{#5}%
  \selectfont}%
\fi\endgroup%
\begin{picture}(516,1686)(4828,-4324)
\thicklines
{\color[rgb]{0,0,0}\put(4861,-2851){\line( 1, 0){450}}
}%
{\color[rgb]{0,0,0}\put(4861,-3031){\line( 1, 0){450}}
}%
{\color[rgb]{0,0,0}\put(4861,-3211){\line( 1, 0){450}}
}%
{\color[rgb]{0,0,0}\put(4861,-3391){\line( 1, 0){450}}
}%
{\color[rgb]{0,0,0}\put(4861,-3571){\line( 1, 0){450}}
}%
{\color[rgb]{0,0,0}\put(4861,-3751){\line( 1, 0){450}}
}%
{\color[rgb]{0,0,0}\put(4861,-3931){\line( 1, 0){450}}
}%
{\color[rgb]{0,0,0}\put(4861,-4111){\line( 1, 0){450}}
}%
{\color[rgb]{0,0,0}\put(4861,-4066){\framebox(270,990){}}
}%
{\color[rgb]{0,0,0}\put(4861,-4291){\framebox(450,1620){}}
}%
\put(4996,-3526){\makebox(0,0)[b]{\smash{{\SetFigFont{12}{24.0}{\rmdefault}{\mddefault}{\updefault}{\color[rgb]{0,0,0}$P$}%
}}}}
\end{picture}%
& ~ ~ $\rightarrow$ ~ ~ &
\setlength{\unitlength}{5000sp}%
\begingroup\makeatletter\ifx\SetFigFont\undefined%
\gdef\SetFigFont#1#2#3#4#5{%
  \reset@font\fontsize{#1}{#2pt}%
  \fontfamily{#3}\fontseries{#4}\fontshape{#5}%
  \selectfont}%
\fi\endgroup%
\begin{picture}(516,1686)(4828,-4324)
\thicklines
{\color[rgb]{0,0,0}\put(4861,-3211){\line( 1, 0){450}}
}%
{\color[rgb]{0,0,0}\put(4861,-3391){\line( 1, 0){450}}
}%
{\color[rgb]{0,0,0}\put(4861,-3571){\line( 1, 0){450}}
}%
{\color[rgb]{0,0,0}\put(4861,-3751){\line( 1, 0){450}}
}%
{\color[rgb]{0,0,0}\put(4861,-3931){\line( 1, 0){450}}
}%
{\color[rgb]{0,0,0}\put(4861,-4111){\line( 1, 0){450}}
}%
{\color[rgb]{0,0,0}\put(4861,-4066){\framebox(270,855){}}
}%
\thinlines
{\color[rgb]{0,0,0}\put(4861,-3076){\line( 1, 0){270}}
\put(5131,-3076){\line( 0,-1){135}}
}%
\thicklines
{\color[rgb]{0,0,0}\put(4861,-4291){\framebox(450,1620){}}
}%
\put(4996,-3526){\makebox(0,0)[b]{\smash{{\SetFigFont{12}{24.0}{\rmdefault}{\mddefault}{\updefault}{\color[rgb]{0,0,0}$P$}%
}}}}
\end{picture}%
\end{tabular}
}{
+------+            +------+
|      |            |      |
+------+            |      |
|      |            |      |
+--+---+            |  ....+
|  |   |            |  :   |
+--| - +    ===>    +--|---+
|  | P |            |  | P |
+--| - +            +--|---+
|  |   |            |  |   |
+--+---+            |--+---+
|      |            |      |
+------+            +------+
}

This way, the base of the revised preorder $\min(C \sev(P),\top)$ is guaranteed
to contain some models of $P$, which means that it has been made consistent
with $P$. At the same time, the relative plausibility of two models is never
reversed: a model that is more plausible that another according to $C$ is never
made less plausible than that according to $C \sev(P)$.

\section{Reductions}
\label{reductions}

Many belief change operators exist. Many of them are expressible in terms of
the three presented in the previous section: lexicographic revision, refinement
and severe antiwithdrawal. The reductions do not affect what is before or after
then replaced operator applications, which is not the case for the
transformations shown in the next section.

\begin{example}

The following sequence of revisions is used as a running example. The following
sections show how to make it into a sequence that only contains $\lex$,
$\refi$ and $\sev$.

\[
\emptyset \lex(y) \nat(\neg x) \res(x \wedge z) \rad(\neg z)
\]

\end{example}

\subsection{Natural revision}

This revision was first considered and discarded by Spohn~\cite{spoh-88}, and
later independently reintroduced by Boutilier~\cite{bout-96-a}. Among
revisions, it can be considered at further opposite to lexicographic revision,
in that a formula $P$ is made true by a minimal change to the ordering. This
amounts to making the minimal models of $P$ the new class zero of the ordering,
and changing nothing else.

Formally, if $\min(C,P)=C(i) \cap \mod(P)$ then:

\begin{eqnarray*}
\lefteqn{[C(0),\ldots,C(m)] \nat(P) = }			\\
&& [
C(i) \cap \mod(P),
C(0),
\ldots
C(i-1),
C(i) \backslash \mod(P),
C(i+1),
\ldots,
C(m)
]							\\
\end{eqnarray*}

Graphically, $\min(C,P)$ is ``cut out'' from the total preorder $C$ and moved
to the beginning of the sequence, making it the new class zero $C \nat(P)(0)$.
Since by definition $\min(C,P)$ is not empty, it holds $\min(C \nat(P),\top) =
C \nat(P)(0) = \min(C,P)$, meaning that it is an AGM revision operator.

\begin{theorem}

For every total preorder $C$ and formula $P$,
it holds $C \nat(P) \equiv C \lex(K)$ where $K=\form(\min(C,P))$.

\end{theorem}

\proof Let $K=\form(\min(C,P))$ and $i$ the index such that $\min(C,P)=C(i)
\cap \mod(P)$. By the properties of set difference, it holds $C(i) \backslash
\mod(P) = C(i) \backslash (C(i) \cap \mod(P)) = C(i) \backslash \mod(K)$. Since
classes do not share variable and $\mod(K) \subseteq C(i)$, it holds $C(j) =
C(j) \backslash \mod(K)$ for every $j \not= i$. Natural revision can therefore
be recast as:

\begin{eqnarray*}
\lefteqn{[C(0),\ldots,C(m)] \nat(P) = }			\\
&=& [
C(i) \cap \mod(M),
C(0),
\ldots
C(i-1),
C(i) \backslash \mod(P),
C(i+1),
\ldots,
C(m)
]							\\
&\equiv& [
\mod(K),						\\
&&~
C(0)	\backslash \mod(K),
\ldots
C(i-1)	\backslash \mod(K),
C(i)	\backslash \mod(K),				\\
&&~
C(i+1)	\backslash \mod(K),
\ldots,
C(m)	\backslash \mod(K)
]							\\
&\equiv& [
\mod(K), \emptyset, \ldots, \emptyset,			\\
&&~
C(0)	\backslash \mod(K),
\ldots,
C(m)	\backslash \mod(K)
]							\\
&\equiv& [
C(0)	\cap \mod(K),
C(1)	\cap \mod(K),
\ldots
C(m)	\cap \mod(K),					\\
&&~
C(0)	\backslash \mod(K),
\ldots
C(m)	\backslash \mod(K)
]							\\
\end{eqnarray*}

The equivalences are correct because: first, $C(i) \cap \mod(K) = \mod(K)$
since $\mod(K) \subseteq C(i)$; second, empty classes can be introduced at
every point of every ordering, and $C(j) \cap \mod(K)=\emptyset$ for every $j
\not= i$. The resulting total preorder is $C \lex(K)$.~\qed

This transformation does not just tell how to compute the propositional result
of natural revision, that is, the base $\form(C \nat(P)(0))$ of the revised
ordering. To the contrary, it requires it, as $\mod(K) = \min(C,P) = C
\nat(P)(0)$. After $K$ has been calculated, $\lex(K)$ produces the same exact
preorder as $\nat(P)$ when applied to the same preorder, not just two preorders
having the same base. This means that all subsequent revisions are unaffected
by the replacement. In other words, for every initial preorder $C$ and every
sequence of previous and future belief changes, it holds:

\begin{eqnarray*}
\lefteqn{
C
\ope_1(P_1) \ldots \ope_{n-1}(P_{n-1})
\nat(P) 
\ope_{n+1}(P_{n+1}) \ldots \ope_{n'}(P_{n'})
\equiv }						\\
&&
C
\ope_1(P_1) \ldots \ope_{n-1}(P_{n-1})
\lex(K) 
\ope_{n+1}(P_{n+1}) \ldots \ope_{n'}(P_{n'})
\end{eqnarray*}

The other reductions in this section have all this property, that an operator
application in whichever position of a sequence can be replaced without
affecting the final ordering. Natural revision requires the minimal models of
$P$ to be calculated, some other operators do not. Natural revision is replaced
by a single lexicographic revision, the others may require some lexicographic
revisions, refinements and severe antiwithdrawals.

\begin{example}

The second operation in the running example is a natural revision.

\[
\emptyset \lex(y) \nat(\neg x) \res(x \wedge z) \rad(\neg z)
\]

Since $\emptyset lex(y)$ is $[\mod(y),\mod(\neg y)]$, and class zero of this
ordering contains models of $\neg x$, then $\min(\emptyset,\neg x)=\mod(\neg x
\wedge y)$. As a result, the sequence can be simplified into:

\[
\emptyset \lex(y) \lex(\neg x \wedge y) \res(x \wedge z) \rad(\neg z)
\]

\end{example}

Some other operators are reduced to natural revision, which can in turn be
reduced to lexicographic revision. For example, restrained revision is a
refinement followed by natural revision (or vice versa). The above theorem
shows that it can be further reformulated as a refinement and a lexicographic
revision.

\subsection{Restrained revision}

Restrained revision~\cite{boot-meye-06} can be seen as a minimal modification
of refinement to turn it into a form of revision. Indeed, refining a total
preorder by a formula $P$ does not generally makes $P$ entailed by the refined
total preorder. This is indeed the case only if $\min(C,\top)$ contains some
models of $P$.


Restrained revision can be seen as an intermediate form of revision: while
natural revision changes the preorder in a minimal way to make the revising
formula entailed and lexicographic revision makes the formula to be preferred
in all possible cases, restrained revision makes it to be preferred only when
this is consistent with previous beliefs, and makes it entailed by a minimal
change in the ordering.

Restrained revision is defined as follows, where $\min(C,P)=C(i) \cap \mod(P)$.

\begin{eqnarray*}
\lefteqn{[C(0),\ldots,C(m)] \res(P) = }			\\
&=& [
C(i)	\cap \mod(P),					\\
&&~
C(0)	\cap \mod(P),
C(0)	\backslash \mod(P),
\ldots,
C(i-1)	\cap \mod(P),
C(i-1)	\backslash \mod(P),				\\
&&~
C(i)	\backslash \mod(P),				\\
&&~
C(i+1)	\cap \mod(P),
C(i+1)	\backslash \mod(P),
\ldots
C(m)	\cap \mod(P),
C(m)	\backslash \mod(P)
]
\end{eqnarray*}

The following quite obvious theorem is proved only for the sake of
completeness, its statement being almost a direct consequence of the
definition.

\begin{theorem}

For every total preorder $C$ and formula $P$, it holds
$C \mathrm{res}(P) \equiv C \refi(P) \nat(P)$

\end{theorem}

\proof Let $\min(C,P)=C(i) \cap \mod(P)$. The ordering $C \refi(P) \nat(P)$ is:

\begin{eqnarray*}
\lefteqn{[C(0),\ldots,C(m)] \refi(P) \nat(P) = }		\\
&=& [
C(0)	\cap \mod(P),
C(0)	\backslash \mod(P),
\ldots
C(m)	\cap \mod(P),
C(m)	\backslash \mod(P)
] \nat(P)							\\
&=& [
C(i)	\cap \mod(P),						\\
&&~
C(0)	\cap \mod(P),
C(0)	\backslash \mod(P),
\ldots
C(i-1)	\cap \mod(P),
C(i-1)	\backslash \mod(P),					\\
&&~
\emptyset,
C(i)	\backslash \mod(P),					\\
&&~
C(i+1)	\cap \mod(P),
C(i+1)	\backslash \mod(P),				
\ldots
C(m)	\cap \mod(P),
C(m)	\backslash \mod(P)
]								\\
&=& [
\min(C \refi(P),P),
C(0)	\cap \mod(P),
C(0)	\backslash \mod(P),
\ldots
\emptyset,							\\
&&~
C(i)	\backslash \mod(P),
\ldots
C(m)	\cap \mod(P),
C(m)	\backslash \mod(P)
]								\\
\end{eqnarray*}

By assumption, the minimal class of $C$ containing models of $P$ is $C(i)$. As
a result, the minimal class of $C \refi(P)$ containing models of $P$ is $C(i)
\cap \mod(P)$. As a result, $\min(C \refi(P),P)=\min(C,P)$. The total preorder
above is therefore equivalent to $C \refi(P)$, since empty classes do not
affect equivalence.~\qed

This reduction is applied to the running example.

\begin{example}

Restrained revision can be replaced by a refinement followed by natural
revision.

\[
\emptyset \lex(y) \lex(\neg x \wedge y) \res(x \wedge z) \rad(\neg z)
\]

This operation results into the following sequence:

\[
\emptyset
\lex(y)
\lex(\neg x \wedge y)
\refi(x \wedge z) \nat(x \wedge z)
\rad(\neg z)
\]

The resulting natural revision can be then replaced by lexicographic revision.
It can be seen that the minimal models of $x \wedge z$ in the ordering just
before the natural revision are these of $x \wedge y \wedge z$:

\[
\emptyset
\lex(y)
\lex(\neg x \wedge y)
\refi(x \wedge z) \lex(x \wedge y \wedge z)
\rad(\neg z)
\]

\end{example}

\subsection{Very radical revision}

Irrevocable revision~\cite{sege-98} formalizes hypothetical reasoning by
excluding from consideration all models that do not satisfy the assumption.
Formally, these models are made inaccessible to revision, which cannot
therefore recover them (hence the name). The scope of this article is limited
to revisions that consider all models. While irrevocable revision excludes some
model, the very radical revision variant by Rott~\cite{rott-09} does not.
Formally, it is defined as follows.

\begin{eqnarray*}
\lefteqn{[C(0),\ldots,C(m)] \rad(P) = }			\\
&=& [
C(0) \cap \mod(P),
\ldots
C(m) \cap \mod(P),
(C(0) \cup \cdots \cup C(m)) \backslash \mod(P)
]
\end{eqnarray*}

The original definition has the first part $C(0) \cap \mod(P),\ldots,C(m) \cap
\mod(P)$ only for the classes that intersect $\mod(P)$. The difference is
inessential since the other classes are empty and empty classes are irrelevant.

Very radical revision can be expressed in terms of a sequence of a
lexicographic revision, a severe antiwithdrawal and a second lexicographic
revision. Intuitively, this is because very radical revision merges the classes
not satisfying $P$, which is equivalent to make them minimal by a lexicographic
revision by $\neg P$ and then by a severe antiwithdrawal by $P$; a further
lexicographic revision is needed to restore the correct ordering.

\begin{theorem}

For every total preorder $C$ and formula $P$, it holds
$C \rad(P) \equiv C \lex(\neg P) \sev(P) \lex(P)$.

\end{theorem}

\proof By definition, $C \lex(\neg P) \sev(P) \lex(P)$ is the following
total preorder:

\begin{eqnarray*}
\lefteqn{[C(0),\ldots,C(m)] \lex(\neg P) \sev(P) \lex(P) = }	\\
&=& [
C(0)	\cap \mod(\neg P),
\ldots,
C(m)	\cap \mod(\neg P),					\\
&&~
C(0)	\backslash \mod(\neg P),
\ldots,
C(m)	\backslash \mod(\neg P)
] \sev(P) \lex(P)						\\
&=& [
C(0)	\backslash \mod(P),
\ldots,
C(m)	\backslash \mod(P),					\\
&&~
C(0)	\cap \mod(P),
\ldots,
C(m)	\cap \mod(P)
] \sev(P) \lex(P)						\\
&=& [
C(0)	\backslash \mod(P)
\cup \cdots \cup
C(m)	\backslash \mod(P)
\cup
C(0)	\cap \mod(P),						\\
&&~
C(1)	\cap \mod(P),
\ldots,
C(m)	\cap \mod(P)
] \lex(P)							\\
&=& [
(C(0) \cup \cdots \cup C(m))	\backslash \mod(P)
\cup C(0)	\cap \mod(P),					\\
&&~
C(1)	\cap \mod(P),					
\ldots,
C(m)	\cap \mod(P)
] \lex(P)							\\
&=& [
((C(0) \cup \cdots \cup C(m))	\backslash \mod(P)
\cup C(0)	\cap \mod(P)) \cap \mod(P),			\\
&&~
C(1)	\cap \mod(P) \cap \mod(P),
\ldots,
C(m)	\cap \mod(P) \cap \mod(P),				\\
&&~
((C(0) \cup \cdots \cup C(m))	\backslash \mod(P)
\cup C(0)	\cap \mod(P)) \backslash \mod(P),		\\
&&~
C(1)	\cap \mod(P) \backslash \mod(P),
\ldots,
C(m)	\cap \mod(P) \backslash \mod(P)
]								\\
&=& [
C(0)	\cap \mod(P),						\\
&&~
C(1)	\cap \mod(P),
\ldots,
C(m)	\cap \mod(P),						\\
&&~
(C(0) \cup \cdots \cup C(m)	\backslash \mod(P),
\emptyset,
\ldots,
\emptyset
]
\end{eqnarray*}

Apart from the empty classes, this total preorder is $C \rad(P)$.~\qed

The reduction is applied to the running example.

\begin{example}

The previous replacements turned the sequence of revisions of the running
example into the following.

\[
\emptyset
\lex(y)
\lex(\neg x \wedge y)
\refi(x \wedge z) \lex(x \wedge y \wedge z)
\rad(\neg z)
\]

The last revision of the sequence $\rad(\neg z)$ is replaced by $\lex(\neg \neg
z) \sev(\neg z) \lex(\neg z)$:

\[
\emptyset
\lex(y)
\lex(\neg x \wedge y)
\refi(x \wedge z) \lex(x \wedge y \wedge z)
\lex(z) \sev(\neg z) \lex(\neg z)
\]

This sequence contains only lexicographic revisions, a refinement and a severe
antiwithdrawal.

\end{example}

\subsection{Severe revisions}

The Levi identity~\cite{hans-11} allows constructing a revision operator from a
contraction. This can be applied to severe withdrawal, leading to the
definition of severe revision. However, the Levy identity only specifies the
base of the revised ordering, the set of its minimal models. The rest of the
ordering can be obtained in at least three different ways, leading to different
revision operators~\cite{rott-09}.

The first definition is called just ``severe revision''. Since the symbol
$\sev$ is already taken for severe antiwithdrawal, $\sevr$ is used for this
revision.

\begin{definition}

If $\min(C,P)=C(i) \cap \mod(P)$, the severe revision $\sevr$ revises the
total preorder $C$ by formula $P$ as follows.

\begin{eqnarray*}
\lefteqn{[C(0),\ldots,C(m)] \sevr(P) = } \\
&=& [
  C(i) \cap \mod(P),
  (C(0) \cup \cdots \cup C(i)) \backslash \mod(P),
  C(i+1),\ldots,C(m)
]
\end{eqnarray*}

\end{definition}

This operator can be shown to be reducible to a severe antiwithdrawal followed
by a natural revision, the latter being reducible to lexicographic revision as
proved above.

\begin{theorem}

For every total preorder $C$ and formula $P$, it holds
$C \sevr(P) \equiv \sev(P) \nat(P)$.

\end{theorem}

\proof Let $C=[C(0),\ldots,C(m)]$ and $i$ be the minimal index such that $C(i)
\cap \mod(P) \not=\emptyset$. Revising $C$ by $\sev(P)$ and then $\nat(P)$
produces:

\begin{eqnarray*}
\lefteqn{[C(0),\ldots,C(m)] \sev(P) \nat(P) = }			\\	
&& [
C(0) \cup \cdots \cup C(i),
C(i+1),
\ldots
C(m)
] \nat(P)
\end{eqnarray*}

Since class zero of this ordering is $C(0) \cup \cdots \cup C(i)$ and $C(i)$
intersects $\mod(P)$, it follows that the minimal index of a class of this
ordering interesting $\mod(P)$ is zero. As a result, natural revision
produces:

\begin{eqnarray*}
\lefteqn{
[
C(0) \cup \cdots \cup C(i),
C(i+1),
\ldots
C(m)
] \nat(P)
= }							\\
&=& [
(C(0) \cup \cdots \cup C(i)) \cap \mod(P),
(C(0) \cup \cdots \cup C(i)) \backslash \mod(P),	\\
&&~
C(i+1),
\ldots
C(m)
] \nat(P)						\\
&=& [
C(i) \cap \mod(P),
(C(0) \cup \cdots \cup C(i)) \backslash \mod(P),
C(i+1),
\ldots
C(m)
] \nat(P)
\end{eqnarray*}

The last step follows from $C(i)$ being the minimal-index class of $C$
intersecting $\mod(P)$, which implies $C(j) \cap \mod(P) = \emptyset$ for
every $j<i$. The last total preorder is $C \sevr(P)$.~\qed

Moderate severe revision mixes a severe withdrawal with the changes
lexicographic revision makes to a preorder. It will indeed be proved to be
equivalent as a sequence of a severe antiwithdrawal and a lexicographic
revision.

\begin{definition}

If $\min(C,P)=C(i) \cap \mod(P)$, the moderate severe revision $\msev$ revises
the total preorder $C$ by formula $P$ as follows.

\begin{eqnarray*}
\lefteqn{[C(0),\ldots,C(m)] \msev(P) = }			\\
&=& [
  C(0) \cap \mod(P),\ldots,C(m) \cap \mod(P),			\\
&&~
  (C(0) \cup \cdots \cup C(i)) \backslash \mod(P),		
  C(i+1) \backslash \mod(P),\ldots,C(m) \backslash \mod(P)
]
\end{eqnarray*}

\end{definition}

Moderate severe revision can be proved to be equivalent to a severe
antiwithdrawal followed by a lexicographic revision.

\begin{theorem}

For every total preorder $C$ and formula $P$, it holds
$C \msev(P) \equiv \sev(P) \lex(P)$.

\end{theorem}

\proof Let $C=[C(0),\ldots,C(m)]$ and $i$ be the index such that
$\min(C,P)=C(i) \cap \mod(P)$. Revising $C$ by $\sev(P)$ and then $\lex(P)$
produces:

\begin{eqnarray*}
\lefteqn{[C(0),\ldots,C(m)] \sev(P) \lex(P) = }			\\	
&=& [
C(0) \cup \cdots \cup C(i),
C(i+1),
\ldots
C(m)
] \lex(P)							\\
&=& [
(C(0) \cup \cdots \cup C(i)) \cap \mod(P),
C(i+1) \cap \mod(P),
\ldots
C(m) \cap \mod(P),						\\
&&~
(C(0) \cup \cdots \cup C(i)) \backslash \mod(P),
C(i+1) \backslash \mod(P),
\ldots
C(m) \backslash \mod(P)
] 								\\
&=& [
C(i) \cap \mod(P),
C(i+1) \cap \mod(P),
\ldots
C(m) \cap \mod(P),						\\
&&~
(C(0) \cup \cdots \cup C(i)) \backslash \mod(P),
C(i+1) \backslash \mod(P),
\ldots
C(m) \backslash \mod(P)
] 								\\
&\equiv& [
C(0) \cap \mod(P),
\ldots
C(i-1) \cap \mod(P),						\\
&&~
C(i) \cap \mod(P),
C(i+1) \cap \mod(P),
\ldots
C(m) \cap \mod(P),						\\
&&~
(C(0) \cup \cdots \cup C(i)) \backslash \mod(P),
C(i+1) \backslash \mod(P),
\ldots
C(m) \backslash \mod(P)
] 								\\
\end{eqnarray*}

Equivalence $(C(0) \cup \cdots \cup C(i)) \cap \mod(P) = C(i) \cap \mod(P)$
holds because $C(i)$ is by assumption the lowest-index class intersecting
$\mod(P)$. For the same reason, $C(0) \cap \mod(P),\ldots,C(i-1) \cap
\mod(P)${\plural} are all empty; therefore, their introduction leads to an
equivalent preorder. The preorder obtained this way is $C \msev(P)$.~\qed

The last variant of severe revision is plain severe revision. Let
$\min(C,P)=C(i) \cap \mod(P)$ and $j$ be the minimal index such that $i<j$ and
$C(j) \not= \emptyset$ if any, otherwise $j=i+1$. Plain severe revision is
defined as follows.

\begin{definition}

If $\min(C,P)=C(i) \cap \mod(P)$, the plain severe revision $\psev$ revises
the total preorder $C$ by formula $P$ as follows.

\begin{eqnarray*}
\lefteqn{[C(0),\ldots,C(m)] \psev(P) = }			\\
&=& [
C(i) \cap \mod(P), C(0) \cup \cdots \cup (C(i) \backslash \mod(P)) \cup C(j),
C(j+1),\ldots,C(m)
]
\end{eqnarray*}

\end{definition}

Plain severe revision can be reformulated in terms of severe antiwithdrawal
and lexicographic revision.

\begin{theorem}

For every total preorder $C$ and formula $P$, it holds
$C \psev(P) \equiv C \sev(\neg K') \lex(K)$
where 
$K = \form(\min(C,P))$
and $K' = \form(\min(C \sev(P), \top)$.

\end{theorem}

\begin{eqnarray*}
\lefteqn{[C(0),\ldots,C(i),C(i+1),\ldots,C(m)] \sev(P) = }		\\
&& [
C(0) \cup \cdots \cup C(i),
C(i+1),
\ldots,
C(m)
]
\end{eqnarray*}

\proof Let $i$ and $j$ be the indexes as in the definition of plain severe
revision. The models of $K'$ are the first non-empty class of the ordering
$C \sev(P)$, where $C=[C(0),\ldots,C(m)]$:

\begin{eqnarray*}
\lefteqn{[C(0),\ldots,C(i),C(i+1),\ldots,C(m)] \sev(P) = }		\\
&& [
C(0) \cup \cdots \cup C(i),
C(i+1),
\ldots,
C(m)
]
\end{eqnarray*}

By definition $i$ is such that $C(i) \cap \mod(P) \not=\emptyset$. As a result,
$C(0) \cup \cdots \cup C(i)$ is not empty and therefore defines the set of
models of $K'$. The models of $\neg K'$ are the other ones:

\[
\mod(\neg K') = C(i+1) \cup \cdots \cup C(m)
\]

The ordering $C \sev(\neg K') \lex(K)$ can now be determined. By construction,
none of the classes $C(0),\ldots,C(i)$ intersect $\mod(\neg K')$. The next
class $C(i+1)$ may, but only if it is not empty. In particular, the lowest
index class intersecting $\mod(K')$ is $C(j)$.

\begin{eqnarray*}
\lefteqn{[C(0),\ldots,C(m)] \sev(\neg K') \lex(K) = }		\\
&=& [
C(0) \cup \cdots \cup C(i) \cup \cdots \cup C(j),
C(j+1),
\ldots,
C(m)
] \lex(K)							\\
&=& [
C(i) \cap \mod(P),
C(0) \cup \cdots (C(i) \backslash \mod(P)) \cup \cdots \cup C(j),
C(j+1),
\ldots,
C(m)
]								\\
&\equiv& [
C(i) \cap \mod(P),
C(0) \cup \cdots (C(i) \backslash \mod(P)) \cup C(j),
C(j+1),
\ldots,
C(m)
]								\\
\end{eqnarray*}

The last simplification can be done because all classes between $C(i)$ and
$C(j)$ are by definition empty. What results coincides with the definition of
$C \psev(P)$.~\qed

The following theorem shows that plain severe revision is not able to increase
the number of levels over two. It also links it with full meet revision, to be
defined in the next section.

\begin{theorem}
\label{twoclasses}

If $C$ has at most two non-empty classes, then
{}$	C \psev(P) \equiv
{}	[C(i) \cap \mod(P),\mod(\top) \backslash (C(i) \cup \mod(P))]$
holds for every formula $P$, where $\min(C,P)=C(i) \cap \mod(P)$.

\end{theorem}

\proof Since $C$ has at most two non-empty classes, and removing empty classes
produces an equivalent preorder, it can be assumed $C = [C(0),C(1)]$ where
$C(0) \not= \emptyset$ while $C(1)$ may be empty. The definition of the plain
severe revision depends on the minimal class $i$ intersecting $\mod(P)$ and the
minimal non-empty class of index greater than $i$. Since $C$ has only two
classes, $i$ can only be $0$ or $1$. In the first case, $j$ can only be 1,
regardless of whether $C(1)$ is empty or not. As a result:

\begin{eqnarray*}
\lefteqn{[C(0),C(1)] \psev(P) = }			\\
&=& [
C(0) \cap \mod(P), (C(0) \backslash \mod(P)) \cup C(1)
]							\\
&=& [
C(0) \cap \mod(P), \mod(\top) \backslash (C(0) \cap \mod(P))
]							\\
&=& [
C(0) \cap \mod(P), \mod(\top) \backslash (C(i) \cap \mod(P))
]
\end{eqnarray*}

If $C(0) \cap \mod(P) = \emptyset$, then $i=1$ and $j=i+1=2$ since no class of
index greater than $i$ exists, therefore none is different from the empty set.
As a result:

\begin{eqnarray*}
\lefteqn{[C(0),C(1)] \psev(P) = }			\\
&=& [
C(1) \cap \mod(P), C(0) \cup (C(1) \backslash \mod(P)) \cup C(2)
]							\\
&=& [
C(1) \cap \mod(P), \mod(\top) \backslash (C(1) \cap \mod(P))
]							\\
&=& [
C(1) \cap \mod(P), \mod(\top) \backslash (C(i) \cap \mod(P))
]
\end{eqnarray*}

Since $C$ has only two classes, $C(2)$ is empty and can be removed. Since
{} $C(0) \cap \mod(P) = \emptyset$, the set
{} $C(0) \cup (C(1) \backslash \mod(P))$ is equal to
{} $(C(0) \cup C(1)) \backslash \mod(P)$, in turn equal to
{} $\mod(\top) \backslash (C(1) \cap \mod(P))$.~\qed

\subsection{Full meet revision}

Full meet revision was initially defined in the one-step revision
case~\cite{alch-gard-maki-85,gard-88}. In particular, it was the result of
disjoining all possible ways of minimally revising a propositional theory,
formalizing both the assumption of minimal change and that of a complete lack
of knowledge about the plausibility of the various choices. The initial
plausibility ordering is not used other than for its set of minimal models. The
resulting ordering only distinguish models in two classes: the base and the
others.

\begin{definition}

The full meet revision $\full$ revises an ordering $C$ by a formula $P$ as
follows, where $\min(C,P) = C(i) \cap \mod(P)$.

\begin{eqnarray*}
\lefteqn{[C(0),\ldots,C(m)] \full(P) = } \\
&=& [
C(i) \cap \mod(P),
\mod(\top) \backslash (C(i) \cap \mod(P))
]
\end{eqnarray*}

\end{definition}

Due to its simplicity, full meet revision can be expressed in a number of ways
in terms of the other operators. For example, it is equivalent to a sequence
made of a lexicographic revision followed by a severe antiwithdrawal and
another lexicographic revision.

\begin{theorem}

For every total preorder $C$ and formula $P$, it holds
$C \full(P) \equiv C \lex(\neg K) \sev(K) \lex(K)$, where 
$K = \form(\min(C,P))$.

\end{theorem}

\proof Let $C=[C(0),\ldots,C(m)]$ and $i$ be the lowest index such that $C(i)
\cap \mod(P) \not= \emptyset$. By definition, $\mod(K)=C(i) \cap \mod(P)$.
Since $\mod(K) \subseteq C(i)$, it holds $\mod(K) \cap C(j) = \emptyset$ for
every $j \not= i$.

\begin{eqnarray*}
\lefteqn{[C(0),\ldots,C(m)] \lex(\neg K) \sev(K) \lex(K) = }	\\
&=& [
C(0) \cap \mod(\neg K),\ldots,C(m) \cap \mod(\neg K),		\\
&&~
C(0) \backslash \mod(\neg K),\ldots,C(m) \backslash \mod(\neg K)
] \sev(K) \lex(K)						\\
&=& [
C(0) \backslash \mod(K),\ldots,C(m) \backslash \mod(K),		\\
&&~
C(0) \cap \mod(K),\ldots,C(m) \cap \mod(K)
] \sev(K) \lex(K)						\\
&=& [
C(0) \backslash \mod(K),\ldots,C(m) \backslash \mod(K),
\mod(K)
] \sev(K) \lex(K)						\\
&& \mbox{since $\mod(K) \cap C(j)=\emptyset$ if $j \not= i$
         and $\mod(K) \cap C(i) = \mod(K)$}			\\
&=& [
\mod(\top),
\emptyset
] \lex(K)							\\
&\equiv& [
\mod(\top)
] \lex(K)							\\
&=& [
\mod(K),
\mod(\neg K)
]
\end{eqnarray*}

Since $\mod(K)=C(i) \cap \mod(P)$, the final total preorder is $C
\full(P)$.~\qed

An alternative reduction is $C \full(P) = C \lex(\neg \form(\{I\}))
\sev(\form(\{I\})) \lex(K)$, where $I$ is an arbitrary propositional
interpretation. Indeed, the proof relies on $C \lex(\neg F) \sev(F) =
[\mod(\top)]$, which holds for every formula $F$ such that $\mod(F)$ is
contained in a single class of $C$. This is the case for $\min(C,P)$, but also
for every formula having only one model.

Yet another reduction is $C \full(P) = \emptyset \lex(K)$ dove $K =
\form(\min(C,P))$. This is however not used because, contrarily to the other
reductions it affects the previous sequence of revisions. For example, $C
\nat(N) \full(P)$ is turned into $\emptyset \lex(K)$, therefore making the
initial natural revision disappear.

The previous revisions are instead preserved by the reduction $C \full(P)
\equiv C \rad(K)$ where $K=\form(\min(C,P))$, which however requires the
calculation of the minimal models of $P$ in the total preorder $C$ before being
applied. The very radical revision can be then expressed in terms of two
lexicographic revisions and a severe antiwithdrawal.

\

Starting from an empty ordering, full meet revision and plain severe revision
behave in exactly the same way. Formally, for every sequence of formulae
$P_1,\ldots,P_n$ it holds:

\[
\emptyset \full(P_1) \ldots \full(P_n)
\equiv
\emptyset \psev(P_1) \ldots \psev(P_n)
\]

This means that a sequence of mixed plain severe and full meet revisions can be
turned into one containing only one type of revisions. This fact is a
consequence of how they change an ordering comprising one or two classes: they
both produce an ordering containing the class $\min(C,P)$ and the class
containing all other models. For full meet revision, this is the definition and
holds in all cases. For plain severe revision this is proved in
Theorem~\ref{twoclasses}.

\section{The algorithm}
\label{method}

The previous section shows that every considered belief change operator can be
reduced to a sequence of lexicographic revisions, refinements and severe
antiwithdrawals. As a result, every sequence of operators can be turned into
one made of these three only. This section presents an algorithm for computing
the base of the ordering at every time step of such a sequence.

This is done by first proving that refinements and severe antiwithdrawals can
be removed by suitably modifying the sequence. This is done differently than
the reductions in the previous section, which only modify the sequence locally:
nothing is changed before or after the operator that is replaced. Removing
refinements instead requires introducing lexicographic revisions in other
points of the sequence, and removing severe antiwithdrawals requires changing
the previous lexicographic revisions.

The algorithm that computes the bases of a sequence of lexicographic revisions
is then modified to work on the original sequence. The detour to the sequence
of lexicographic revisions is necessary to prove that the final algorithm
works. In particular, it is shown to do the same as the original algorithm on
the simplified sequence.

All sequences are assumed to start with the empty ordering $\emptyset$. Every
other ordering $C=[C(0),\ldots,C(m)]$ is the result of a sequence of
lexicographic revision applied to the empty ordering: $\emptyset
\lex(\form(C(m))) \ldots \lex(\form(C(0)))$.

\subsection{Simplification}

A sequence of lexicographic revisions ending in either a refinement or a severe
antiwithdrawal can be turned into a sequence of lexicographic revisions that
has exactly the same final ordering when applied to the same original ordering.
As a result, a sequence containing every of these three operators can be
scanned from the beginning until the first operator that is not a lexicographic
revising is found. The initial part of the sequence is then turned into a
sequence containing only lexicographic revisions, and the process restarted.

Removal of the refinements is done thanks to the following theorems, which
proves that a refinement can be moved at the beginning of a sequence of
lexicographic revisions, and then turned into a lexicographic revision itself.

\begin{theorem}

For every ordering $C$ and two formulae $R$ and $L$, it holds $C \lex(L)
\refi(R) = C \refi(R) \lex(L)$.

\end{theorem}

\proof According to the definitions of $\lex$ and $\refi$, the ordering $C
\lex(L) \refi(R)$ is:

\begin{eqnarray*}
\lefteqn{[C(0), \ldots, C(m)] \lex(L) \refi(R) = } 		\\
&=&
[
C(0) \cap \mod(L), \ldots, C(m) \cap \mod(L),
C(0) \backslash \mod(L), \ldots, C(m) \backslash \mod(L)
] \refi(R)							\\
&=&
[
C(0) \cap \mod(L) \cap \mod(R),
C(0) \cap \mod(L) \backslash \mod(R)				\\
&&~
\ldots
C(m) \cap \mod(L) \cap \mod(R),
C(m) \cap \mod(L) \backslash \mod(R),				\\
&&~
C(0) \backslash \mod(L) \cap \mod(R),
C(0) \backslash \mod(L) \backslash \mod(R)			\\
&&~
\ldots
C(m) \backslash \mod(L) \cap \mod(R)
C(m) \backslash \mod(L) \backslash \mod(R)
]
\end{eqnarray*}

The ordering resulting from the opposite application $C \refi(R) \lex(L)$ is:

\begin{eqnarray*}
\lefteqn{[C(0), \ldots, C(m)] \refi(R) \lex(L) = } 		\\
&=&
[
C(0) \cap \mod(R),
C(0) \backslash \mod(R)
\ldots
C(m) \cap \mod(R),
C(m) \backslash \mod(R)
] \lex(L) \\
&=&
[
C(0) \cap \mod(R) \cap \mod(L),
C(0) \backslash \mod(R) \cap \mod(L)				\\
&&~
\ldots
C(m) \cap \mod(R) \cap \mod(L),
C(m) \backslash \mod(R) \cap \mod(L),				\\
&&~
C(0) \cap \mod(R) \backslash \mod(L),
C(0) \backslash \mod(R) \backslash \mod(L)			\\
&&~
\ldots
C(m) \cap \mod(R) \backslash \mod(L),
C(m) \backslash \mod(R) \backslash \mod(L)
]
\end{eqnarray*}

Since $\cap$ and $\backslash$ commute, these two sequences are the same.~\qed

This proves that $\emptyset \lex(L_1) \ldots \lex(L_{n-1}) \lex(L_n) \refi(R)$
is equal to $\emptyset \lex(L_1) \ldots \lex(L_{n-1}) \refi(R) \lex(L_n)$.
Iteratively applying commutativity produces $\emptyset \refi(R) \lex(L_1)
\ldots \lex(L_n)$. Since $\emptyset=[\mod(T)]$, by definition $\emptyset
\refi(R) = [\mod(R),\mod(\neg R)]$, and this is also the total preorder
$\emptyset \lex(R)$. As a result, the whole sequence is equivalent to
$\emptyset \lex(R) \lex(L_1) \ldots \lex(L_n)$.

\begin{corollary}
\label{ref-lex}

For every formulae $L_1,\ldots,L_n$ and $R$, it holds:

\[
\emptyset \lex(L_1) \ldots \lex(L_{n-1}) \lex(L_n) \refi(R)
\equiv
\emptyset \lex(R) \lex(L_1) \ldots \lex(L_{n-1}) \lex(L_n)
\]

\end{corollary}

If a sequence contains $\lex$ and $\refi$, every $\refi(R)$ in order can be moved
to the beginning of the sequence and then replaced by $\lex(L)$. What results
is a sequence containing only lexicographic revisions.

\begin{example}

The sequence in the running example was changed to comprise lexicographic
revisions, refinements and severe antiwithdrawals only:

\[
\emptyset
\lex(y)
\lex(\neg x \wedge y)
\refi(x \wedge z) \lex(x \wedge y \wedge z)
\lex(z) \sev(\neg z) \lex(\neg z)
\]

The corollary above shows that $\refi(x \wedge z)$ can be moved to the beginning
of the sequence and there turned into a lexicographic revision:

\[
\emptyset
\lex(x \wedge z)
\lex(y)
\lex(\neg x \wedge y)
\lex(x \wedge y \wedge z)
\lex(z) \sev(\neg z) \lex(\neg z)
\]

\end{example}

This transformation can be used to prove the folklore theorem linking
lexicographic revision with maxsets:

\begin{eqnarray*}
\maxset(F) &\equiv& F					\\
\maxset(F_1, F_2) &\equiv&
\left\{
\begin{array}{ll}
F_1 \wedge F_2		& \mbox{if consistent}		\\
F_1			& \mbox{otherwise}
\end{array}
\right.								\\
\maxset(F_1,\ldots,F_n)
&\equiv&
\maxset(\maxset(F_1,\ldots,F_{n-1}),F_n)
\end{eqnarray*}

The theorem establishes that $\maxset$ can be used to determine the minimal
models of a formula in the ordering resulting from a sequence of lexicographic
revisions. The proof is included here for the sake of completeness.

\begin{theorem}

For every formula $P$ and sequence of formulae $L_1,\ldots,L_n$, it holds:

\[
\min(\emptyset \lex(L_1) \ldots \lex(L_n),P)
=
\mod(\maxset(P,L_n,\ldots,L_1))
\]

\end{theorem}

\proof Proved by induction on the length of the sequence. With $n=0$,
$\maxset(P)=P$ and $\min(\emptyset,P)=\mod(P)$. The claim therefore holds.

If the claim holds for $n-1$ formulae, then
$\min(\emptyset \lex(L_2) \ldots \lex(L_n),P) =
\mod(\maxset(P,L_n,\ldots,L_2))$.
The same has to be proved with a formula $L_1$ more.

Let $C=[C(0),\ldots,C(m)]=\emptyset \lex(L_2) \ldots \lex(L_n)$. By the above
theorem, $\emptyset \lex(L_1) \lex(L_2) \ldots \lex(L_n) = \emptyset \lex(L_2)
\ldots \lex(L_n) \refi(L_1) = C \refi(L_1)$.

Let $i$ be the index such that $\min(C,P)=C(i) \cap \mod(P)$. This implies that
$C(0),\ldots,C(i-1)$ do not intersect $\mod(P)$. By definition, $C \refi(L_1)$
is:

\begin{eqnarray*}
\lefteqn{[C(0),\ldots,C(m)] \refi(L_1) = }		\\
&=& [
C(0) \cap \mod(L_1),
C(0) \backslash \mod(L_1),
\ldots,
C(i-1) \cap \mod(L_1),
C(i-1) \backslash \mod(L_1),				\\
&&~
C(i) \cap \mod(L_1),
C(i) \backslash \mod(L_1),
\ldots
C(m) \cap \mod(L_1),
C(m) \backslash \mod(L_1)
]
\end{eqnarray*}

Since $C(0),\ldots,C(i-1)$ do not intersect $\mod(P)$, the minimal class of $C
\refi(P)$ doing that is $C(i) \cap \mod(L_1)$ if not empty and $C(i) \backslash
\mod(L_1)$ otherwise. In the second case, $C(i) \backslash \mod(L_1)=C(i)$
since $C(i) \cap \mod(L_1)=\emptyset$. Therefore, $\min(C \refi(L_1),P)$ is
$C(i) \cap \mod(L_1) \cap \mod(P)$ if not empty and $C(i) \cap \mod(P)$
otherwise.

By the inductive assumption, $\min(C,P)=\mod(\maxset(P,L_n,\ldots,L_2))$, and
by definition of $i$ it holds $C(i) \cap \mod(P)=\min(C,P)$. Therefore, $\min(C
\refi(L_1),P)$ is $\mod(\maxset(P,L_n,\ldots,L_2)) \cap \mod(L_1)$ if not empty,
and $\mod(\maxset(P,L_n,\ldots,L_2)$ otherwise. In terms of formulae, $\min(C
\refi(P),P)$ is the set of models of $\maxset(P,L_n,\ldots,L_2) \wedge L_2$ if
this formula is consistent and of $\maxset(P,L_n,\ldots,L_2)$ otherwise. By the
recursive definition of maxset, $\min(C \refi(L_1),P) =
\mod(\maxset(P,L_n,\ldots,L_2,L_1))$.~\qed

In a sequence of lexicographic revisions, refinements and severe
antiwithdrawals, if the first operator of the sequence that is not a
lexicographic revision is a refinement it can be turned into a lexicographic
revision and moved to the beginning of the sequence. If it is a severe
antiwithdrawal, a more complex change needs to be applied to the sequence.

As the previous refinements can be turned into lexicographic revisions, the
previous belief change operators can be all assumed to be lexicographic
revisions. In other words, the considered sequence has all lexicographic
revisions but the last operator, which is a severe antiwithdrawal. Such a
sequence can be modified as follows, where $B=\under(S;L_1,\ldots,L_1)$ is a
formula defined below.

\[
\emptyset
\lex(L_1) \ldots \lex(L_n)
\sev(S)
=
\emptyset
\lex(L_1 \vee B) \ldots \lex(L_n \vee B) \lex(B)
\]

Intuitively, $B$ is constructed so that it collects all models that are in the
same class of the minimal ones of $S$ or in lower classes. Disjoining every
revising formula with $B$ ensures that these models remain in class zero over
each revision. The claim therefore requires two proofs: first, that $B$
actually comprises these models; second, that modifying the lexicographic
sequence this way does not change the resulting total preorder.

\begin{definition}

The {\em underformula} of a sequence of formulae is:

\begin{eqnarray*}
\lefteqn{\under(S;\epsilon) = \top }		\\
\lefteqn{\under(S;L_n,L_{n-1},\ldots,L_1) = } 	\\
&=&
\left\{
\begin{array}{ll}
L_n \wedge \under(S \wedge L_n;L_{n-1},\ldots,L_1)
			& \mbox{if $S \wedge L_n$ is consistent} \\
L_n \vee \under(S;L_{n-1},\ldots,L_1)   & \mbox{otherwise}
\end{array}
\right.
\end{eqnarray*}

\end{definition}

Informally, this construction includes as alternatives the formulae that are
excluded from $\maxset(P,L_n,\ldots,L_1)$ because they are inconsistent with
the partially built maxset $M$. Starting from $M=S$, the procedure of maxset
construction adds $L_i$ to $M$ if $M \wedge L_i$ is consistent. Otherwise,
$L_i$ is skipped. This procedure results in the minimal models of $S$. If $M
\wedge L_i$ is inconsistent only because of $S$, its models are in lower
classes than all models of $S$ in the final total preorder. Disjoining $M$ with
$L_i$ gathers all such models. This is obtained in the last case of the
definition: $L_i$ is disjoined with the underformula but not added to $M$. This
intuition is formalized by the following theorem.

\begin{lemma}

If $C=[C(0),\ldots,C(m)]=\emptyset \lex(L_1) \ldots \lex(L_n)$ and $i$ is the
minimal index such that $C(i) \cap \mod(S) \not= \emptyset$, then:

\[
C(0) \cup \cdots \cup C(i) = \mod(\under(S;L_n,\ldots,L_1))
\]

\end{lemma}

\proof The class of a model $I$ is lower or equal than all classes of $\mod(S)$
if and only if $I \in \min(C,S \vee \form(I))$. As a result, $C(0) \cup \cdots
\cup C(i)$ is defined by $\{I ~|~ I \models \maxset(S \vee
\form(I),L_n,\ldots,L_1)\}$. This set can be inductively proved to be equal to
$\mod(\under(S;L_n,\ldots,L_1))$. By induction on $n$, the following is proved:

\[
I \models \maxset(S \vee \form(I),L_n,\ldots,L_1)\}
\mbox{ iff } 
I \models \under(S;L_n,\ldots,L_1)
\]

With $m=0$, by definition $\under(S;\epsilon)=\top$, which contains all models;
$\maxset(S \vee \form(I))=S \vee \form(I)$, which always contains $I$. The base
case is therefore proved.

The induction step assumes:

\[
I \models \maxset(S \vee \form(I),L_{n-1},\ldots,L_1)
\mbox{ iff }
I \models \under(S;L_{n-1},\ldots,L_1)
\]

\

The claim is the same with $L_n$ added. The underformula of the claim is by
definition:

\begin{eqnarray*}
\lefteqn{\under(S;L_n,L_{n-1},\ldots,L_1) = }		\\
&
\left\{
\begin{array}{ll}
L_n \wedge \under(S \wedge L_n;L_{n-1},\ldots,L_1)
	& \mbox{ if $S \wedge L_1$ is consistent }	\\
L_n \vee \under(S;L_{n-1},\ldots,L_1)
	& \mbox{otherwise}
\end{array}
\right.
\end{eqnarray*}

The maxset is instead:

\begin{eqnarray*}
\lefteqn{\maxset(S \vee \form(I),L_n,L_{n-1},\ldots,L_1) = }		\\
&
\left\{
\begin{array}{ll}
\maxset((S \vee \form(I)) \wedge L_n,L_{n-1},\ldots,L_1)
	& \mbox{if $(S \vee \form(I)) \wedge L_n$ is consistent}	\\
\maxset(S \vee \form(I),L_{n-1},\ldots,L_1)
	& \mbox{otherwise}
\end{array}
\right.
\end{eqnarray*}

The claim is proved if
$I \models \maxset(S \vee \form(I),L_n,L_{n-1},\ldots,L_1)$
is shown to be the same as
$I \models \under(S;L_n,L_{n-1},\ldots,L_1)$.
This is done by reformulating the first condition. Since $(S \vee \form(I))
\wedge L_n$ is consistent if and only if $S \wedge L_n$ is consistent or $I
\models L_n$, the first case in the definition of maxset can be divided into
three:

\begin{eqnarray*}
\lefteqn{\maxset(S \vee \form(I),L_n,L_{n-1},\ldots,L_1) = }		\\
&
\left\{
\begin{array}{ll}
\maxset((S \vee \form(I)) \wedge L_n,L_{n-1},\ldots,L_1)
	& \mbox{if $S \wedge L_n$ is consistent and $I \models L_n$}	\\
\maxset((S \vee \form(I)) \wedge L_n,L_{n-1},\ldots,L_1)
	& \mbox{if $S \wedge L_n$ is consistent and $I \not\models L_n$}\\
\maxset((S \vee \form(I)) \wedge L_n,L_{n-1},\ldots,L_1)
	& \mbox{if $S \wedge L_n$ is inconsistent and $I \models L_n$}	\\
\maxset(S \vee \form(I),L_{n-1},\ldots,L_1)
	& \mbox{if $S \wedge L_n$ is inconsistent and $I \not\models L_n$}
\end{array}
\right.
\end{eqnarray*}

In the first case, since $I \models L_n$ it follows that 
$I \models \maxset((S \vee \form(I)) \wedge L_n,L_{n-1},\ldots,L_1)$
is equivalent to 
$I \models L_n \wedge \maxset((S \vee \form(I)) \wedge L_n,L_{n-1},\ldots,L_1)$.
The first argument $(S \vee \form(I)) \wedge L_n$ can be
rewritten $(S \wedge L_n) \vee \form(I)$ since $I \models L_n$.
By the induction assumption, 
$I \models L_n \wedge \maxset((S \wedge L_n) \vee \form(I),L_{n-1},\ldots,L_1)$
is the same as
$I \models L_n \wedge \under(S \wedge L_n;L_{n-1},\ldots,L_1)$.
Since $S \wedge L_1$ is consistent, the latter is equivalent to
$I \models \under(S;L_n,\ldots,L_1)$.

In the second case, the claim is proved by showing that $I$ satisfies neither
$\maxset((S \vee \form(I)),L_n,L_{n-1},\ldots,L_1)$ nor
$\under(S;L_n,L_{n-1},\ldots,L_1)$. Since $S \wedge L_n$ is consistent, $(S
\vee \form(I)) \wedge L_n$ is also consistent. As a result, $\maxset((S \vee
\form(I)),L_n,L_{n-1},\ldots,L_1)$ implies $(S \vee \form(I)) \wedge L_n$,
which implies $L_n$. Since $I \not\models L_n$, it follows $I \not\models
\maxset((S \vee \form(I)),L_n,L_{n-1},\ldots,L_1)$. This model does not satisfy
$\under(S;L_n,L_{n-1},\ldots,L_1)$ either, because this formula is equal to
$L_n \wedge \under(S \wedge L_n;L_{n-1},\ldots,L_1)$ when $S \wedge L_n$ is
consistent.

In the third case, $(S \vee \form(I)) \wedge L_n \equiv \form(I)$ since $S
\wedge L_n$ is inconsistent and $I \models L_n$. As a result, $I$ is the only
model of $\maxset(S \vee \form(I),L_n,L_{n-1},\ldots,L_1)$.

Together with the fourth case, this means that if $S \wedge L_n$ is
inconsistent then $I \models \maxset(S \vee \form(I),L_n,L_{n-1},\ldots,L_1)$
if and only if $I \models L_n$ or $I \models \maxset(S \vee
\form(I),L_{n-1},\ldots,L_1)$. By the induction assumption, the latter is
equivalent to $I \models \under(S;L_{n-1},\ldots,L_1)$. Therefore, the condition
can be rewritten as $I \models L_n \vee \under(S;L_{n-1},\ldots,L_1)$. Since $S
\wedge L_n$ is inconsistent, $L_n \vee
\under(S;L_{n-1},\ldots,L_1)=\under(S;L_n,L_{n-1},\ldots,L_1)$.~\qed

It is now shown that $\under(S,L_n,\ldots,L_1)$ allows rewriting the sequence
without affecting the final total preorder.

\begin{theorem}

If $B=\under(S;L_n,\ldots,L_1)$, then:

\[
\emptyset \lex(L_1) \ldots \lex(L_n) \sev(S)
\equiv
\emptyset \lex(L_1 \vee B) \ldots \lex(L_n \vee B) \lex(B)
\]

\end{theorem}

\proof The claim is proved by showing that for every consistent formula $P$,
its minimal models according to the two total preorders are the same.

By the previous theorem, if $B=\under(S;L_n,\ldots,L_1)$ then $\mod(B) = C(0)
\cup \cdots \cup C(i)$, where $C=[C(0),\ldots,C(m)]=\emptyset \lex(L_1) \ldots
\lex(L_n)$ and $i$ is the minimal index such that $C(i) \cap \mod(S) \not=
\emptyset$. Since $C \sev(S)=[C(0) \cup \cdots \cup
C(i),C(i+1),\ldots,C(m)]=[\mod(B),C(i+1),\ldots,C(m)]$, it follows that:

\[
\min(P,C \sev(S)) =
\left\{
\begin{array}{ll}
\mod(P) \cap \mod(B)	& \mbox{if not empty}				\\
\mod(P) \cap C(j)	& \mbox{otherwise, for $j$ minimal index}	\\
			& \mbox{such that this set is not empty}
\end{array}
\right.
\]

Since $C=\emptyset \lex(L_1) \ldots \lex(L_n)$, $\mod(P) \cap C(j)$ for the
minimal $j$ for which this set is not empty is the set of models of
$\maxset(P,L_n,\ldots,L_1)$. As a result, $\min(P,C \sev(S))$ can be rewritten
as:

\[
\min(P,C \sev(S)) =
\left\{
\begin{array}{ll}
\mod(P \wedge B)		& \mbox{if consistent}		\\
\mod(\maxset(P,L_n,\ldots,L_1))	& \mbox{otherwise}
\end{array}
\right.
\]

Let $C'=\emptyset \lex(L_1 \vee B) \ldots \lex(L_n \vee B) \lex(B)$. It is now
shown that $\min(C \sev(S),P) = \min(C',P)$. Since $C'$ results from applying a
number of lexicographic revisions to an empty total preorder, it holds:

\begin{eqnarray*}
\min(P,C') &=& \maxset(P,B,L_n \vee B,\ldots,L_1 \vee B)
\end{eqnarray*}

If $P \wedge B$ is consistent, then it is consistent with all formulae $L_i
\vee B$ but also entails all of them. As a result:

\begin{eqnarray*}
\min(P,C')
&=& \maxset(P,B,L_n \vee B,\ldots,L_1 \vee B)				\\
&=& \maxset(P \wedge B,L_n \vee B,\ldots,L_1 \vee B)			\\
&=& P \wedge B \wedge (L_n \vee B) \wedge \cdots \wedge (L_1 \vee B)	\\
&=& P \wedge B
\end{eqnarray*}

If $P \wedge B$ is inconsistent, then $P \models \neg B$; therefore:

\begin{eqnarray*}
\min(P,C')
&=& \maxset(P,B,L_n \vee B,\ldots,L_1 \vee B)				\\
&=& \maxset(P,L_n \vee B,\ldots,L_1 \vee B)				\\
\end{eqnarray*}

Since $P \models \neg B$, then $P \models \neg B$. As a result, $P \wedge (L_n
\vee B)$ is equivalent to $P \wedge L_i$. This implies that $\maxset(P,L_n \vee
B,\ldots,L_1 \vee B)$ is $\maxset(P \wedge L_n,L_{n-1}\ldots,L_1 \vee B)$ if $P
\wedge L_n$ is consistent, otherwise it is $\maxset(P,L_{m-1}\ldots,L_1 \vee
B)$. This argument can be repeated for every $P \wedge \bigwedge L$ with $L
\subseteq \{L_n,\ldots,L_i\}$, proving that the result is
$\maxset(P,L_n,\ldots,L_1)$.~\qed

These theorems tell how to modify a sequence into one that only contains
lexicographic revisions: starting from the beginning, the first operator that
is not $\lex$ can be $\refi(R)$ or $\sev(S)$; in the first case, it is turned
into $\lex(R)$ and moved at the beginning of the sequence; in the second case,
the underformula $B$ of the previous revisions (which are all lexicographic by
assumption) is used to replace $\sev(S)$ with $\lex(B)$ and is disjoined to all
previous revisions. This part of the sequence now contains only lexicographic
revisions, and the process can therefore be repeated for the next $\refi$ or
$\sev$ operator. The final result is a sequence of lexicographic revisions
applied to the empty preorder.

Such a sequence not only has the correct vale $\min(C,P)$ at each step, but
also the same final preorder of the original sequence. This implies that it is
equivalent to it even regarding subsequent revisions.

\begin{example}

The sequence in the running example has been shown to be equivalent to the
following one, which only contains lexicographic revisions and a severe
antiwithdrawal.

\[
\emptyset
\lex(x \wedge z)
\lex(y)
\lex(\neg x \wedge y)
\lex(x \wedge y \wedge z)
\lex(z)
\sev(\neg z)
\lex(\neg z)
\]

The severe antiwithdrawal can be turned into a lexicographic revision by first
calculating its underformula:

\begin{eqnarray*}
\lefteqn{
\under(\neg z; z, x \wedge y \wedge z, \neg x \wedge y, y, x \wedge z) =
}									\\
&=&
z \vee \under(\neg z; x \wedge y \wedge z, \neg x \wedge y, y, x \wedge z) \\
&=&
z \vee (x \wedge y \wedge z) \vee
\under(\neg z; \neg x \wedge y, y, x \wedge z) \\
&=&
z \vee (x \wedge y \wedge z) \vee
(\neg x \wedge y \wedge \under(\neg z \wedge \neg x \wedge y, y, x \wedge z)) \\
&=&
z \vee (x \wedge y \wedge z) \vee
(\neg x \wedge y \wedge y \wedge
\under(\neg z \wedge \neg x \wedge y \wedge y, x \wedge z)) \\
&=&
z \vee (x \wedge y \wedge z) \vee
(\neg x \wedge y \wedge y \wedge
((x \wedge z) \vee
\under(\neg z \wedge \neg x \wedge y \wedge y))) \\
&=&
z \vee (x \wedge y \wedge z) \vee
(\neg x \wedge y \wedge y \wedge
((x \wedge z) \vee \top))
\end{eqnarray*}

This formula is equivalent to $B=z \vee (\neg x \wedge y)$. The sequence is
therefore turned into:

\[
\emptyset
\lex((x \wedge z) \vee B)
\lex(y \vee B)
\lex((\neg x \wedge y) \vee B)
\lex((x \wedge y \wedge z) \vee B)
\lex(z \vee B)
\lex(B)
\lex(\neg z)
\]

Some simplifications can be then applied. For example, $(x \wedge z) \vee B =
(x \wedge z) \vee (z \vee (\neg x \wedge y)) \equiv z \vee (\neg x \wedge y)$
and $y \vee (z \vee (\neg x \wedge y)) \equiv y \vee z$.

In this particular case, only one severe antiwithdrawal occurs. More generally,
they are transformed into lexicographic revisions starting from the first.

\end{example}

The only apparent drawback of this procedure is that every $\sev(S)$ requires
the underformula $B$ to be disjoined to all previous formulae. This makes $B$
to be included in the underformula of the next $\sev(S')$. This problem is
solved by leaving the sequence as it is and processing it as if the
transformation has been done.

\subsection{Algorithm}

A sequence contains only lexicographic revisions and refinements applied to the
empty ordering can be turned into a sequence of lexicographic revisions by
moving all refinements to the beginning. After this change, the minimal models
of a formula $P$ can be calculated using the maxset construction. Since the
refinements are moved to the start of the sequence in the order in which they
are encountered, they end up there in reverse order. As a result, the maxset
can be calculated from the original sequence following the order that would
result from the simplification:

\noindent%
\framebox{
\hskip -12pt
\vbox{
\begin{enumerate}

\item start with $M=P$;

\item proceeding from the end to the start of the sequence,
for every $\lex(L_i)$ turn $M$ into $M \wedge L_i$ if this formula is
consistent;

\item from the start to the end of the sequence, for every $\refi(R_i)$
turn $M$ into $M \wedge R_i$ if this formula is consistent.

\end{enumerate}

{\bf The back and forth algorithm}.

\

}
}

The following figure shows how the algorithm proceeds when computing a formula
equivalent to the set of the minimal models of $P$. Every formula encountered
following the arrows is conjoined with it if that does not result in
contradiction.

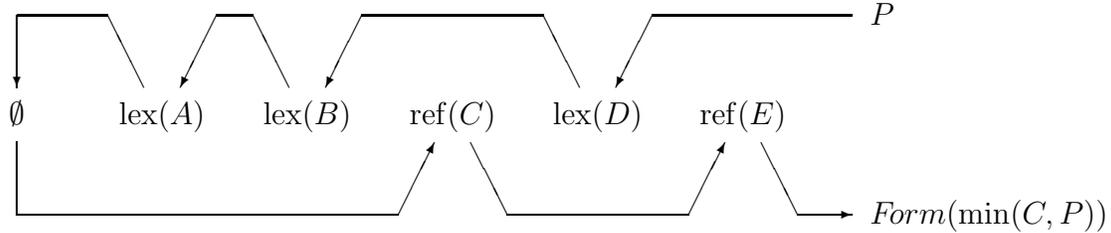
\begingroup
\expandafter\ifx\ttyfig\relax\else\let\-=\%\fi
\noindent
\setlength{\unitlength}{5000sp}%
\begingroup\makeatletter\ifx\SetFigFont\undefined%
\gdef\SetFigFont#1#2#3#4#5{%
  \reset@font\fontsize{#1}{#2pt}%
  \fontfamily{#3}\fontseries{#4}\fontshape{#5}%
  \selectfont}%
\fi\endgroup%
\begin{picture}(4287,1204)(4639,-3770)
\thinlines
{\color[rgb]{0,0,0}\put(8821,-2671){\line(-1, 0){990}}
\put(7831,-2671){\vector(-1,-2){180}}
}%
{\color[rgb]{0,0,0}\put(7471,-3031){\line(-1, 2){180}}
\put(7291,-2671){\line(-1, 0){900}}
\put(6391,-2671){\vector(-1,-2){180}}
}%
{\color[rgb]{0,0,0}\put(6031,-3031){\line(-1, 2){180}}
\put(5851,-2671){\line(-1, 0){180}}
\put(5671,-2671){\vector(-1,-2){180}}
}%
{\color[rgb]{0,0,0}\put(5311,-3031){\line(-1, 2){180}}
\put(5131,-2671){\line(-1, 0){450}}
\put(4681,-2671){\vector( 0,-1){360}}
}%
{\color[rgb]{0,0,0}\put(4681,-3301){\line( 0,-1){360}}
\put(4681,-3661){\line( 1, 0){1890}}
\put(6571,-3661){\vector( 1, 2){180}}
}%
{\color[rgb]{0,0,0}\put(6931,-3301){\line( 1,-2){180}}
\put(7111,-3661){\line( 1, 0){900}}
\put(8011,-3661){\vector( 1, 2){180}}
}%
{\color[rgb]{0,0,0}\put(8371,-3301){\line( 1,-2){180}}
\put(8551,-3661){\vector( 1, 0){270}}
}%
\put(8911,-2716){\makebox(0,0)[lb]{\smash{{\SetFigFont{12}{24.0}{\rmdefault}{\mddefault}{\updefault}{\color[rgb]{0,0,0}$P$}%
}}}}
\put(7561,-3211){\makebox(0,0)[b]{\smash{{\SetFigFont{12}{24.0}{\rmdefault}{\mddefault}{\updefault}{\color[rgb]{0,0,0}$\lex(D)$}%
}}}}
\put(6121,-3211){\makebox(0,0)[b]{\smash{{\SetFigFont{12}{24.0}{\rmdefault}{\mddefault}{\updefault}{\color[rgb]{0,0,0}$\lex(B)$}%
}}}}
\put(5401,-3211){\makebox(0,0)[b]{\smash{{\SetFigFont{12}{24.0}{\rmdefault}{\mddefault}{\updefault}{\color[rgb]{0,0,0}$\lex(A)$}%
}}}}
\put(4681,-3211){\makebox(0,0)[b]{\smash{{\SetFigFont{12}{24.0}{\rmdefault}{\mddefault}{\updefault}{\color[rgb]{0,0,0}$\emptyset$}%
}}}}
\put(8911,-3706){\makebox(0,0)[lb]{\smash{{\SetFigFont{12}{24.0}{\rmdefault}{\mddefault}{\updefault}{\color[rgb]{0,0,0}$\form(\min(C,P))$}%
}}}}
\put(8281,-3211){\makebox(0,0)[b]{\smash{{\SetFigFont{12}{24.0}{\rmdefault}{\mddefault}{\updefault}{\color[rgb]{0,0,0}$\refi(E)$}%
}}}}
\put(6841,-3211){\makebox(0,0)[b]{\smash{{\SetFigFont{12}{24.0}{\rmdefault}{\mddefault}{\updefault}{\color[rgb]{0,0,0}$\refi(C)$}%
}}}}
\end{picture}%
\nop{
  /-<-\   /-<-\   /--------<--\   /----------<-- P
 |     | |     | |             | |
 V     | V     | V             | V
 0    lex(A)  lex(B)  ref(C)  lex(D)  ref(E)
 |                     ^ |             ^ |
 |                     | |             | |
  \--->---------------/   \--->-------/   \---> form(min(C,P))
}
\endgroup

The back and forth algorithm works because it builds the maxset starting from
$P$ and proceding in the same order as if the refinements were moved to the
start of the sequence. Its correctness is therefore proved by
Theorem~\ref{ref-lex}. For the same reason, a similar mechanism can be used to
determine an underformula instead of a maxset.


This is important because a sequence may contain lexicographic revisions,
refinements and severe antiwithdrawals. Assuming that the underformulae for the
latter have all been determined, at each severe antiwithdrawal encountered
while going back, if $B$ is consistent with the current maxset $M$ then $M$ is
turned into $M \wedge B$ and the procedure ``bounces'' back in the forward
direction. This is because if $M \wedge B$ is consistent then the previous
$\lex(L)$ in the original sequence would be turned into $\lex(L \vee B)$ in the
modified sequence. As a result, $M$ is consistent with all of them, but their
addition is irrelevant because $M$ is already conjoined with $B$.

\noindent%
\framebox{
\hskip -12pt
\vbox{
\begin{enumerate}

\item Start at the end of the sequence with $M=P$ and go back;

\item for every $\lex(L_i)$ turn $M$ into $M \wedge L_i$ if this formula is
consistent; regardless, continue going backwards;

\item for every $\sev(S)$, if $M$ is consistent with its underformula $B$ then
turn $M$ into $M \wedge B$ and bounce forward, toward the end of the sequence;
otherwise, continue going backwards;

\item at the start of the sequence, bounce forward, toward the end of the
sequence;

\item when proceding forward, for every $\refi(R_i)$ turn $M$ into $M \wedge
R_i$ if this formula is consistent; regardless, continue going forward.

\end{enumerate}

{\bf The back, bounce and forth algorithm}

\

}
}

The fourth point can be omitted by placing $\sev(\top)$ at the very beginning
of the sequence. This marker signals the algorithm to bounce forward without
the need to verify whether the sequence is at the start. At the end $M$ has
models $\min(C,P)$ where $C$ is the final ordering.

The algorithm works because it builds a formula that is the same that would
have been produced when creating the maxset of the modified sequence that only
contains lexicographic revisions.

The following figure shows how the algorithm moves in a segment of the
sequence. When it reaches $\sev(F)$, if the formula that is currently being
built is consistent with the underformula of this severe antiwithdrawal then
the algorithm bounces forward to $\refi(H)$, otherwise it keeps going back, to
$\lex(D)$.

\begingroup
\expandafter\ifx\ttyfig\relax\else\let\-=\%\fi
\setlength{\unitlength}{5000sp}%
\begingroup\makeatletter\ifx\SetFigFont\undefined%
\gdef\SetFigFont#1#2#3#4#5{%
  \reset@font\fontsize{#1}{#2pt}%
  \fontfamily{#3}\fontseries{#4}\fontshape{#5}%
  \selectfont}%
\fi\endgroup%
\begin{picture}(4884,1074)(6199,-3703)
\thinlines
{\color[rgb]{0,0,0}\put(11071,-2671){\line(-1, 0){1080}}
\put(9991,-2671){\vector(-1,-2){180}}
}%
{\color[rgb]{0,0,0}\put(9631,-3031){\line(-1, 2){180}}
\put(9451,-2671){\line(-1, 0){180}}
\put(9271,-2671){\vector(-1,-2){180}}
}%
{\color[rgb]{0,0,0}\put(8911,-3031){\line(-1, 2){180}}
\put(8731,-2671){\line(-1, 0){900}}
\put(7831,-2671){\vector(-1,-2){180}}
}%
{\color[rgb]{0,0,0}\put(7471,-3031){\line(-1, 2){180}}
\put(7291,-2671){\vector(-1, 0){1080}}
}%
{\color[rgb]{0,0,0}\put(6211,-3661){\line( 1, 0){360}}
\put(6571,-3661){\vector( 1, 2){180}}
}%
{\color[rgb]{0,0,0}\put(9091,-3301){\line( 1,-2){180}}
\put(9271,-3661){\line( 1, 0){900}}
\put(10171,-3661){\vector( 1, 2){180}}
}%
{\color[rgb]{0,0,0}\put(10531,-3301){\line( 1,-2){180}}
\put(10711,-3661){\vector( 1, 0){360}}
}%
{\color[rgb]{0,0,0}\put(6931,-3301){\line( 1,-2){180}}
\put(7111,-3661){\line( 1, 0){900}}
\put(8011,-3661){\vector( 1, 2){180}}
}%
{\color[rgb]{0,0,0}\put(8371,-3301){\line( 1,-2){180}}
\put(8551,-3661){\vector( 1, 0){630}}
}%
\put(7561,-3211){\makebox(0,0)[b]{\smash{{\SetFigFont{12}{24.0}{\rmdefault}{\mddefault}{\updefault}{\color[rgb]{0,0,0}$\lex(D)$}%
}}}}
\put(9001,-3211){\makebox(0,0)[b]{\smash{{\SetFigFont{12}{24.0}{\rmdefault}{\mddefault}{\updefault}{\color[rgb]{0,0,0}$\sev(F)$}%
}}}}
\put(9721,-3211){\makebox(0,0)[b]{\smash{{\SetFigFont{12}{24.0}{\rmdefault}{\mddefault}{\updefault}{\color[rgb]{0,0,0}$\lex(G)$}%
}}}}
\put(6841,-3211){\makebox(0,0)[b]{\smash{{\SetFigFont{12}{24.0}{\rmdefault}{\mddefault}{\updefault}{\color[rgb]{0,0,0}$\refi(C)$}%
}}}}
\put(8281,-3211){\makebox(0,0)[b]{\smash{{\SetFigFont{12}{24.0}{\rmdefault}{\mddefault}{\updefault}{\color[rgb]{0,0,0}$\refi(E)$}%
}}}}
\put(10441,-3211){\makebox(0,0)[b]{\smash{{\SetFigFont{12}{24.0}{\rmdefault}{\mddefault}{\updefault}{\color[rgb]{0,0,0}$\refi(H)$}%
}}}}
\end{picture}%
\nop{
<----------\   /--------<--\   /-<-\   /-----------<--
            | |             | |     | |
            | V             | V     | V
   ref(C)  lex(D)  ref(E)  sev(F)  lex(G)  ref(H)
    ^ |             ^ |       |             ^ |
    | |             | |       |             | |
->-/   \-->--------/   \---->  \-----------/   \-------->
}
\endgroup

This construction produces a maxset. The underformula of each severe
antiwithdrawal is built similarly. The effect of a severe anticontraction
$\sev(S')$ with underformula $B'$ encountered during the construction of
another underformula may have only two possible effects: $B'$ is ignored; or,
$B'$ is added but makes the algorithm bounce forward. This proves that an
underformula $B$ may at most contain a single previous underformula $B'$ and
some formulae in between. As a result, the underformulae do not exponentially
blow up.

\begin{example}

The algorithm is applied to the sequence of the running example. It first
determines the underformula of the severe antiwithdrawal, then the base at the
end of the sequence.

\[
\emptyset
\lex(y)
\lex(\neg x \wedge y)
\refi(x \wedge z)
\lex(x \wedge y \wedge z)
\lex(z)
\sev(\neg z)
\lex(\neg z)
\]

The first step is to determine the underformula of the first severe revision in
the sequence. This is done by following the back and forth procedure: first go
back through the lexicographic revisions, then come forth through the
refinements.

\begingroup
\expandafter\ifx\ttyfig\relax\else\let\-=\%\fi
\ttytex{
\[
\begin{array}{cccccccc}
0 & \lex(y) & \lex(\neg x \wedge y) & \refi(x \wedge z) &
\lex(x \wedge y \wedge z) & \lex(z) & \sev(\neg z) & \lex(\neg z)	\\
  & 5 & 4 &   & 3 & 2 & 1						\\
  &   &   & 6
\end{array}
\]
}{
0 lex(y) lex(-x&y) ref(x&z) lex(x&y&z) lex(z) sev(-z) lex(-z)
     5  <--  4  <--------------  3  <---  2 <--- 1
      \------------->  6  -------->
}
\endgroup

Following the numbers, the formulae are in the sequence $\neg z$, $z$, $x
\wedge y \wedge z$, $\neg x \wedge y$, $y$ and $x \wedge z$. As a result, the
underformula of the severe antiwithdrawal is calculated on this sequence:

\[
\under(\neg z; z, x \wedge y \wedge z, \neg x \wedge y, y, x \wedge z)
\]

This has been previously shown to be equivalent to $B=z \vee (\neg x \wedge
y)$. It allows determining the final base by following the arrows as in the
back, bounce and forth algorithm.

\begingroup
\expandafter\ifx\ttyfig\relax\else\let\-=\%\fi
\ttytex{
\[
\begin{array}{ccccccccc}
0 & \lex(y) & \lex(\neg x \wedge y) & \refi(x \wedge z) &
\lex(x \wedge y \wedge z) & \lex(z) & \sev(\neg z) & \lex(\neg z)	\\
  &   &   &   &   & 3?  & 2 & 1	\\ 
  &   &   &   &   &     &   &   & 3?
\end{array}
\]
}{
0 lex(y) lex(-x&y) ref(x&z) lex(x&y&z) lex(z) sev(-z) lex(-z)
                                       ...  <--  2  <--- 1
                                                  \---->
}
\endgroup

The choice of keeping going back or bouncing forth depends on the consistency
of the formula under construction with the underformula of $\sev(\neg z)$. In
this case, the formula is $\neg z$ and the underformula $z \vee (\neg x \wedge
y)$. Since their conjunction $\neg z \wedge \neg x \wedge y$ is consistent, the
algorithm bounces. Since there are no refinement after the severe
antiwithdrawal, the resulting base is this formula.

\end{example}

The algorithm can be adapted to work with the other considered revisions
without replacing them with $\lex$, $\refi$ and $\sev$, since each of them can
be ``locally'' replaced with a sequence of these three. As a result, while
going forward or backwards, sufficies to behave in the same way as if the
replacement has been done.

\section{Complexity}
\label{hardness}

The reductions shown in the previous sections prove that each considered belief
change operators can be turned into a lexicographic revision, possibly by first
calculating a maxset or an underformula. Both operations can be done by a
polynomial number of calls to a propositional satisfiability solver. Therefore,
the complexity of a sequence of arbitrary and mixed belief change operators is
in the complexity class \D{2}, which contains all problems that can be solved
by a polynomial number of calls to an \np\ oracle. The problem is also easily
shown to be hard for the same class even if all operators are lexicographic
revisions. This was previously published without proof~\cite{libe-97-c}.

\begin{theorem}
\label{lexi-hard}

The problems of establishing whether the base resulting from a sequence of
lexicographic, natural, restrained, very radical and severe revisions,
refinements and severe antiwithdrawals applied to the empty ordering implies a
formula is in \D{2}, and is \D{2}-hard even if the sequence comprises
lexicographic revisions only or refinements only.

\end{theorem}

\proof The back, bounce and forth algorithm shown in the previous section
calculates a formula equivalent to the set of the minimal models of a formula
in polynomial time, not counting that needed to determine propositional
satisfiability. This proves that the problem is in \D{2} for sequence
comprising only lexicographic revisions, refinements and severe antiwithdrawal.
Since all other considered operators can be reduced to these three, the problem
is in \D{2} for all of them.

Hardness is proved by reduction from the problem of establishing whether the
maximal lexicographic model of a formula $F$ over the alphabet
$\{x_1,\ldots,x_n\}$ satisfies $x_n$, which is \D{2}-complete~\cite{kren-88}. A
simple reduction translates this problem into the similar one where the formula
is satisfiable: a possibly unsatisfiable formula $F$ is turned into the
satisfiable formula
{} $(\neg x_0 \wedge \neg x_1 \wedge \cdots \neg x_n) \vee F$,
where $x_0$ is a new variable. A further reduction shows that the maximal model
of a satisfiable formula $F$ satisfies $x_n$ if and only if the base of
$\emptyset \lex(x_n) \ldots \lex(x_1) \lex(F)$ implies $x_n$. This proves that
entailment for a sequence of lexicographic revisions is \D{2}-hard. The
sequence is equivalent to $\emptyset \refi(F) \refi(x_1) \ldots \refi(x_n)$,
proving the \D{2}-hardness for refinements only. More generally, it is hard for
every alternation of these two belief change operators.~\qed

Since severe antiwithdrawal turns an empty total preorder into an empty
total preorder, a sequence comprising only this operator has low complexity:
entailment is equal to validity, \conp\  complete.

Sequences of mixed belief change operators are now considered. Two classes can
be shown to be \D{2}-hard: operators that can produce a lexicographically
maximal model, and operators that can refine the lowest class of an ordering.
In both cases, the alternation of operators does not matter, as long as they
have the given behavior.

\subsection{Lexicographic-finding revisions}

Entailment from a sequence of lexicographic revisions is \D{2}-hard by
Theorem~\ref{lexi-hard}. Some other belief change operators can be intermixed
without changing complexity. These are the ones that produce the same results
when applied after a sequence of revisions whose formulae are consistent.

\begin{theorem}

If $\rev$ is any of a class of revision operators such that
{} $\emptyset \rev(S_1) \ldots \rev(S_n)
{} \equiv
{} \emptyset \lex(S_1) \ldots \lex(S_n)$
whenever $S_1 \wedge \cdots \wedge S_n$ is consistent, then entailment for
$\rev$ is \D{2}-hard.

\end{theorem}

\proof Checking whether the lexicographically maximal model of a formula $F$
satisfies $x_n$ is \D{2}-hard~\cite{kren-88}. This model is also the only
element of class zero of $\emptyset \lex(x_n) \ldots \lex(x_1) \lex(F)$. Since
$x_1,\ldots,x_n$ is consistent,
{} $\emptyset \lex(x_n) \ldots \lex(x_1)$
is equivalent to
{} $\emptyset \rev(x_n) \ldots \rev(x_1)$
by assumption. Therefore, entailment from these two sequences is the same.~\qed

Moderate severe revision satisfies the premise of this theorem: it coincides
with lexicographic revision on all consistent sequences.

\begin{theorem}

If $S_1 \wedge \cdots \wedge S_n$ is consistent, then:

\[
\emptyset \msev(S_1) \ldots \msev(S_n)
=
\emptyset \lex(S_1) \ldots \lex(S_n)
\]

\end{theorem}

\proof It is inductively proved that 
$\emptyset \msev(S_1) \ldots \msev(S_n) = \emptyset \lex(S_1) \ldots \lex(S_n)$
and that
$\emptyset \msev(S_1) \ldots \msev(S_n)(0) =
\mod(S_1 \wedge \cdots \wedge S_n)$.

The base case is with $n=0$, where the claim holds because $\emptyset =
\emptyset$ and the conjunction of an empty sequence is $\top$.

Assuming that 
$\emptyset \msev(S_1) \ldots \msev(S_{n-1}) =
\emptyset \lex(S_1) \ldots \lex(S_{n-1})$,
and that
$\emptyset \msev(S_1) \ldots \msev(S_{n-1})(0) =
\mod(S_1 \wedge \cdots \wedge S_{n-1})$,
the same are proved with the addition of $S_n$.

Let $C=[C(0),\ldots,C(m)]$ be $\emptyset \msev(S_1) \ldots \msev(S_{n-1})$,
and
$M=S_1 \wedge \cdots \wedge S_{n-1}$.
By the induction assumptions,
$C=\emptyset \lex(S_1) \ldots \lex(S_{n-1})$, and
$C(0)=\mod(M)$.

The definition of $C \msev(S_n)$ depends
on $\min(C,S_n)=C(i) \cap \mod(S_n)$. Since $C(0)=\mod(M)$ and $M \wedge S_n$
is by assumption consistent, $i=0$. The definition of moderate severe revision
specializes to $i=0$ as:

\begin{eqnarray*}
\lefteqn{[C(0),\ldots,C(m)] \msev(S_n) = }			\\
&=& [
  C(0) \cap \mod(S_n),\ldots,C(m) \cap \mod(S_n),		\\
&&~
  (C(0) \cup \cdots \cup C(0)) \backslash \mod(S_n),
  C(1) \backslash \mod(S_n),\ldots,C(m) \backslash \mod(S_n)
]								\\
&=& [
  C(0) \cap \mod(S_n),\ldots,C(m) \cap \mod(S_n),		\\
&&~
  C(0) \backslash \mod(S_n),
  C(1) \backslash \mod(S_n),\ldots,C(m) \backslash \mod(S_n)
]								\\
&=&
[C(0),\ldots,C(m)] \lex(S_n)
\end{eqnarray*}

The total preorder $[C(0),\ldots,C(m)] \lex(S_n)$ is $\emptyset \lex(S_1)
\ldots \lex(S_{n-1}) \lex(S_n)$ because of the induction assumption
$C=\emptyset \lex(S_1) \ldots \lex(S_{n-1})$. Since $C(0)=\mod(S_1 \wedge
\cdots \wedge S_{n-1})$, it follows that the class zero of this ordering is $C
\msev(S_n)(0)=C(0) \cap \mod(S_n)=\mod(S_1 \wedge \cdots \wedge S_{n-1} \wedge
S_n)$, which concludes the proof of the induction claim.~\qed

A consequence of the two theorems above is that entailment from a sequence of
moderate severe revision is \D{2}-hard. The same holds even if lexicographic
and moderate severe revisions are mixed.

\begin{corollary}

Entailment from a sequence of moderate severe revision is \D{2}-hard.

\end{corollary}

\subsection{Bottom-refining revisions}

A revision operator is bottom-refining if it ``refines'' the lowest-index class
of the ordering that has models of the revising formula.

\begin{definition}

An operator $\rev$ is a revision if $\min(C \rev(P),T) = \min(C,P)$ and is a
bottom-revising revision if $C \rev(P)(1)=C(0) \backslash \mod(P)$ also holds
whenever $C(0) \cap \mod(P) \not= \emptyset$.

\end{definition}

A revision makes minimal the models that are the minimal models satisfying the
revising formula. Removing empty classes, class zero of $C \rev(P)$ is the
non-empty intersection $C(i) \cap P$ of minimal $i$. If $\rev$ is also
bottom-refining, this is $C(0) \cap \mod(P)$ if this intersection is not empty.
In this case, revising $C$ by $P$ splits $C(0)$ based on $P$, as shown in the
following example.

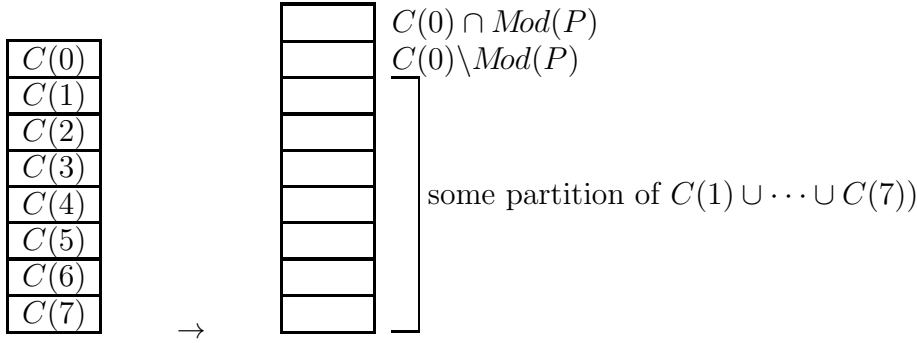
\ttytex{
\begin{tabular}{ccc}
\setlength{\unitlength}{5000sp}%
\begingroup\makeatletter\ifx\SetFigFont\undefined%
\gdef\SetFigFont#1#2#3#4#5{%
  \reset@font\fontsize{#1}{#2pt}%
  \fontfamily{#3}\fontseries{#4}\fontshape{#5}%
  \selectfont}%
\fi\endgroup%
\begin{picture}(516,1506)(4828,-4144)
\thicklines
{\color[rgb]{0,0,0}\put(4861,-2851){\line( 1, 0){450}}
}%
{\color[rgb]{0,0,0}\put(4861,-3031){\line( 1, 0){450}}
}%
{\color[rgb]{0,0,0}\put(4861,-3211){\line( 1, 0){450}}
}%
{\color[rgb]{0,0,0}\put(4861,-3391){\line( 1, 0){450}}
}%
{\color[rgb]{0,0,0}\put(4861,-3571){\line( 1, 0){450}}
}%
{\color[rgb]{0,0,0}\put(4861,-3751){\line( 1, 0){450}}
}%
{\color[rgb]{0,0,0}\put(4861,-3931){\line( 1, 0){450}}
}%
{\color[rgb]{0,0,0}\put(4861,-4111){\framebox(450,1440){}}
}%
\put(5086,-2806){\makebox(0,0)[b]{\smash{{\SetFigFont{12}{24.0}{\rmdefault}{\mddefault}{\itdefault}{\color[rgb]{0,0,0}$C(0)$}%
}}}}
\put(5086,-2986){\makebox(0,0)[b]{\smash{{\SetFigFont{12}{24.0}{\rmdefault}{\mddefault}{\itdefault}{\color[rgb]{0,0,0}$C(1)$}%
}}}}
\put(5086,-3166){\makebox(0,0)[b]{\smash{{\SetFigFont{12}{24.0}{\rmdefault}{\mddefault}{\itdefault}{\color[rgb]{0,0,0}$C(2)$}%
}}}}
\put(5086,-3346){\makebox(0,0)[b]{\smash{{\SetFigFont{12}{24.0}{\rmdefault}{\mddefault}{\itdefault}{\color[rgb]{0,0,0}$C(3)$}%
}}}}
\put(5086,-3526){\makebox(0,0)[b]{\smash{{\SetFigFont{12}{24.0}{\rmdefault}{\mddefault}{\itdefault}{\color[rgb]{0,0,0}$C(4)$}%
}}}}
\put(5086,-3706){\makebox(0,0)[b]{\smash{{\SetFigFont{12}{24.0}{\rmdefault}{\mddefault}{\itdefault}{\color[rgb]{0,0,0}$C(5)$}%
}}}}
\put(5086,-3886){\makebox(0,0)[b]{\smash{{\SetFigFont{12}{24.0}{\rmdefault}{\mddefault}{\itdefault}{\color[rgb]{0,0,0}$C(6)$}%
}}}}
\put(5086,-4066){\makebox(0,0)[b]{\smash{{\SetFigFont{12}{24.0}{\rmdefault}{\mddefault}{\itdefault}{\color[rgb]{0,0,0}$C(7)$}%
}}}}
\end{picture}%
& ~ ~ $\rightarrow$ ~ ~ &
\setlength{\unitlength}{5000sp}%
\begingroup\makeatletter\ifx\SetFigFont\undefined%
\gdef\SetFigFont#1#2#3#4#5{%
  \reset@font\fontsize{#1}{#2pt}%
  \fontfamily{#3}\fontseries{#4}\fontshape{#5}%
  \selectfont}%
\fi\endgroup%
\begin{picture}(768,1686)(4828,-4324)
\thicklines
{\color[rgb]{0,0,0}\put(4861,-2851){\line( 1, 0){450}}
}%
{\color[rgb]{0,0,0}\put(4861,-3031){\line( 1, 0){450}}
}%
{\color[rgb]{0,0,0}\put(4861,-3211){\line( 1, 0){450}}
}%
{\color[rgb]{0,0,0}\put(4861,-3391){\line( 1, 0){450}}
}%
{\color[rgb]{0,0,0}\put(4861,-3571){\line( 1, 0){450}}
}%
{\color[rgb]{0,0,0}\put(4861,-3751){\line( 1, 0){450}}
}%
{\color[rgb]{0,0,0}\put(4861,-3931){\line( 1, 0){450}}
}%
{\color[rgb]{0,0,0}\put(4861,-4111){\line( 1, 0){450}}
}%
{\color[rgb]{0,0,0}\put(4861,-4291){\framebox(450,1620){}}
}%
\thinlines
{\color[rgb]{0,0,0}\put(5401,-3031){\line( 1, 0){135}}
\put(5536,-3031){\line( 0,-1){1260}}
\put(5536,-4291){\line(-1, 0){135}}
}%
\put(5401,-2806){\makebox(0,0)[lb]{\smash{{\SetFigFont{12}{24.0}{\rmdefault}{\mddefault}{\updefault}{\color[rgb]{0,0,0}$C(0) \cap \mod(P)$}%
}}}}
\put(5401,-2986){\makebox(0,0)[lb]{\smash{{\SetFigFont{12}{24.0}{\rmdefault}{\mddefault}{\updefault}{\color[rgb]{0,0,0}$C(0) \backslash \mod(P)$}%
}}}}
\put(5581,-3661){\makebox(0,0)[lb]{\smash{{\SetFigFont{12}{24.0}{\rmdefault}{\mddefault}{\updefault}{\color[rgb]{0,0,0}$\mbox{some partition of } C(1) \cup \cdots \cup C(7))$}%
}}}}
\end{picture}%
\end{tabular}
}{
+------------+           +------------+
|    C(0)    |           |C(0)&mod(P) |
+------------+           +------------+
|    C(1)    |           |C(0)&mod(-P)|
+------------+           +------------+ 
|    C(2)    |           |            |   \                    .
+------------+    ==>    +------------+    |
|    C(3)    |           |            |    |
+------------+           +------------+    |
|    C(4)    |           |            |     \ some partition
+------------+           +------------+     / of C(1)u...uC(7)
|    C(5)    |           |            |    |
+------------+           +------------+    |
|    C(6)    |           |            |    |
+------------+           +------------+    |
|    C(7)    |           |            |   /
+------------+           +------------+
}

The name bottom-refining derives from the way the ``bottom'' class $C(0)$ is
partitioned (refined) into the part satisfying $P$ and the part not satisfy
$P$. How the other classes are changed is not constrained.

\begin{theorem}

Natural revision, restrained revision and severe revision are bottom-refining
revisions.

\end{theorem}

\proof All three operators are revisions, since they make the non-empty
intersection $C(i) \cap \mod(P)$ of minimal $i$ the new class zero.

The condition of bottom-refining only concerns the case where $C(0) \cap
\mod(P) \not= \emptyset$, where this index $i$ is $0$. The definition of
natural revisions specializes as follows:

\begin{eqnarray*}
\lefteqn{[C(0),\ldots,C(m)] \nat(P) = }			\\
&& [
C(i) \cap \mod(M),
C(0),
\ldots
C(i-1),
C(i) \backslash \mod(P),
C(i+1),
\ldots,
C(m)
]							\\
&& [
C(0) \cap \mod(M),
C(0) \backslash \mod(P),
C(1),
\ldots
C(m)
]							\\
\end{eqnarray*}

The definition of restrained revisions specializes as follows:

\begin{eqnarray*}
\lefteqn{[C(0),\ldots,C(m)] \mathrm{res}(P) = }			\\
&=& [
C(i)	\cap \mod(P),						\\
&&~
C(0)	\cap \mod(P),
C(0)	\backslash \mod(P),
\ldots,
C(i-1)	\cap \mod(P),
C(i-1)	\backslash \mod(P),					\\
&&~
C(i)	\backslash \mod(P),					\\
&&~
C(i+1)	\cap \mod(P),
C(i+1)	\backslash \mod(P),
\ldots
C(m)	\cap \mod(P),
C(m)	\backslash \mod(P)
]								\\
&=& [
C(0)	\cap \mod(P),
C(0)	\backslash \mod(P),
C(1)	\cap \mod(P),
C(1)	\backslash \mod(P),					\\
&&~
\ldots
C(m)	\cap \mod(P),
C(m)	\backslash \mod(P)
]
\end{eqnarray*}

The definition of severe revisions specializes as follows:

\begin{eqnarray*}
\lefteqn{[C(0),\ldots,C(m)] \sevr(P) = }			\\
&=& [
  C(i) \cap \mod(P),
  (C(0) \cup \cdots \cup C(i)) \backslash \mod(P),
  C(i+1),\ldots,C(m)
]								\\
&=& [
  C(0) \cap \mod(P),
  C(0) \backslash \mod(P),
  C(1),\ldots,C(m)
]
\end{eqnarray*}

All three revisions makes $C(0) \cap \mod(P)$ the new class zero and $C(0)
\backslash \mod(P)$ the new class one.~\qed

The complexity of arbitrary sequences of bottom-refining revisions is
established by the following theorem.

\begin{theorem}

Inference from a sequence of bottom-refining revisions is \D{2}-hard.

\end{theorem}

\proof The claim is proved by reduction from the problem of establishing
whether the lexicographically maximal model of a consistent formula $F$ over
alphabet $\{x_1,\ldots,x_n\}$ satisfies $x_n$.
The corresponding sequence of bottom refining revisions $\rev$ is the
following, where $y_1,\ldots,y_n${\plural} are fresh variables in bijective
correspondence with $x_1,\ldots,x_n$:

\[
\emptyset 
\rev(F)
\rev(y_1)
\rev(y_1 \rightarrow x_1)
\ldots
\rev(y_n)
\rev(y_n \rightarrow x_n)
\]

The reduction first introduces the models of $F$ as the class zero, then refines
it by $x_1$ if consistent, then by $x_2$ if consistent, etc. A single
bottom-refining revision for each variable cannot do that: if $x_1$ is
inconsistent with $F$ the effect of $F \rev(x_1)$ is to move the models of
$x_1$ in a class lower than that of $F$. This is why the new variable $y_1$ is
introduced first.

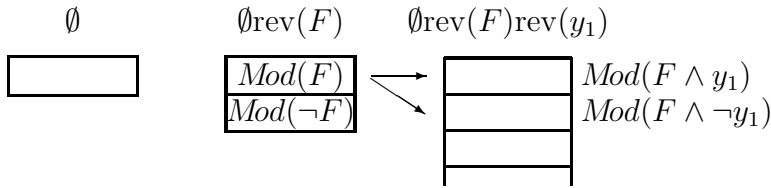
\setlength{\unitlength}{5000sp}%
\begingroup\makeatletter\ifx\SetFigFont\undefined%
\gdef\SetFigFont#1#2#3#4#5{%
  \reset@font\fontsize{#1}{#2pt}%
  \fontfamily{#3}\fontseries{#4}\fontshape{#5}%
  \selectfont}%
\fi\endgroup%
\begin{picture}(2883,957)(3568,-3334)
\thicklines
{\color[rgb]{0,0,0}\put(5761,-2851){\line( 1, 0){630}}
}%
{\color[rgb]{0,0,0}\put(5761,-3031){\line( 1, 0){630}}
}%
{\color[rgb]{0,0,0}\put(5761,-3211){\line( 1, 0){630}}
}%
{\color[rgb]{0,0,0}\put(5761,-3301){\line( 0, 1){630}}
\put(5761,-2671){\line( 1, 0){630}}
\put(6391,-2671){\line( 0,-1){630}}
}%
{\color[rgb]{0,0,0}\put(3601,-2851){\framebox(630,180){}}
}%
{\color[rgb]{0,0,0}\put(4681,-3031){\framebox(630,360){}}
}%
{\color[rgb]{0,0,0}\put(4681,-2851){\line( 1, 0){630}}
}%
\thinlines
{\color[rgb]{0,0,0}\put(5401,-2761){\vector( 1, 0){270}}
}%
{\color[rgb]{0,0,0}\put(5394,-2771){\vector( 3,-2){270}}
}%
\put(4996,-2806){\makebox(0,0)[b]{\smash{{\SetFigFont{12}{24.0}{\rmdefault}{\mddefault}{\updefault}{\color[rgb]{0,0,0}$\mod(F)$}%
}}}}
\put(4996,-2986){\makebox(0,0)[b]{\smash{{\SetFigFont{12}{24.0}{\rmdefault}{\mddefault}{\updefault}{\color[rgb]{0,0,0}$\mod(\neg F)$}%
}}}}
\put(6436,-2806){\makebox(0,0)[lb]{\smash{{\SetFigFont{12}{24.0}{\rmdefault}{\mddefault}{\updefault}{\color[rgb]{0,0,0}$\mod(F \wedge y_1)$}%
}}}}
\put(6436,-2986){\makebox(0,0)[lb]{\smash{{\SetFigFont{12}{24.0}{\rmdefault}{\mddefault}{\updefault}{\color[rgb]{0,0,0}$\mod(F \wedge \neg y_1)$}%
}}}}
\put(6076,-2536){\makebox(0,0)[b]{\smash{{\SetFigFont{12}{24.0}{\rmdefault}{\mddefault}{\updefault}{\color[rgb]{0,0,0}$\emptyset \rev(F) \rev(y_1)$}%
}}}}
\put(4996,-2536){\makebox(0,0)[b]{\smash{{\SetFigFont{12}{24.0}{\rmdefault}{\mddefault}{\updefault}{\color[rgb]{0,0,0}$\emptyset \rev(F)$}%
}}}}
\put(3916,-2536){\makebox(0,0)[b]{\smash{{\SetFigFont{12}{24.0}{\rmdefault}{\mddefault}{\updefault}{\color[rgb]{0,0,0}$\emptyset$}%
}}}}
\end{picture}%
\nop{
      0               0 rev(F)         0 rev(F)rev(y1)
+-----------+      +-----------+        +----------+
|           |      |  mod(F)   |        |mod(F&y1) |
+-----------+      +-----------+        +----------+
                   |  mod(-F)  |        |mod(F&-y1)|
                   +-----------+        +----------+
                                        |   ...    |
}

The empty ordering has all models in class zero. A revision operator cuts
$\mod(F)$ out from it to make the new class zero when revising by a consistent
formula $F$. The class $\mod(\neg F)$ is created only if $F$ is not
tautological, but this is irrelevant.

The new class zero $\mod(F)$ is refined into $\mod(F \wedge y_1)$ and $\mod(F
\wedge \neg y_1)$ because of the bottom-refining condition: since $F$ is
consistent and does not contain $y_1$, the conjunctions $F \wedge y_1$ and $F
\wedge \neg y_1$ are consistent.

Revising this ordering by $y_1 \rightarrow x_1$ depends on the consistency of
$F \wedge x_1$. If $F$ is consistent with $x_1$ then $F \wedge y_1 \wedge (y_1
\rightarrow x_1)$ is consistent; therefore, class zero $\mod(F \wedge y_1)$
contains some models of $y_1 \rightarrow x_1$. The resulting class zero
comprises them:
{} $\mod(F \wedge y_1 \wedge (y_1 \rightarrow x_1)) =
{}  \mod(F \wedge x_1 \wedge y_1)$.
If $F$ is inconsistent with $x_1$, then $F \wedge y_1 \wedge(y_1 \rightarrow
x_1)$ is inconsistent; therefore, $\mod(F \wedge y_1)$ does not contain any
model of $y_1 \rightarrow x_1$. Instead, $\mod(F \wedge \neg y_1)$ does, since
{} $\mod(F \wedge \neg y_1 \wedge (y_1 \rightarrow x_1)$
is the same as
{} $\mod(F \wedge \neg y_1)$,
which is consistent because $F$ is consistent and does not mention $y_1$. The
class zero resulting from revising by $y_1 \rightarrow x_1$ is
{} $\mod(F \wedge \neg y_1 \wedge (y_1 \rightarrow x_1)) =
{}  \mod(F \wedge \neg y_1)$
since $\rev$ is a revision.

This proves that the result of revising first by $y_1$ and then by $y_1
\rightarrow x_1$ is an ordering that has $\mod(F \wedge x_1 \wedge y_1)$ as its
class zero if $F \wedge x_1$ is consistent and $\mod(F \wedge \neg y_1)$
otherwise. Apart from $y_1$, which is unlinked to the rest of the formula and
the other variables $y_i$, these are the models of $F \wedge x_1$ if consistent
and the models of $F$ otherwise.

Iterating the procedure on the remaining variables $x_2,\ldots,x_n$ produces an
ordering whose class zero comprises the lexicographically maximally model of
$F$ only. Checking whether it entails $x_n$ is the final step of the
translation.

\

This intuition is made a formal proof by induction. For every $i$, class zero
of
{} $\emptyset \rev(F) \rev(y_1) \rev(y_1 \rightarrow x_1) \ldots
{}                    \rev(y_i) \rev(y_i \rightarrow x_i)$
is
{} $\mod(\bigwedge Y' \wedge \maxset(F,x_1,\ldots,x_i))$
{} for some consistent $Y' \subseteq \{y_1,\neg y_1,\ldots,y_i,\neg y_i\}$.
This is the lexicographically maximal partial model over variables
$x_1,\ldots,x_i$, apart from some of the variables $y_1,\ldots,y_i$. Assuming
that this condition is true for $i$, it is shown to remain true after revising
by $y_{i+1}$ and $y_{i+1} \rightarrow x_{i+1}$.

Let
{} $C=\emptyset \rev(F) \rev(y_1) \rev(y_1 \rightarrow x_1) \ldots \rev(y_i)
{}                      \rev(y_i \rightarrow x_i)$
and
{} $M=\bigwedge Y' \wedge \maxset(F,x_1,\ldots,x_i)$.
The inductive assumption is $C(0)=\mod(M)$. Since $M$ does not mention
$y_{i+1}$ and is consistent, $M \wedge y_{i+1}$ is consistent. As a result,
$C(0) \cap \mod(y_{i+1})$ is not empty. A bottom-refining revision splits the
class zero in two:

\begin{eqnarray*}
C \rev(y_{i+1}) (0) &=& \mod(M \wedge y_{i+1}) \\
C \rev(y_{i+1}) (1) &=& \mod(M \wedge \neg y_{i+1})
\end{eqnarray*}

This ordering is further revised by $y_{i+1} \rightarrow x_{i+1}$. The
resulting class zero of $C \rev(y_{i+1}) \rev(y_{i+1} \rightarrow x_{i+1})$ is
the first non-empty of the following two sets, since $\rev$ is by assumption a
revision operator and at least the second is not empty since $M$ is consistent
and does not contain $y_{i+1}$.

\begin{eqnarray*}
C \rev(y_{i+1})(0) \cap \mod(y_{i+1} \rightarrow x_{i+1})
&=&
\mod(M \wedge y_{i+1} \wedge (y_{i+1} \rightarrow x_{i+1}))	\\
&=&
\mod(M \wedge y_{i+1} \wedge x_{i+1})				\\
C \rev(y_{i+1})(1) \cap \mod(y_{i+1} \rightarrow x_{i+1})
&=&
\mod(M \wedge \neg y_{i+1} \wedge (y_{i+1} \rightarrow x_{i+1})) \\
&=&
\mod(M \wedge \neg y_{i+1}) \\
\end{eqnarray*}

Since $M$ is $\bigwedge Y' \wedge \maxset(F,x_1,\ldots,x_i)$ and $F$ is a
formula over variables $\{x_1,\ldots,x_n\}$ only, $M \wedge y_{i+1} \wedge
x_{i+1}$ is consistent if and only if $\maxset(F,x_1,\ldots,x_i) \wedge
x_{i+1}$ is consistent. Depending on this condition:

\begin{description}

\item[$\maxset(F,x_1,\ldots,x_i) \wedge x_{i+1}$ is consistent:] $M \wedge
y_{i+1} \wedge x_{i+1}$ is also consistent; therefore, the models of this
formula are the new class zero; $\bigwedge Y' \wedge \maxset(F,x_1,\ldots,x_i)
\wedge y_{i+1} \wedge x_{i+1}$ is the same as $\bigwedge Y'' \wedge
\maxset(F,x_1,\ldots,x_{i+1})$ where $Y'' = Y' \cup \{y_{i+1}\}$ since by
assumption $\maxset(F,x_1,\ldots,x_i) \wedge x_{i+1}$ is consistent;

\item[$\maxset(F,x_1,\ldots,x_i) \wedge x_{i+1}$ is inconsistent:] since $M
\wedge y_{i+1} \wedge x_{i+1}$ is inconsistent, the first non-empty of the two
sets above is $\mod(M \wedge \neg y_{i+1})$; replacing $M$ with its definition,
$M \wedge y_{i+1}$ becomes $\bigwedge Y' \wedge \maxset(F,x_1,\ldots,x_i)
\wedge \neg y_{i+1}$ and this is equal to $\bigwedge Y'' \wedge
\maxset(F,x_1,\ldots,x_i,x_{i+1})$ where $Y'' = Y' \cup \{\neg y_i\}$. Indeed,
$\maxset(F,x_1,\ldots,x_i,x_{i+1})=\maxset(F,x_1,\ldots,x_i)$ since by
assumption $\maxset(F,x_1,\ldots,x_i) \wedge x_{i+1}$ is inconsistent.

\end{description}~\qed

\subsection{Very radical revision}

Very radical revision is neither lexicographic-finding nor bottom-refining. It
is indeed easier, as the classes of $\emptyset \rad(R_1) \ldots \rad(R_n)$ are
relatively easy to determine.

\begin{theorem}

For every formulae $R_1,\ldots,R_n$, the total preorder $\emptyset \rad(R_1)
\ldots \rad(R_n)$ is equivalent to the following preorder $C$.

\begin{eqnarray*}
C(0)	&=& \mod(\neg \bot \wedge R_1 \wedge R_2 \wedge
                 R_3 \wedge \cdots \wedge R_n)				\\
C(1)	&=& \mod(\neg R_1 \wedge R_2 \wedge R_3 \wedge \cdots \wedge R_n) \\
C(2)	&=& \mod(\neg R_2 \wedge R_3 \wedge \cdots \wedge R_n)		\\
& \vdots &								\\
C(n-1)	&=& \mod(\neg R_{n-1} \wedge R_n)				\\	
C(n)	&=& \mod(\neg R_n)
\end{eqnarray*}

\end{theorem}

\proof The proof is by induction on the length of the sequence. For $n=1$, the
total preorder $C=\emptyset \rad(R_1)$ splits the single class
$\emptyset(0)=\mod(\top)$ into the two classes $C(0)=\mod(\top) \cap
\mod(R_1)=\mod(\neg \bot \wedge R_1)$ and $C(1)=\mod(\top) \backslash
\mod(R_1)=\mod(\neg R_1)$. The claim therefore holds.

Assuming that the claim holds for the preorder $C=[C(0),\ldots,C(n-1)] =
\emptyset \rad(R_1) \ldots \rad(R_{n-1})$, it is proved for $C \rad(R_n)$. From
the definition of $\rad$:

\begin{eqnarray*}
\lefteqn{[C(0),\ldots,C(m)] \rad(R_n) = }			\\
&=& [
C(0) \cap \mod(R_n),
\ldots,
C(n-1) \cap \mod(R_n),
(C(0) \cup \cdots \cup C(n-1)) \backslash \mod(R_n)
]								\\
&=& [
\mod(\neg \bot \wedge R_1 \wedge \cdots \wedge R_{n-1}) \cap \mod(R_n), \\
&&~
\ldots,
\mod(\neg R_{n-1}) \cap \mod(R_n),
\mod(\top) \backslash \mod(R_n)
]								\\
&=& [
\mod(\neg \bot \wedge R_1 \wedge \cdots \wedge R_{n-1} \wedge R_n),
\ldots,
\mod(\neg R_{n-1} \wedge R_n),
\mod(\neg R_n)
]
\end{eqnarray*}

The second equality holds because by definition $C$ includes all models;
therefore, $C(0) \cup \cdots \cup C(n)=\mod(\top)$. The third holds because
$\mod(\top) \backslash \mod(R_n)$ is the set of all models but the ones of
$R_n$.~\qed

This theorem tells how to determine $\min(\emptyset \rad(R_1) \ldots
\rad(R_n),\top)$: by conjoining $R_n$ with $R_{n-1}$, then with $R_{n-2}$ and
so on until consistent.

\begin{definition}

The longest consistent conjunction of a sequence of formulae
$\longest(L_1,\ldots,L_n)$ is $L_1 \wedge \cdots \wedge L_i$ such that either
$i=n$ or $L_1 \wedge \cdots \wedge L_i \wedge L_{i+1}$ is inconsistent.

\end{definition}

A longest sequence is a simplified form of maxset: the maxsets conjoin formulae
in order skipping every one that would create an inconsistency; the longest
sequences stop altogether at the first. A sequence of very radical revisions
from the empty ordering can be reformulated in terms of this definition.

\begin{theorem}

For every formulae $R_1,\ldots,R_n$, formula $\form(\min(\emptyset \rad(R_1)
\ldots \rad(R_n),\top))$ is equivalent to $\longest(R_n,\ldots,R_1)$.

\end{theorem}

\proof The previous theorem shows that the classes of $\emptyset \rad(R_1)
\ldots \rad(R_n)$ are the models of the following formulae:

\begin{eqnarray*}
&& \neg \bot \wedge R_1 \wedge R_2 \wedge R_3 \wedge \cdots \wedge R_n \\
&& \neg R_1 \wedge R_2 \wedge R_3 \wedge \cdots \wedge R_n		\\
&& \neg R_2 \wedge R_3 \wedge \cdots \wedge R_n				\\
&& \vdots								\\
&& \neg R_{n-1} \wedge R_n						\\
&& \neg R_n
\end{eqnarray*}

As a result, $\min(\emptyset \rad(R_1) \ldots \rad(R_n),\top)$ is the set of
models of the first consistent formula in the list. This formula may the first
or any of the others. The first is
{} $\neg \bot \wedge R_1 \wedge R_2 \wedge R_3 \wedge \cdots \wedge R_n$,
which is equivalent to
{} $R_1 \wedge \cdots \wedge R_n$.
If it is consistent, then
{} $R_n \wedge \cdots \wedge R_i$
is consistent for $i=1$, and is therefore the same as
$\longest(R_n,\ldots,R_1)$.

The other case is that the first consistent formula of the list is
{} $\neg R_{i-1} \wedge R_i \wedge \cdots \wedge R_n$
for some index $i$. Since it is consistent, its subformula
{} $R_i \wedge \cdots \wedge R_n$
is consistent too. This is the first part of the definition of the longest
consistent conjunction, the second being the inconsistency of
{} $R_{i-1} \wedge R_i \wedge \cdots \wedge R_n$.

To the contrary, let $M$ be a model of
{} $R_{i-1} \wedge R_i \wedge \cdots \wedge R_n$.
If $M$ satisfies all formulae $R_1,\ldots,R_{i-2}$ then $R_1 \wedge \cdots
\wedge R_n \wedge \neg \bot$ is consistent, contrary to assumption. Therefore,
$M$ falsifies some formula among $R_1,\ldots,R_{i-2}$. Let $j \leq i-2$ be the
highest index such that $M$ falsifies $R_j$. Since $M$ falsifies this formula,
it satisfies its negation $\neg R_j$. Because of the highest index, $M$
satisfies all formulae $R_{j+1},\ldots,R_{i-2}$ if any. As a result, $M$
satisfies
{} $\neg R_j \wedge R_{j+1} \wedge \cdots \wedge R_{i-2}$.
Since it also satisfies
{} $R_{i-1} \wedge R_i \wedge \cdots \wedge R_n$,
it satisfies
{} $\neg R_j \wedge R_{j+1} \wedge \cdots \wedge R_1$.
The consistency of this sequence with $j \leq i-2$ contradicts the assumption
that $i$ is the lowest index such that
{} $\neg R_{i-1} \wedge R_i \wedge \cdots \wedge R_1$
is consistent.

This proves that
{} $R_{i-1} \wedge R_i \wedge \cdots \wedge R_n$
is inconsistent. Since
{} $R_i \wedge \cdots \wedge R_n$
is consistent, this is $\longest(R_n,\ldots,R_1)$.~\qed

By this theorem, the complexity of $\longest(L_1,\ldots,L_n) \models Q$ is the
same as inference from a sequence of very radical revisions from an empty
sequence. This problem is investigated under the condition that each formula
$L_i$ is consistent.

\begin{theorem}

For every sequence of consistent formulae $L_1,\ldots,L_n$ and formula $Q$,
checking whether $\longest(L_1,\ldots,L_n) \models Q$ is \Dlog{2}-complete, and 
\bh{2n-1}-complete if $n$ is a constant.

\end{theorem}

\proof Entailment $\longest(L_1,\ldots,L_n) \models Q$ holds if $L_1 \wedge
\cdots \wedge L_i$ is consistent and entails $Q$ for some $i \in
\{1,\ldots,n\}$. The check for the inconsistency of $L_1 \wedge \cdots \wedge
L_i \wedge L_{i+1}$ is not necessary: if it does not hold, then $L_{i+1}$ is
added to the conjunction $L_1 \wedge \cdots \wedge L_i \models Q$, and the
result still entails $Q$.

These consistency and entailment tests can be done in parallel; if they succeed
for the same index $i$, then $\longest(L_1,\ldots,L_n)$ entails $Q$. The
problem is therefore in \Dlog{2}. If $n$ is a constant, an exact computation of
the tests to be performed is needed; these are:

\begin{enumerate}

\item $\models Q$; or

\item $L_1$ is consistent and $L_1 \models Q$; or

\item $L_1 \wedge L_2$ is consistent and $L_1 \wedge L_2 \models Q$; or

\item[] \vdots

\item[$n+1$] $L_1 \wedge \cdots \wedge L_n$ is consistent and entails $Q$.

\end{enumerate}

Under the assumption that every single $L_i$ is consistent, the first two ones
can be simplified. Indeed, the second one ``$L_1$ is consistent and $L_1 \models
Q$'' is the same as $L_1 \models Q$. This condition is entailed by $\models Q$,
which becomes unnecessary. The conditions can therefore be rewritten as:

\begin{enumerate}

\item $L_1 \models Q$; or

\item $L_1 \wedge L_2$ is consistent and $L_1 \wedge L_2 \models Q$; or

\item[] \vdots

\item[$n$] $L_1 \wedge \cdots \wedge L_n$ is consistent and entails $Q$.

\end{enumerate}

The first test is in \conp, the other $n-1$ ones are in \Dp. By the definition
of the Boolean hierarchy~\cite{wagn-87}, the problem is in \bh{2n-1}.

\

Hardness for \Dlog{2} and \bh{2n-1} is proved for unbounded and constant $n$ by
a reduction from the following problem: given $F_1,\ldots,F_{2n-1}$ with $F_i$
consistent implying $F_{i-1}$ consistent, check whether the number of
consistent $F_i$ is even~\cite{eite-gott-97}. This problem is reduced to
checking $\longest(L_1,\ldots,L_n) \models Q$. This implies that entailment
from the longest sequence is both \Dlog{2}-hard in the general case and
\bh{2n-1}-hard if $n$ is constant.

The first step of the reduction is rewriting of each formula $F_i$ on a private
alphabet. This is assumed already done, and does not change consistency.

New variables are introduced, one for each formula: $\{y_1,\ldots,y_m\}$. The
query $Q$ is $\neg y_1 \vee \neg y_3 \vee \cdots$, the disjunction of all
literals $\neg y_i$ with $i$ odd.

The first formula $L_1$ is $y_1 \rightarrow F_1$, which is consistent because
it is satisfied by setting $y_1$ to false. If $F_1$ is inconsistent, $\neg y_1$
is entailed. This is correct since the number of consistent formulae is zero,
which is even.

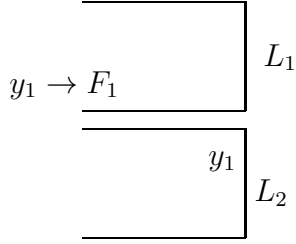
\setlength{\unitlength}{5000sp}%
\begingroup\makeatletter\ifx\SetFigFont\undefined%
\gdef\SetFigFont#1#2#3#4#5{%
  \reset@font\fontsize{#1}{#2pt}%
  \fontfamily{#3}\fontseries{#4}\fontshape{#5}%
  \selectfont}%
\fi\endgroup%
\begin{picture}(1290,1194)(5296,-4573)
\thinlines
{\color[rgb]{0,0,0}\put(5671,-3931){\line( 1, 0){810}}
\put(6481,-3931){\line( 0, 1){540}}
\put(6481,-3391){\line(-1, 0){810}}
}%
{\color[rgb]{0,0,0}\put(5671,-4561){\line( 1, 0){810}}
\put(6481,-4561){\line( 0, 1){540}}
\put(6481,-4021){\line(-1, 0){810}}
}%
\put(6571,-3706){\makebox(0,0)[lb]{\smash{{\SetFigFont{12}{24.0}{\rmdefault}{\mddefault}{\updefault}{\color[rgb]{0,0,0}$L_1$}%
}}}}
\put(6526,-4381){\makebox(0,0)[lb]{\smash{{\SetFigFont{12}{24.0}{\rmdefault}{\mddefault}{\updefault}{\color[rgb]{0,0,0}$L_2$}%
}}}}
\put(6436,-4201){\makebox(0,0)[rb]{\smash{{\SetFigFont{12}{24.0}{\rmdefault}{\mddefault}{\updefault}{\color[rgb]{0,0,0}$y_1$}%
}}}}
\put(5311,-3841){\makebox(0,0)[lb]{\smash{{\SetFigFont{12}{24.0}{\rmdefault}{\mddefault}{\updefault}{\color[rgb]{0,0,0}$y_1 \rightarrow F_1$}%
}}}}
\end{picture}%
\nop{
+--------------------+
|                    |
|   y1 -> F1         |    L1
+--------------------+
|               y1   |
|               ...  |    L2
}

The second formula $L_2$ contains $y_1$ and an unrelated variable. The
conjunction of the first two formulae contains
{} $(y_1 \rightarrow F_1) \wedge y_1 \equiv F_1 \wedge y_1$.
If $F_1$ is consistent, this formula does not entail $\neg y_1$ and is
consistent. The construction of the longest consistent sequence continues.

Overall, if $F_1$ is inconsistent then $\neg y_1$ is entailed. Otherwise, $\neg
y_1$ is not entailed and the construction of the longest consistent sequence
continues.

The rest of the sequence works similarly: the construction stops at the first
inconsistent formula $F_i$; if $i$ is odd then $\neg y_i$ is entailed, which is
correct since the number of consistent formulae is $i-1$, even.

The other formulae are as follows, where $i$ is odd.

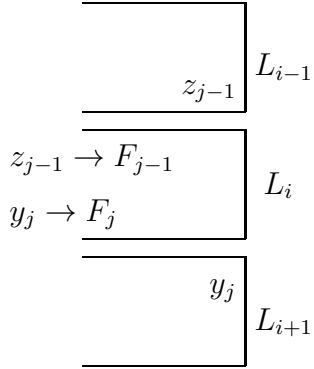
\setlength{\unitlength}{5000sp}%
\begingroup\makeatletter\ifx\SetFigFont\undefined%
\gdef\SetFigFont#1#2#3#4#5{%
  \reset@font\fontsize{#1}{#2pt}%
  \fontfamily{#3}\fontseries{#4}\fontshape{#5}%
  \selectfont}%
\fi\endgroup%
\begin{picture}(1290,1824)(5296,-4573)
\thinlines
{\color[rgb]{0,0,0}\put(5671,-3931){\line( 1, 0){810}}
\put(6481,-3931){\line( 0, 1){540}}
\put(6481,-3391){\line(-1, 0){810}}
}%
{\color[rgb]{0,0,0}\put(5671,-4561){\line( 1, 0){810}}
\put(6481,-4561){\line( 0, 1){540}}
\put(6481,-4021){\line(-1, 0){810}}
}%
{\color[rgb]{0,0,0}\put(5671,-3301){\line( 1, 0){810}}
\put(6481,-3301){\line( 0, 1){540}}
\put(6481,-2761){\line(-1, 0){810}}
}%
\put(6436,-3211){\makebox(0,0)[rb]{\smash{{\SetFigFont{12}{24.0}{\rmdefault}{\mddefault}{\updefault}{\color[rgb]{0,0,0}$z_{j-1}$}%
}}}}
\put(6526,-3121){\makebox(0,0)[lb]{\smash{{\SetFigFont{12}{24.0}{\rmdefault}{\mddefault}{\updefault}{\color[rgb]{0,0,0}$L_{i-1}$}%
}}}}
\put(5311,-3571){\makebox(0,0)[lb]{\smash{{\SetFigFont{12}{24.0}{\rmdefault}{\mddefault}{\updefault}{\color[rgb]{0,0,0}$z_{j-1} \rightarrow F_{j-1}$}%
}}}}
\put(6571,-3706){\makebox(0,0)[lb]{\smash{{\SetFigFont{12}{24.0}{\rmdefault}{\mddefault}{\updefault}{\color[rgb]{0,0,0}$L_i$}%
}}}}
\put(5311,-3841){\makebox(0,0)[lb]{\smash{{\SetFigFont{12}{24.0}{\rmdefault}{\mddefault}{\updefault}{\color[rgb]{0,0,0}$y_j \rightarrow F_j$}%
}}}}
\put(6526,-4381){\makebox(0,0)[lb]{\smash{{\SetFigFont{12}{24.0}{\rmdefault}{\mddefault}{\updefault}{\color[rgb]{0,0,0}$L_{i+1}$}%
}}}}
\put(6436,-4201){\makebox(0,0)[rb]{\smash{{\SetFigFont{12}{24.0}{\rmdefault}{\mddefault}{\updefault}{\color[rgb]{0,0,0}$y_j$}%
}}}}
\end{picture}%
\nop{
|               ...  |
|               yi-1 |    Li-1
+--------------------+
| yi-1 -> Fi-1       |
|   yi -> Fi         |    Li
+--------------------+
|               yi   |    Li+1
|               ...  |
}

Assuming that the construction of the longest consistent sequence includes
$L_{i-1}$, it proceeds as follows.

Three cases are possible: $F_{i-1}$ is inconsistent; it is consistent but $F_i$
is not; they are both consistent.

If $F_{i-1}$ is inconsistent, $L_i$ is not added to the sequence because
$L_{i-1} \wedge L_i$ contains $y_{i-1} \wedge (y_{i-1} \rightarrow F_i)$, which
implies the inconsistent formula $F_{i-1}$. Therefore, the longest consistent
sequence does not contain $y_i$, and it therefore does not entail $\neg y_i$.
This is correct since the number of consistent formulae is $i-2$, odd.

If $F_{i-1}$ is consistent, then $L_i$ is added to the sequence because the
final part of its conjunction is
{} $y_{i-1} \wedge (y_{i-1} \rightarrow F_{i-1}) \wedge (y_i \rightarrow F_i)$,
which is equivalent to
{} $y_{i-1} \wedge F_{i-1} \wedge (y_i \rightarrow F_i)$
and is therefore consistent. If $F_i$ is inconsistent this formula entails
$\neg y_i$, which is correct because the number of consistent formulae is
$i-1$, even.

The final case is that $F_i$ is consistent. Not only
{} $y_{i-1} \wedge F_{i-1} \wedge (y_i \rightarrow F_i)$
is consistent, but is also consistent with $L_{i+1}$, which is $y_i$ plus an
unrelated variable. Therefore, the construction of the longest consistent
sequence continues.

\

Technically, the formulae $L_i, L_{i+1}$ and $Q$ are as follows, where $i$ is
odd:

\begin{eqnarray*}
L_1 &=& y_1 \rightarrow F_1						\\
L_2 &=& y_1 \wedge y_2							\\
&\vdots&								\\
L_i &=&	(y_{i-1} \rightarrow F_{i-1}) \wedge (y_i \rightarrow F_i)	\\
L_{i+1} &=& y_i \wedge y_{i+1}						\\
Q &=& \neg y_1 \vee \neg y_3 \vee \cdots
\end{eqnarray*}

Each formula is consistent by itself: the formulae of odd index are satisfied
by the model that assigns false to all variables, the formulae of even index by
that assigning true to all variables.

The claim is that $\longest(L_1,\ldots,L_n)$ entails $Q$ if and only if the
number of consistent formulae $F_i$ is even.

The entailment $\longest(L_1,\ldots,L_m) \models Q$ simplifies because
$\longest(L_1,\ldots,L_m)$ is a conjunction of formulae $L_i$, where the
variables $y_i${\plural} occur in separate subformulae. This means that
$\longest(L_1,\ldots,L_m)$ entails $Q$ if and only if it entails some variables
$\neg y_i$ with $i$ odd.

The number of consistent formulae $F_i$ is even is the same as the consistency
of $F_1 \wedge \cdots \wedge F_{i-1}$ and the inconsistency of $F_1 \wedge
\cdots \wedge F_{i-1} \wedge F_i$ with $i$ odd because of the separation of the
variables and the consistency of all formulae preceding a consistent one.

The conjunction $L_1 \wedge \cdots \wedge L_i$ with $i$ odd is
{} $y_1 \wedge (y_1 \rightarrow F_1) \wedge \cdots
{}  \wedge y_{i-1} \wedge (y_{i-1} \rightarrow F_{i-1})
{}  \wedge (y_i \rightarrow F_i)$.
This formula is equivalent to
{} $y_1 \wedge F_1 \wedge \cdots
{}  \wedge y_{i-1} \wedge F_{i-1}
{}  \wedge (y_i \rightarrow F_i)$.

The conjunction with $i$ even is obtained by taking $L_1 \wedge \cdots \wedge
L_{i+1}$ in the definition above and decreasing $i$ by $1$. It is
{} $y_1 \wedge (y_1 \rightarrow F_1) \wedge \cdots
{}  \wedge y_{i-1} \wedge (y_{i-1} \rightarrow F_{i-1})
{}  \wedge y_i$,
which is equivalent to
{} $y_1 \wedge F_1 \wedge \cdots
{}  \wedge y_{i-1} \wedge F_{i-1}
{}  \wedge y_i$.

The claim can now be proved in each direction.

If $F_1 \wedge \cdots \wedge F_{i-1}$ is consistent and $F_1 \wedge \cdots
\wedge F_{i-1} \wedge F_i$ is not with $i$ odd then
{} $L_1 \wedge \cdots \wedge L_i =
{}  y_1 \wedge F_1 \wedge \cdots
{}  \wedge y_{i-1} \wedge F_{i-1}
{}  \wedge (y_i \rightarrow F_i)$
is consistent and entails $\neg y_i$.
The longest consistent conjunction contains it and therefore entails $\neg y_i$ as
well.

To prove the other direction, the longest consistent conjunction $L_1 \wedge
\cdots \wedge L_i$ is assumed to entail a literal $\neg y_j$. Since this
conjunction is consistent, it does not contain $y_j$. Since it contains all
variables $y_j$ with $j < i$ regardless of whether $i$ is even or odd, the only
possible $j$ is $i$. If $i$ is even then $L_1 \wedge \cdots \wedge L_i$
contains $y_i$. Therefore, $i$ is odd.~\qed

Since a sequence of very radical revisions from the empty ordering is exactly
the same as the longest consistent conjunction, and all formulae are consistent
by assumption, the problem of entailment for very radical revision is
\Dlog{2}-complete in the general case and \bh{2n-1}-complete for constant $n$.

\subsection{Plain severe revision and full meet revision}

On total preorders comprising at most two classes, plain severe and full meet
revision coincide, and always generate an ordering of at most two classes. As a
result, when the initial ordering is empty, sequence of plain severe and full
meet revision coincide:

\[
\emptyset \psev(P_1) \ldots \psev(P_n)
\equiv
\emptyset \full(P_1) \ldots \full(P_n)
\]

More generally, mixed sequences of plain severe and full meet revisions applied
to an ordering comprising at most two classes are equivalent to sequence of
full meet revisions only and to sequences of plain severe revisions only.

These two revisions are neither lexicographic-finding nor bottom-refining.
A lexicographic-finding sequence of revisions $x_1 \wedge x_3, \neg x_1, x_2$
produces $\neg x_1 \wedge x_2 \wedge x_3$, but the same sequence of full meet
revisions instead produces to $\neg x_1 \wedge x_2$.
A sequence of bottom-refining revisions $x_1, x_2$ produces an ordering with a
class one equal to $x_1 \wedge \neg x_2$, but the same sequence of full meet
revisions instead gives $\neg x_1 \vee \neg x_2$. They are also different from
very radical revision, as seen from the sequence of revisions $x_1, x_2, \neg
x_1$, where very radical revision produces $x_2 \wedge \neg x_1$ while full
meet produces $\neg x_1$.

\begin{theorem}

Inference from a sequence of full meet and plain severe revisions applied to
the empty ordering is \Dlog{2}-complete.

\end{theorem}

\proof The \Dlog{2} class includes all problems that can be solved by a
polynomial number of nonadaptive calls to an {\np}-oracle. Nonadaptive means
that no call depends on the others. Equivalently, these calls are in
parallel~\cite{hema-89,buss-hay-91}.

A sequence of full meet revisions
{} $\emptyset \full(S_1) \ldots \full (S_n)$
requires establishing the satisfiability of the following quadratic number of
formulae.

\begin{eqnarray*}
&& S_1						\\
&& S_1 \wedge S_2				\\
&& \vdots					\\
&& S_1 \wedge S_2 \wedge \cdots \wedge S_n	\\
&& S_2						\\
&& S_2 \wedge S_3				\\
&& \vdots					\\
&& S_2 \wedge S_3 \wedge \cdots \wedge S_n	\\
&& \vdots					\\
&& S_n
\end{eqnarray*}

The first group of formulae starts with $S_1$ and adds a formula at time until
the last. The second starts with $S_2$ and does the same. This is repeated for
all subsequent formulae $S_3,\ldots,S_n$. All these conjunctions are checked
for satisfiability regardless of the satisfiability of the others.

Given the result of these tests, the result of full meet revision is calculated
in polynomial time. First, the longest continuous conjunction $F_1 \wedge
\cdots \wedge F_i$ is determined. If $i=n$, it is the final result. Otherwise,
since $i < n$ then $i+1$ is less than or equal to $n$. Therefore, $F_{i+1}$ is
a formula of the sequence. The longest continuous conjunction $F_{i+1} \wedge
\cdots \wedge F_j$ is again determined. If $j = n$, it is the final result.
Otherwise, the process continues with $F_{j+1}$. This is repeated until $F_n$
is in the conjunction.

Hardness was announced for full meet revision in a previous article, but
without proof~\cite{libe-97-c}. The proof provided here is by reduction from
the problem of deciding $\longest(L_1,\ldots,L_n) \models Q$. Given the
consistent formulae $L_1,\ldots,L_n$, the reduction builds the following
sequence.

\begin{eqnarray*}
F_n &=& a_n \rightarrow (L_1 \wedge \cdots \wedge L_n)		\\
F_{n-1} &=&
	a_n \wedge (a_{n-1} \rightarrow (L_1 \wedge \cdots \wedge L_{n-1})) \\
F_{n-2} &=&
	a_{n-1} \wedge (a_{n-2}\rightarrow(L_1 \wedge \cdots \wedge L_{n-2}))\\
& \vdots &							\\
F_2 &=& a_3 \wedge (a_2 \rightarrow (L_1 \wedge L_2)		\\
F_1 &=& a_2 \wedge L_1
\end{eqnarray*}

The sequence of revisions $F_n,\ldots,F_1$ applied to the empty order entails
$Q$ if and only if $\longest(L_1,\ldots,L_n)$ does. This is the case because an
inconsistent conjunction $L_1 \wedge \cdots \wedge L_i$ makes $F_i$
inconsistent with the following formulae $F_{i+1}$, which therefore takes its
place. Otherwise, they are conjoined and the process continues.

Technically, if $L_1 \wedge \cdots \wedge L_i$ is the longest consistent
conjunction then $L_1 \wedge \cdots \wedge L_{i+1}$ is inconsistent. Since
$F_{i+1}$ contains $a_{i+1} \rightarrow (L_1 \wedge \cdots \wedge L_{i+1})$, it
is inconsistent with $a_{i+1}$, which is contained in $F_i$. As a result, full
meet revision produces $F_i$. The subsequent formulae $F_{i-1},\ldots,F_1$ are
consistent with $F_i$. Indeed, $F_i \wedge F_{i+1}$ implies $L_1 \wedge \cdots
\wedge L_i$, which is consistent and entails all implications $a_j \rightarrow
(L_1 \wedge \cdots \wedge L_j)$ in the following formulae. What remains after
their removal is only a number of positive literals $a_j$, which are therefore
consistent.

This proves that the result of the sequence of revision is $F_i \wedge \cdots
\wedge F_1$, which is equivalent to the longest consistent conjunction $L_1
\wedge \cdots \wedge L_i$ apart from some unrelated variables $a_j$. Entailment
of $Q$ is therefore the same.~\qed

\section{Conclusions}
\label{conclusions}

This article advocates and studies mixed sequences of belief change operators,
in which revisions, refinements and withdrawals may occur. With some
exceptions~\cite{lehm-95,koni-pere-00,gabb-etal-03,delg-dubo-lang-06,%
arav-etal-19,boot-chan-20}, the semantics for iterated belief revision mostly
work on objects that are equivalent to total preorders, which lets using
different kinds of changes at different times. Even the memoryless operators
such as full meet revision~\cite{alch-gard-maki-85} and the distance-based
revision~\cite{dala-88,pepp-will-16} can be embedded in this framework: they
produce a plausibility order which does not depend at all on the previous one
except for their zero class.

The main technical result of this article is a method for computing the result
of a mixed sequence of revisions. It directly works on sequences of
lexicographic revisions, refinements and severe antiwithdrawals, which may
result from translating an arbitrary sequence of lexicographic revisions,
refinements, severe withdrawal, natural, severe, plain severe, moderate severe
and very radical revisions, alternating in every possible way. The requirement
of being able to solve propositional satisfiability problems is not too
demanding, given the current efficiency of SAT
algorithms~\cite{bier-etal-09,baly-etal-17} and given that belief revision
cannot be easier than its underlying logical
language~\cite{eite-gott-91-b,libe-scha-00}. The polynomial running time (not
counting the satisfiability tests) implies that the required amount of memory
is also polynomial, as well as the resulting knowledge bases at each step. This
was not obvious, as some belief change operators may produce orderings
comprising an exponential number of classes, which forbids storing them
explicitly in practice.


\begin{example}

The running example can be solved by an explicit representation of the
preorders.

\[
\emptyset \lex(y) \nat(\neg x) \refi(x \wedge z) \rad(\neg z)
\]

The initial preorder is empty: $\emptyset = [\mod(\top)]$. The revisions change
it as follows, where $\mod()$ is omitted from the classes for simplicity.

\setlength{\unitlength}{5000sp}%
\begingroup\makeatletter\ifx\SetFigFont\undefined%
\gdef\SetFigFont#1#2#3#4#5{%
  \reset@font\fontsize{#1}{#2pt}%
  \fontfamily{#3}\fontseries{#4}\fontshape{#5}%
  \selectfont}%
\fi\endgroup%
\begin{picture}(4704,1218)(5569,-3685)
\thinlines
{\color[rgb]{0,0,0}\put(5581,-2941){\framebox(450,180){}}
}%
{\color[rgb]{0,0,0}\put(6301,-3121){\framebox(450,360){}}
}%
{\color[rgb]{0,0,0}\put(6301,-2941){\line( 1, 0){450}}
}%
{\color[rgb]{0,0,0}\put(7021,-2941){\line( 1, 0){540}}
}%
{\color[rgb]{0,0,0}\put(7021,-3121){\line( 1, 0){540}}
}%
{\color[rgb]{0,0,0}\put(7021,-3301){\framebox(540,540){}}
}%
{\color[rgb]{0,0,0}\put(7831,-3661){\framebox(1080,900){}}
}%
{\color[rgb]{0,0,0}\put(7831,-3481){\line( 1, 0){1080}}
}%
{\color[rgb]{0,0,0}\put(7831,-3301){\line( 1, 0){1080}}
}%
{\color[rgb]{0,0,0}\put(7831,-3121){\line( 1, 0){1080}}
}%
{\color[rgb]{0,0,0}\put(7831,-2941){\line( 1, 0){1080}}
}%
{\color[rgb]{0,0,0}\put(9181,-3481){\framebox(1080,720){}}
}%
{\color[rgb]{0,0,0}\put(9181,-2941){\line( 1, 0){1080}}
}%
{\color[rgb]{0,0,0}\put(9181,-3121){\line( 1, 0){1080}}
}%
{\color[rgb]{0,0,0}\put(9181,-3301){\line( 1, 0){1080}}
}%
\put(6526,-2896){\makebox(0,0)[b]{\smash{{\SetFigFont{12}{24.0}{\rmdefault}{\mddefault}{\updefault}{\color[rgb]{0,0,0}$y$}%
}}}}
\put(8371,-3616){\makebox(0,0)[b]{\smash{{\SetFigFont{12}{24.0}{\rmdefault}{\mddefault}{\updefault}{\color[rgb]{0,0,0}$\neg y \wedge (\neg x \vee \neg z)$}%
}}}}
\put(7291,-2896){\makebox(0,0)[b]{\smash{{\SetFigFont{12}{24.0}{\rmdefault}{\mddefault}{\updefault}{\color[rgb]{0,0,0}$\neg x \wedge y$}%
}}}}
\put(7291,-3076){\makebox(0,0)[b]{\smash{{\SetFigFont{12}{24.0}{\rmdefault}{\mddefault}{\updefault}{\color[rgb]{0,0,0}$x \wedge y$}%
}}}}
\put(7291,-3256){\makebox(0,0)[b]{\smash{{\SetFigFont{12}{24.0}{\rmdefault}{\mddefault}{\updefault}{\color[rgb]{0,0,0}$\neg y$}%
}}}}
\put(8371,-2896){\makebox(0,0)[b]{\smash{{\SetFigFont{12}{24.0}{\rmdefault}{\mddefault}{\updefault}{\color[rgb]{0,0,0}$x \wedge y \wedge z$}%
}}}}
\put(8371,-3076){\makebox(0,0)[b]{\smash{{\SetFigFont{12}{24.0}{\rmdefault}{\mddefault}{\updefault}{\color[rgb]{0,0,0}$\neg x \wedge y$}%
}}}}
\put(8371,-3436){\makebox(0,0)[b]{\smash{{\SetFigFont{12}{24.0}{\rmdefault}{\mddefault}{\updefault}{\color[rgb]{0,0,0}$\neg y \wedge x \wedge z$}%
}}}}
\put(8371,-3256){\makebox(0,0)[b]{\smash{{\SetFigFont{12}{24.0}{\rmdefault}{\mddefault}{\updefault}{\color[rgb]{0,0,0}$x \wedge y \wedge \neg z$}%
}}}}
\put(9721,-3256){\makebox(0,0)[b]{\smash{{\SetFigFont{12}{24.0}{\rmdefault}{\mddefault}{\updefault}{\color[rgb]{0,0,0}$\neg y \wedge \neg z$}%
}}}}
\put(9721,-3436){\makebox(0,0)[b]{\smash{{\SetFigFont{12}{24.0}{\rmdefault}{\mddefault}{\updefault}{\color[rgb]{0,0,0}$\neg z$}%
}}}}
\put(9721,-2896){\makebox(0,0)[b]{\smash{{\SetFigFont{12}{24.0}{\rmdefault}{\mddefault}{\updefault}{\color[rgb]{0,0,0}$\neg x \wedge y \wedge \neg z$}%
}}}}
\put(9721,-3076){\makebox(0,0)[b]{\smash{{\SetFigFont{12}{24.0}{\rmdefault}{\mddefault}{\updefault}{\color[rgb]{0,0,0}$x \wedge y \wedge \neg z$}%
}}}}
\put(5806,-2626){\makebox(0,0)[b]{\smash{{\SetFigFont{12}{24.0}{\rmdefault}{\mddefault}{\updefault}{\color[rgb]{0,0,0}$\emptyset$}%
}}}}
\put(5806,-2896){\makebox(0,0)[b]{\smash{{\SetFigFont{12}{24.0}{\rmdefault}{\mddefault}{\updefault}{\color[rgb]{0,0,0}$\top$}%
}}}}
\put(6526,-3076){\makebox(0,0)[b]{\smash{{\SetFigFont{12}{24.0}{\rmdefault}{\mddefault}{\updefault}{\color[rgb]{0,0,0}$\neg y$}%
}}}}
\put(6526,-2626){\makebox(0,0)[b]{\smash{{\SetFigFont{12}{24.0}{\rmdefault}{\mddefault}{\updefault}{\color[rgb]{0,0,0}$\lex(y)$}%
}}}}
\put(7291,-2626){\makebox(0,0)[b]{\smash{{\SetFigFont{12}{24.0}{\rmdefault}{\mddefault}{\updefault}{\color[rgb]{0,0,0}$\nat(\neg x)$}%
}}}}
\put(8371,-2626){\makebox(0,0)[b]{\smash{{\SetFigFont{12}{24.0}{\rmdefault}{\mddefault}{\updefault}{\color[rgb]{0,0,0}$\res(x \wedge z)$}%
}}}}
\put(9721,-2626){\makebox(0,0)[b]{\smash{{\SetFigFont{12}{24.0}{\rmdefault}{\mddefault}{\updefault}{\color[rgb]{0,0,0}$\rad(\neg z)$}%
}}}}
\end{picture}%
\nop{
 0     lex(y)  nat(-x)     res(xz)      rad(-z)
+---+  +----+  +-----+  +-----------+  +-------+
| T |  |  y |  | -xy |  |    xyz    |  | -xy-z |
+---+  +----+  +-----+  +-----------+  +-------+
       | -y |  |  xy |  |    -xy    |  | xy-z  |
       +----+  +-----+  +-----------+  +-------+
               | -y  |  |   xy-z    |  | -y-z  |
               +-----+  +-----------+  +-------+
                        |   -yxz    |  |  -z   |
                        +-----------+  +-------+
                        | -y(-x|-z) |
                        +-----------+
}

The final result is the same as obtained by the algorithm: the base of the last
preorder is $\neg x \wedge y \wedge \neg z$. However, explicitly storing the
preorder means representing all its classes, which in this example increased in
number up to five. In general, with $n$ variables there may be as many as $2^n$
models, and therefore as many as $2^n$ nonempty classes. In this case, the
bound $2^3=8$ was almost reached after $\res(x \wedge z)$.

\end{example}

A side result is that the resulting knowledge base only takes polynomial space
since it is generated by an algorithm that works in polynomial space. This is
not the case for several one-step revisions~\cite{cado-etal-00-b}. A stricter
characterization can also be given: if none of the original operators is of a
kind that is translated using severe antiwithdrawal, the result is the
conjunction of some formulae in the sequence. Otherwise, it may also contain an
underformula, which by definition may include disjunctions. The ability of
generating results that contain both conjuctions and disjunction can be seen as
informal evidence that mixed sequences of revisions have superior expressive
power~\cite{janh-99} than sequences of a single kind of revisions.

Other articles explored the translations from different belief change operators
into a single formalism. Rott has shown that severe withdrawal, irrevocable and
irrefutable revision can be expressed in terms of revision by
comparison~\cite{rott-06}, natural and lexicographic in terms of bounded
revision~\cite{rott-12}. Several single-step revisions can be recast in some
forms of circumscription~\cite{libe-scha-97}.


Several computational complexity results about belief revision are known. Eiter
and Gottlob~\cite{eite-gott-92-d} proved that most distance-based and syntax
approaches are \P{2}-complete in the single-step case. In a further
article~\cite{eite-gott-96}, the same authors proved (among other results) that
the same applies to positive right-nested counterfactuals, which are equivalent
to a form of iterated revision. Nebel~\cite{nebe-98} proved a number of
results, the most relevant to the present article being the that one-step
syntactic-lexicographic revision is \D{2}-complete. This operator can encode
lexicographic revision as defined in the iterated case by placing each formula
in a separate priority class. Other iterated revisions have a similar degree of
complexity~\cite{libe-97-c}.

A number of problems are left open. The algorithm requires a SAT solver, which
is unavoidable given that the underlying language is propositional logic and
SAT expresses its basic problems of satisfiability, mutual consistency and
entailment. However, some restricted languages such as Horn and Krom require
only polynomial time for checking satisfiability~\cite{scha-78}. As a result,
it makes sense to investigate their computational properties on iterated
change. The analysis would not be obvious because underformulae include both
disjunction and conjunction, which may result in a non-Horn and non-Krom
formula. The Horn restriction has been studied in single-step revisions by
Eiter and Gottlob~\cite{eite-gott-92-c}, and has recently been considered as a
contributor to the semantics of revision~\cite{crei-etal-18}.

Some iterated belief change operators such as radical revision (as opposed to
very radical revision, considered in this article) consider some models
``inaccessible''~\cite{sege-98,ferm-00}. In terms of total preorders, this
amounts to shifting from a partition into ordered classes into a sequence of
non-overlapping subsets; the models that are not in any of them are the
inaccessible one. Alternatively, the highest-level class is given the special
status of inaccessible model container. These operators have not been
considered in this article, but the analysis could be extended to them.

Other operators not considered in this article include the ones based on
numerical rankings~\cite{spoh-88,will-94,jin-thie-07,saue-etal-20} and
bidimensional ones~\cite{cant-97,ferm-rott-04,rott-12}. They allow for
specifying the strength of a revision either by a number or indirectly by
referring to that of another formula. Either way, revision is by a pair of a
formula and an expression of its strength. A preliminary analysis suggests that
at least a form of bidimensional change, revision by comparison, can be recast
in terms of lexicographic and severe antiwithdrawal, at the cost of first
determining an underformula and a maxset of the previous lexicographic
revisions. Other two-place operators may be amenable to such reductions. Other
recent work include iterated
contraction~\cite{koni-pere-17,boot-chan-19,saue-etal-20-b} and operators where
conditions on the result are specified, rather than mandating a mechanism for
obtaining them~\cite{hans-16}.

Memoryless revision operators~\cite{dala-88,sato-88} may be treated as if they
had memory: this is the case of full meet revision, which is indeed oblivious
to the previous history of revision. The ordering it generates is always
$[\mod(K),\mod(\neg K)]$. In spite of its simplicity, it is still useful to
characterize a {\em tabula rasa} step of reasoning, forgetting all previously
acquired data to start over from a single simple information.

Operators with full
memory~\cite{lehm-95,koni-pere-00,gabb-etal-03,delg-dubo-lang-06} require a
different analysis, since they work from the complete history of revisions
rather than from a total preorder that is modified at each step. The same
applies to operators working from structure more complex than total preorders
over models~\cite{arav-etal-19,boot-chan-20}.

Finally, given that a revision may be performed using different operators, a
question is how to decide which. This is related to merging and non-prioritized
revision. An answer may be to use the history of previous revision to find out
the credibility of a source~\cite{libe-16}, which affects the kind of
incorporation. For example, trustworthy sources produce lexicographic
revisions, plausible but not very reliable sources produce natural revisions,
the others refinements. Still better, sources providing information that turned
out to be valid after all subsequent changes are better treated by
lexicographic revisions; source providing information that turned out to be
specific to the current case are formalized by natural revision. As an
alternative, every new information may be initially treated as a natural
revision; if observations suggest its generality, they are promoted to
lexicographic.

\begin{example}[cont.]

Sound of feathers. A bird, after all?

The hunter and the policeman turn their head, eager to find out. What comes out
from the bushes is a drag queen in red feathers, who stopped by the thicket for
the obvious reason while coming for the parade at the village f\^ete. Not a
bird ($\neg b$) but red ($r$), not to be hunted anyway ($\neg h$).

\end{example}


\bibliographystyle{plain}

\begin{thebibliography}{10}

\bibitem{alch-gard-maki-85}
C.~E. Alchourr\'on, P.~G{\"a}rdenfors, and D.~Makinson.
\newblock On the logic of theory change: Partial meet contraction and revision
  functions.
\newblock {\em Journal of Symbolic Logic}, 50:510--530, 1985.

\bibitem{baly-etal-17}
T.~Balyo, M.J.H. Heule, and M.~J{\"{a}}rvisalo.
\newblock {SAT} competition 2016: Recent developments.
\newblock In {\em Proceedings of the Thirdy-First AAAI Conference on Artificial
  Intelligence (AAAI~2017)}, pages 5061--5063. {AAAI} Press/The {MIT} Press,
  2017.

\bibitem{benf-etal-01}
S.~Benferhat, D.~Dubois, and H.~Prade.
\newblock A computational model for belief change and fusing ordered belief
  bases.
\newblock In M.~Williams and H.~Rott, editors, {\em Frontiers in Belief
  Revision}, pages 109--134. Springer, 2001.

\bibitem{bier-etal-09}
A.~Biere, M.J.M. Heule, H.~van Maaren, and T.~Walsh.
\newblock {\em Handbook of satisfiability}.
\newblock IOS Press, 2009.

\bibitem{boot-chan-19}
R.~Booth and J.~Chandler.
\newblock From iterated revision to iterated contraction: Extending the harper
  identity.
\newblock {\em Artificial Intelligence}, 277, 2019.

\bibitem{boot-chan-20}
R.~Booth and J.~Chandler.
\newblock On strengthening the logic of iterated belief revision: Proper
  ordinal interval operators.
\newblock {\em Artificial Intelligence}, 285:103--289, 2020.

\bibitem{boot-meye-06}
R.~Booth and T.~Meyer.
\newblock Admissible and restrained revision.
\newblock {\em Journal of Artificial Intelligence Research}, 26:127--151, 2006.

\bibitem{boot-nitt-08}
R.~Booth and A.~Nittka.
\newblock Reconstructing an agent's epistemic state from observations about its
  beliefs and non-beliefs.
\newblock {\em Journal of Logic and Computation}, 18:755--782, 2008.

\bibitem{boot-etal-06}
R.~Booth, Meyer T.A., and K.~Wong.
\newblock A bad day surfing is better than a good day working: How to revise a
  total preorder.
\newblock In {\em Proceedings of the Tenth International Conference on
  Principles of Knowledge Representation and Reasoning (KR~2006)}, pages
  230--238. {AAAI} Press/The {MIT} Press, 2006.

\bibitem{bout-96-a}
C.~Boutilier.
\newblock Iterated revision and minimal change of conditional beliefs.
\newblock {\em Journal of Philosophical Logic}, 25(3):263--305, 1996.

\bibitem{buss-hay-91}
S.R. Buss and L.~Hay.
\newblock On truth-table reducibility to {SAT}.
\newblock {\em Information and Computation}, 91(1):86--102, 1991.

\bibitem{cado-etal-00-b}
M.~Cadoli, F.M. Donini, P.~Liberatore, and M.~Schaerf.
\newblock Space efficiency of propositional knowledge representation
  formalisms.
\newblock {\em Journal of Artificial Intelligence Research}, 13(1):1--31, 2000.

\bibitem{cant-97}
J.~Cantwell.
\newblock On the logic of small changes in hypertheories.
\newblock {\em Theoria}, 63(1-2):54--89, 1997.

\bibitem{crei-etal-18}
N.~Creignou, R.~Ktari, and O.~Papini.
\newblock Belief update within propositional fragments.
\newblock {\em Journal of Artificial Intelligence Research}, 61:807--834, 2018.

\bibitem{dala-88}
M.~Dalal.
\newblock Investigations into a theory of knowledge base revision: Preliminary
  report.
\newblock In {\em Proceedings of the Seventh National Conference on Artificial
  Intelligence (AAAI'88)}, pages 475--479, 1988.

\bibitem{darw-pear-97}
A.~Darwiche and J.~Pearl.
\newblock On the logic of iterated belief revision.
\newblock {\em Artificial Intelligence}, 89(1--2):1--29, 1997.

\bibitem{delg-dubo-lang-06}
J.P. Delgrande, D.~Dubois, and J.~Lang.
\newblock Iterated revision as prioritized merging.
\newblock In {\em Proceedings of the Tenth International Conference on
  Principles of Knowledge Representation and Reasoning (KR~2006)}, pages
  210--220, 2006.

\bibitem{eite-gott-92-c}
T.~Eiter and G.~Gottlob.
\newblock Complexity results for disjunctive logic programming and application
  to nonmonotonic logics.
\newblock Technical Report CD-TR 92/41, Technische {Universit\"at} Wien, Vienna
  Austria, Christian Doppler Labor {f\"{u}r} Expertensysteme, 1992.

\bibitem{eite-gott-91-b}
T.~Eiter and G.~Gottlob.
\newblock On the complexity of propositional knowledge base revision, updates
  and counterfactuals.
\newblock {\em Artificial Intelligence}, 57:227--270, 1992.

\bibitem{eite-gott-92-d}
T.~Eiter and G.~Gottlob.
\newblock On the complexity of propositional knowledge base revision, updates
  and counterfactuals.
\newblock {\em Artificial Intelligence}, 57:227--270, 1992.

\bibitem{eite-gott-96}
T.~Eiter and G.~Gottlob.
\newblock The complexity of nested counterfactuals and iterated knowledge base
  revisions.
\newblock {\em Journal of Computer and System Sciences}, 53(3):497--512, 1996.

\bibitem{eite-gott-97}
T.~Eiter and G.~Gottlob.
\newblock The complexity class $\theta_2^p$: Recent results and applications in
  {AI} and modal logic.
\newblock In {\em Proceedings of the Eleventh International Symposium on
  Fundamentals of Computer Theory, (FCT'97)}, pages 1--18. Springer, 1997.

\bibitem{ferm-reis-13}
E.~Ferm{\'e} and M.D.L. Reis.
\newblock Epistemic entrenchment-based multiple contractions.
\newblock {\em Review of Symbolic Logic}, 6(3):460--487, 2013.

\bibitem{ferm-rodr-98}
E.~Ferm{\'e} and R.~Rodriguez.
\newblock A brief note about {R}ott contraction.
\newblock {\em Journal of the Interest Group in Pure and Applied Logic},
  6(6):835--842, 1998.

\bibitem{ferm-rott-04}
E.~Ferm{\'e} and H.~Rott.
\newblock Revision by comparison.
\newblock {\em Artificial Intelligence}, 157(1):5--47, 2004.

\bibitem{ferm-00}
E.L. Ferm{\'{e}}.
\newblock Irrevocable belief revision and epistemic entrenchment.
\newblock {\em Journal of the Interest Group in Pure and Applied Logic},
  8(5):645--652, 2000.

\bibitem{gabb-etal-03}
D.M. Gabbay, G.~Pigozzi, and J.~Woods.
\newblock Controlled revision - an algorithmic approach for belief revision.
\newblock {\em Journal of Logic and Computation}, 13(1):3--22, 2003.

\bibitem{gard-88}
P.~G{\"a}rdenfors.
\newblock {\em Knowledge in Flux: Modeling the Dynamics of Epistemic States}.
\newblock Bradford Books, MIT Press, Cambridge, MA, 1988.

\bibitem{gard-maki-88}
P.~G{\"a}rdenfors and D.~Makinson.
\newblock Revision of knowledge systems using epistemic entrenchment.
\newblock In {\em Proceedings of the Second Conference on Theoretical Aspects
  of Reasoning about Knowledge (TARK'88)}, pages 83--95, 1988.

\bibitem{girl-etal-17}
M.~Girlando, B.~Lellmann, N.~Olivetti, and G.L. Pozzato.
\newblock Hypersequent calculi for lewis' conditional logics with uniformity
  and reflexivity.
\newblock In {\em International Conference on Automated Reasoning with Analytic
  Tableaux and Related Methods}, pages 131--148, 2017.

\bibitem{glai-00}
S.M. Glaister.
\newblock Recovery recovered.
\newblock {\em Journal of Philosophical Logic}, 29(2):171--206, 2000.

\bibitem{grov-88}
A.~Grove.
\newblock Two modellings for theory change.
\newblock {\em Journal of Philosophical Logic}, pages 157--170, 1988.

\bibitem{hans-11}
S.O. Hanson.
\newblock Logic of belief revision.
\newblock In E.N. Zalta, editor, {\em The Stanford Encyclopedia of Philosophy}.
  Metaphysics Research Lab, Stanford University, 2011.

\bibitem{hans-16}
S.O. Hansson.
\newblock Iterated descriptor revision and the logic of ramsey test
  conditionals.
\newblock {\em Journal of Philosophical Logic}, 45(4):429--450, 2016.

\bibitem{hema-89}
L.A. Hemachandra.
\newblock The strong exponential hierarchy collapses.
\newblock {\em Journal of Computer and System Sciences}, 39(3):299--322, 1989.

\bibitem{arav-etal-19}
T.I. I.~Aravanis, P.~Peppas, and M.{-}A. Williams.
\newblock Observations on {D}arwiche and {P}earl's approach for iterated belief
  revision.
\newblock In {\em Proceedings of the Twenty-Eighth International Joint
  Conference on Artificial Intelligence (IJCAI~2019)}, pages 1509--1515, 2019.

\bibitem{janh-99}
T.~Janhunen.
\newblock On the intertranslatability of non-monotonic logics.
\newblock {\em Annals of Mathematics and Artificial Intelligence},
  27(1-4):79--128, 1999.

\bibitem{jin-thie-07}
Y.~Jin and M.~Thielscher.
\newblock Iterated belief revision, revised.
\newblock {\em Artificial Intelligence}, 171(1):1--18, 2007.

\bibitem{kats-mend-91-b}
H.~Katsuno and A.~O. Mendelzon.
\newblock Propositional knowledge base revision and minimal change.
\newblock {\em Artificial Intelligence}, 52:263--294, 1991.

\bibitem{kern-etal-21}
G.~Kern-Isberner, N.~Skovgaard-Olsen, and W.~Spohn.
\newblock {\em Ranking Theory}, chapter 5.3.
\newblock The {MIT} Press, 2021.

\bibitem{koni-98}
S.~Konieczny.
\newblock Operators with memory for iterated revision.
\newblock Technical Report IT-314, Laboratoire d'Informatique Fondamentale de
  Lille, 1998.

\bibitem{koni-pere-00}
S.~Konieczny and R.~Pino~P{\'e}rez.
\newblock A framework for iterated revision.
\newblock {\em Journal of Applied Non-classical logics}, 10(3-4):339--367,
  2000.

\bibitem{koni-pere-17}
S.~Konieczny and R.~Pino~P{\'{e}}rez.
\newblock On iterated contraction: Syntactic characterization, representation
  theorem and limitations of the levi identity.
\newblock In {\em Eleventh International Conference on Scalable Uncertainty
  Management (SUM~2017)}, pages 348--362. Springer, 2017.

\bibitem{kren-88}
M.~W. Krentel.
\newblock The complexity of optimization problems.
\newblock {\em Journal of Computer and System Sciences}, 36:490--509, 1988.

\bibitem{lang-etal-08}
M.~Langlois, R.H. Sloan, B.~Sz{\"o}r{\'e}nyi, and G.~Tur{\'a}n.
\newblock Horn complements: Towards {H}orn-to-{H}orn belief revision.
\newblock In {\em Proceedings of the Twenty-Third AAAI Conference on Artificial
  Intelligence (AAAI~2008)}, pages 466--471, 2008.

\bibitem{lehm-95}
D.~Lehmann.
\newblock Belief revision, revised.
\newblock In {\em Proceedings of the Fourteenth International Joint Conference
  on Artificial Intelligence (IJCAI'95)}, pages 1534--1540, 1995.

\bibitem{libe-97-c}
P.~Liberatore.
\newblock The complexity of iterated belief revision.
\newblock In {\em Proceedings of the Sixth International Conference on Database
  Theory (ICDT'97)}, pages 276--290, 1997.

\bibitem{libe-16}
P.~Liberatore.
\newblock Belief merging by examples.
\newblock {\em {ACM} Transactions on Computational Logic}, 17(2):9:1--9:38,
  2016.

\bibitem{libe-scha-97}
P.~Liberatore and M.~Schaerf.
\newblock Reducing belief revision to circumscription (and viceversa).
\newblock {\em Artificial Intelligence}, 93(1--2):261--296, 1997.

\bibitem{libe-scha-00}
P.~Liberatore and M.~Schaerf.
\newblock Belief revision and update: Complexity of model checking.
\newblock {\em Journal of Computer and System Sciences}, 2000.
\newblock To appear.

\bibitem{meye-etal-02}
T.~Meyer, A.~Ghose, and S.~Chopra.
\newblock Syntactic representations of semantic merging operations.
\newblock In {\em Proceedings of the Seventh Pacific Rim International
  Conference on Artificial Intelligence (PRICAI~2002)}, page 620, 2002.

\bibitem{naya-94}
A.~Nayak.
\newblock Iterated belief change based on epistemic entrenchment.
\newblock {\em Erkenntnis}, 41:353--390, 1994.

\bibitem{naya-etal-03}
A.~Nayak, M.~Pagnucco, and P.~Peppas.
\newblock Dynamic belief revision operators.
\newblock {\em Artificial Intelligence}, 146(2):193--228, 2003.

\bibitem{nebe-98}
B.~Nebel.
\newblock How hard is it to revise a belief base?
\newblock In D.~Dubois and H.~Prade, editors, {\em Belief Change - Handbook of
  Defeasible Reasoning and Uncertainty Management Systems, Vol. 3}. Kluwer
  Academic, 1998.

\bibitem{papi-01}
O.~Papini.
\newblock Iterated revision operations stemming from the history of an agent's
  observations.
\newblock In {\em Frontiers in belief revision}, volume~22 of {\em Applied
  Logic Series}, pages 279--301. Springer, 2001.

\bibitem{pepp-etal-08}
P.~Peppas, A.M. Fotinopoulos, and S.~Seremetaki.
\newblock Conflicts between relevance-sensitive and iterated belief revision.
\newblock In {\em Proceedings of the Eighteenth European Conference on
  Artificial Intelligence (ECAI~2008)}, pages 85--88. IOS Press, 2008.

\bibitem{pepp-will-16}
P.~Peppas and M.A. Williams.
\newblock Kinetic consistency and relevance in belief revision.
\newblock In {\em Proceedings of the Fifthteenth European Conference on Logics
  in Artificial Intelligence (JELIA~2016)}, pages 401--414, 2016.

\bibitem{rott-03-a}
H.~Rott.
\newblock Coherence and conservatism in the dynamics of belief {II:} iterated
  belief change without dispositional coherence.
\newblock {\em Journal of Logic and Computation}, 13(1):111--145, 2003.

\bibitem{rott-06}
H.~Rott.
\newblock Revision by comparison as a unifying framework: Severe withdrawal,
  irrevocable revision and irrefutable revision.
\newblock {\em Theoretical Computer Science}, 355(2):228--242, 2006.

\bibitem{rott-09}
H.~Rott.
\newblock Shifting priorities: Simple representations for twenty-seven iterated
  theory change operators.
\newblock In D.~Makinson, J.~Malinowski, and H.~Wansing, editors, {\em Towards
  Mathematical Philosophy}, volume~28 of {\em Trends in Logic}, pages 269--296.
  Springer Netherlands, 2009.

\bibitem{rott-12}
H.~Rott.
\newblock Bounded revision: Two-dimensional belief change between conservative
  and moderate revision.
\newblock {\em Journal of Philosophical Logic}, 41(1):173--200, 2012.

\bibitem{rott-pagn-99}
H.~Rott and M.~Pagnucco.
\newblock Severe withdrawal (and recovery).
\newblock {\em Journal of Philosophical Logic}, 28(5):51--547, 1999.

\bibitem{sato-88}
K.~Satoh.
\newblock Nonmonotonic reasoning by minimal belief revision.
\newblock In {\em Proceedings of the International Conference on Fifth
  Generation Computer Systems (FGCS'88)}, pages 455--462, 1988.

\bibitem{saue-etal-20}
K.~Sauerwald, J.~Haldimann, M.~von Berg, and C.~Beierle.
\newblock Descriptor revision for conditionals: Literal descriptors and
  conditional preservation.
\newblock In {\em KI-2020: Advances in Artificial Intelligence - Forty-Third
  German Conference on AI}, pages 204--218. Springer, 2020.

\bibitem{saue-etal-20-b}
K.~Sauerwald, G.~Kern{-}Isberner, and C.~Beierle.
\newblock A conditional perspective for iterated belief contraction.
\newblock In {\em Proceedings of the Twenty-Fourth European Conference on
  Artificial Intelligence (ECAI~2020)}, pages 889--896. IOS Press, 2020.

\bibitem{scha-78}
T.~J. Schaefer.
\newblock The complexity of satisfiability problems.
\newblock In {\em Proceedings of the Tenth ACM Symposium on Theory of Computing
  (STOC'78)}, pages 216--226, 1978.

\bibitem{sege-98}
K.~Segerberg.
\newblock Irrevocable belief revision in dynamic doxastic logic.
\newblock {\em Notre Dame Journal of Formal Logic}, 39:287--306, 1998.

\bibitem{spoh-88}
W.~Spohn.
\newblock Ordinal conditional functions: A dynamic theory of epistemic states.
\newblock In {\em Causation in Decision, Belief Change, and Statistics}, pages
  105--134. Kluwer Academics, 1988.

\bibitem{spoh-99}
W.~Spohn.
\newblock Ranking functions, {AGM} style, 1999.

\bibitem{wagn-87}
K.~Wagner.
\newblock More complicated questions about maxima and minima, and some closures
  of {NP}.
\newblock {\em Theoretical Computer Science}, 51:53--80, 1987.

\bibitem{will-94}
M.~Williams.
\newblock Transmutations of knowledge systems.
\newblock In {\em Proceedings of the Fourth International Conference on the
  Principles of Knowledge Representation and Reasoning (KR'94)}, pages
  619--629, 1994.

\end{thebibliography}

\end{document}